\documentclass{article}

\PassOptionsToPackage{numbers, sort&compress}{natbib}
\PassOptionsToPackage{usenames,dvipsnames}{color}
\PassOptionsToPackage{usenames,dvipsnames}{xcolor}

\usepackage[eandd,final]{neurips_2026}

\usepackage{hyperref}
\usepackage{url}

\usepackage{amsmath,amsfonts,bm}

\def\eqref#1{equation~\ref{#1}}
\def\1{\bm{1}}

\DeclareMathAlphabet{\mathsfit}{\encodingdefault}{\sfdefault}{m}{sl}
\SetMathAlphabet{\mathsfit}{bold}{\encodingdefault}{\sfdefault}{bx}{n}

\usepackage[utf8]{inputenc} % enable UTF-8
\usepackage{lipsum} % write pseudo text

\usepackage[T1]{fontenc}

\usepackage{etoolbox}
\newbool{includeappendix}
\setbool{includeappendix}{true}

\ifdefined\isoverfull
\else
\fi

\usepackage{xcolor} % for definecolor

\definecolor{my-full-blue}{HTML}{1F77B4}
\definecolor{my-full-orange}{HTML}{FF7F0E}
\definecolor{my-full-green}{HTML}{2CA02C}
\definecolor{my-full-red}{HTML}{d62728}
\definecolor{my-full-purple}{HTML}{9467bd}
\colorlet{my-blue}{my-full-blue!30}
\colorlet{my-orange}{my-full-orange!30}
\colorlet{my-green}{my-full-green!30}
\colorlet{my-red}{my-full-red!30}
\colorlet{my-purple}{my-full-purple!30}

\usepackage{listings}

\usepackage{textcomp}

\usepackage{xcolor}

\usepackage[scaled=0.8]{beramono}

\definecolor{ckeyword}{HTML}{7F0055}
\definecolor{ccomment}{HTML}{3F7F5F}
\definecolor{cstring}{HTML}{2A0099}

\lstdefinestyle{numbers}{
	numbers=left,
	framexleftmargin=20pt,
	numberstyle=\tiny,
	firstnumber=auto,
	numbersep=1em,
	xleftmargin=2em
}

\lstdefinestyle{layout}{
	frame=none,
	captionpos=b,
}

\lstdefinestyle{comment-style}{
	morecomment=[l]//,
	morecomment=[s]{/*}{*/},
	commentstyle={\color{ccomment}\itshape},
}

\lstdefinestyle{string-style}{
	morestring=[b]",%
	morestring=[b]',%
	stringstyle={\color{cstring}},
	showstringspaces=false,%
}

\lstdefinestyle{keyword-style}{
	keywordstyle={\ttfamily\bfseries},
	morekeywords={
		function,
		constructor,
		int,
		bool,
		return,
		returns,
		uint
	},
	morekeywords = [2]{},
	keywordstyle = [2]{\text},
	sensitive=true,
}

\lstdefinestyle{input-encoding}{
	inputencoding=utf8,
	extendedchars=true,
	literate=
	{ℝ}{$\reals$}1%
	{→}{$\rightarrow$}1%
	{α}{$\alpha$}1%
	{β}{$\beta$}1%
	{λ}{$\lambda$}1%
	{θ}{$\theta$}1%
	{ϕ}{$\phi$}1%
	{∅}{$\emptyset$}1%
}

\lstdefinestyle{escaping}{
	moredelim={**[is][\color{blue}]{\%}{\%}},
	escapechar=|,
	mathescape=true
}

\lstdefinestyle{default-style}{
	basicstyle=\fontencoding{T1}\ttfamily\footnotesize,
	style=numbers,
	style=layout,
	style=comment-style,
	style=string-style,
	style=keyword-style,
	style=input-encoding,
	style=escaping,
	tabsize=2,
	upquote=true
}

\lstdefinelanguage{BASIC}{
	language=C++,
	style=default-style
}[keywords,comments,strings]%

\usepackage{booktabs}       % professional-quality tables
\usepackage{nicefrac}       % compact symbols for 1/2, etc.
\usepackage{microtype}      % microtypography
\usepackage[flushleft]{threeparttable}

\usepackage{url}

\usepackage{xcolor}
\usepackage{xspace}
\usepackage{etoolbox}
\usepackage{fontawesome}
\usepackage{enumitem}
\usepackage{amsmath,amssymb}
\usepackage{array}
\usepackage{multirow}
\usepackage{float}
\usepackage{wrapfig}
\usepackage{adjustbox,booktabs}
\usepackage{siunitx}
\usepackage{caption}
\usepackage{fontawesome}
\usepackage{stackengine}
\usepackage{svg}
\usepackage{mathtools}
\usepackage{xspace}
\usepackage{makecell}
\usepackage{graphicx}
\usepackage{booktabs}
\usepackage{longtable}
\usepackage{hyperref}
\usepackage[labelformat=simple]{subcaption}
\usepackage[most]{tcolorbox}
\usepackage{diagbox}
\usepackage{ulem}
\usepackage[capitalize,noabbrev]{cleveref}

\usepackage{colortbl}

\usepackage{tikz}

\usetikzlibrary{calc,decorations,decorations.pathmorphing}
\usetikzlibrary{positioning,fit,arrows}
\usetikzlibrary{decorations.markings}
\usetikzlibrary{shapes,shapes.geometric}
\usetikzlibrary{shadows,patterns,snakes}
\usetikzlibrary{backgrounds,decorations.pathreplacing,calligraphy,automata}
\usetikzlibrary{intersections}
\usetikzlibrary{angles,quotes}
\usetikzlibrary{plotmarks}
\usetikzlibrary{patterns}
\usetikzlibrary{arrows.meta}
\usetikzlibrary{shapes.misc}
\usetikzlibrary{chains}
\usetikzlibrary{shapes.callouts}

\usepackage{pgfplots}
\pgfplotsset{compat=1.18}

\usepackage{amsthm}
\usepackage{stmaryrd}

\theoremstyle{plain}

\theoremstyle{definition}

\theoremstyle{remark}

\newcolumntype{x}[2]{S[table-format=#1.#2,table-auto-round]}
\newcolumntype{M}{>{$}l<{$}}
\renewcommand{\emph}[1]{\textit{#1}}

\newcommand{\bench}{\textsc{ProofRank}\xspace}

\newcommand{\qwenthreebig}{\textsc{Qwen3.5-397B-A17B}\xspace}
\newcommand{\qwenthreeshort}{\textsc{Qwen3.5-397B}\xspace}

\newcommand{\opcrone}{\textsc{OPC-R1-8B}\xspace}

\newcommand{\geminipro}{\textsc{Gemini-3.1-Pro}\xspace}

\newcommand{\geminiflash}{\textsc{Gemini-3-Flash}\xspace}
\newcommand{\geminiflashseven}{\textsc{Gemini-3.7-Flash}\xspace}
\newcommand{\geminiflashlow}{\textsc{Gemini-3-Flash-Low}\xspace}

\newcommand{\gptfivefour}{\textsc{GPT-5.4}\xspace}
\newcommand{\gptfivesix}{\textsc{GPT-5.6-Sol}\xspace}

\newcommand{\qwenthreeeight}{\textsc{Qwen3-8B}\xspace}

\newcommand{\grokfourfast}{\textsc{Grok-4.1-Fast}\xspace}
\newcommand{\dsvthreetwo}{\textsc{DeepSeek-V3.2}\xspace}
\newcommand{\dsvfourpro}{\textsc{DeepSeek-V4-Pro}\xspace}
\newcommand{\gptoss}{\textsc{GPT-OSS-120B}\xspace}
\newcommand{\glmfive}{\textsc{GLM-5}\xspace}
\newcommand{\kimi}{\textsc{Kimi-K2.5-Think}\xspace}
\newcommand{\stepfun}{\textsc{Step-3.5-Flash}\xspace}

\newcommand{\answerbench}{\textsc{IMO-AnswerBench}\xspace}
\newcommand{\ma}{\textsc{MathArena}\xspace}

\definecolor{acceptblue}{HTML}{6494EA}
\hypersetup{citecolor=acceptblue}

\definecolor{lightred}{HTML}{ffcbc7}
\definecolor{gemini}{HTML}{4285F4}
\definecolor{claude}{HTML}{f3e9d7}
\definecolor{deepseek}{HTML}{FADA4B}
\definecolor{qwen}{HTML}{FA574B}
\definecolor{oai}{HTML}{10a37f}
\definecolor{LightGreen}{HTML}{CCFFCC}

\lstdefinestyle{mystyle}{
    breaklines=true,
    basicstyle=\scriptsize\ttfamily,
    numbers=none,
    language={},
    framextopmargin=0pt,
    framexbottommargin=0pt,
    breakindent=0pt,
    showspaces = false,
    keywordstyle=\bfseries,
    showstringspaces=false,
    columns=fullflexible,
    morekeywords={Answer}
    moredelim=[**][\bfseries]{!!}
}

\newtcblisting{prompt}[2][]{
    arc=3pt, outer arc=3pt,
    width=\linewidth,
    left=1mm,
    top=0mm,
    bottom=0mm,
    title=#2, 
    colback=lightred!5!white,
    colframe=black,
    fonttitle=\bfseries,
    listing only, 
    listing options={style=mystyle},
    breakable,
    #1
}

\newtcblisting{smallprompt}[2][]{
    arc=3pt, outer arc=3pt,
    width=0.9\linewidth,
    left=1mm,
    top=0mm,
    bottom=0mm,
    title=#2, 
    colback=lightred!5!white,
    colframe=black,
    fonttitle=\bfseries,
    listing only, 
    listing options={style=mystyle},
    breakable,
    #1
}

\newtcblisting{response}[2][]{
    arc=3pt, outer arc=3pt,
    width=0.9\linewidth,
    left=1mm,
    top=0mm,
    bottom=0mm,
    title=#2, 
    colback=lightred!5!white,
    colframe=oai,
    fonttitle=\bfseries,
    listing only, 
    listing options={style=mystyle},
    breakable,
    #1
}

\definecolor{glm}{HTML}{6C4BB6}
\definecolor{stepfun}{HTML}{0B7285}
\definecolor{exgray}{HTML}{4A4A4A}

\newtcolorbox{exproblem}[2][]{
    enhanced, breakable,
    arc=3pt, outer arc=3pt,
    width=\linewidth,
    colback=exgray!4!white,
    colframe=exgray,
    coltitle=white,
    fonttitle=\bfseries\small,
    fontupper=\small,
    boxrule=0.7pt,
    left=2mm, right=2mm, top=1mm, bottom=1mm,
    title={#2},
    #1
}

\newtcolorbox{exverdict}[2][]{
    enhanced, breakable,
    arc=3pt, outer arc=3pt,
    width=\linewidth,
    colback=lightred!60!white,
    colframe=my-full-red,
    coltitle=white,
    fonttitle=\bfseries\small,
    fontupper=\small,
    boxrule=0.7pt,
    left=2mm, right=2mm, top=1mm, bottom=1mm,
    title={#2},
    #1
}

\usepackage{paracol}
\usepackage{needspace}

\tcbset{exsolstyle/.style={
    enhanced, breakable,
    arc=3pt, outer arc=3pt,
    width=\linewidth,
    coltitle=white,
    fonttitle=\bfseries\footnotesize,
    fontupper=\footnotesize,
    boxrule=0.7pt,
    left=1.5mm, right=1.5mm, top=1mm, bottom=1mm,
    boxsep=1mm,
    before upper={\setlength{\parindent}{0pt}\setlength{\parskip}{0.4em}},
}}

\newenvironment{excompare}[4]{%
    \par\medskip
    \Needspace*{9\baselineskip}
    \setlength{\columnsep}{5mm}%
    \def\exswitch{%
        \end{tcolorbox}%
        \switchcolumn
        \begin{tcolorbox}[exsolstyle, colframe=#3, colback=#3!5!white, title={#4}]%
    }%
    \begin{paracol}{2}%
    \let\clearpage\newpage
    \begin{tcolorbox}[exsolstyle, colframe=#1, colback=#1!5!white, title={#2}]%
}{%
    \end{tcolorbox}%
    \end{paracol}%
    \par\medskip
}

\newtcolorbox{exsolution}[3][]{
    exsolstyle,
    colframe=#2,
    colback=#2!5!white,
    title={#3},
    fonttitle=\bfseries\small,
    fontupper=\small,
    #1
}

\usepackage[capitalize]{cleveref}

\crefformat{section}{\S#2#1#3}

\crefrangeformat{section}{\S#3#1#4\crefrangeconjunction\S#5#2#6}

\crefmultiformat{section}{\S#2#1#3}{\crefpairconjunction\S#2#1#3}{\crefmiddleconjunction\S#2#1#3}{\creflastconjunction\S#2#1#3}

\newcommand{\crefrangeconjunction}{--}

\crefname{listing}{Lst.}{listings}
\crefname{line}{Lin.}{Lin.}
\crefname{appendix}{App.}{App.}
\crefname{subappendix}{App.}{App.}
\crefname{subsubappendix}{App.}{App.}

\newcommand{\app}[1]{%
	\ifbool{includeappendix}{\cref{#1}}{the appendix}%
}
\newcommand{\App}[1]{%
	\ifbool{includeappendix}{\cref{#1}}{The appendix}%
}

\title{Not All Proofs Are Equal: Evaluating LLM Proof Quality Beyond Correctness}
\author{Ivo Petrov\textsuperscript{1}, Jasper Dekoninck\textsuperscript{2}, Dimitar I. Dimitrov\textsuperscript{1}, Martin Vechev\textsuperscript{1,2}
\\
\textsuperscript{1}INSAIT, Sofia University "St. Kliment Ohridski", \textsuperscript{2}ETH Zurich \\
\texttt{\{ivo.petrov\footnote{Corresponding author.}, martin.vechev\}@insait.ai} \\
\vspace{1mm}\\
\raisebox{-0.16em}{\includegraphics[height=1em]{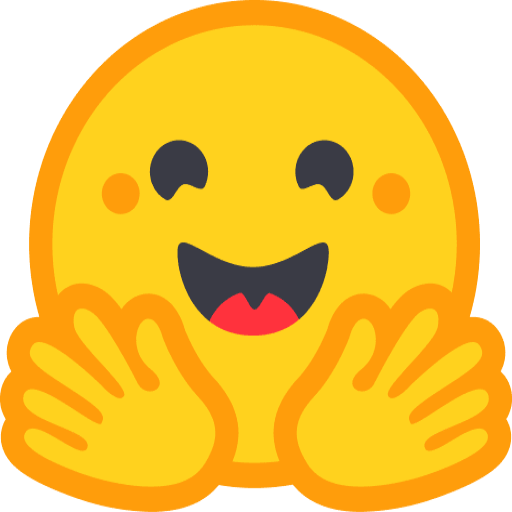}} \href{https://huggingface.co/collections/INSAIT-Institute/proofrank-evaluating-llm-proof-quality-beyond-correctness-6ae8b4ad6d319934e1af33f2}{INSAIT-Institute/ProofRank}
\hspace{12mm}
\raisebox{-0.16em}{\includegraphics[height=1em]{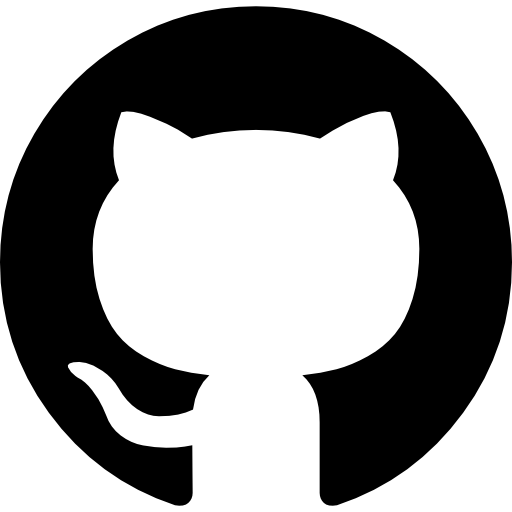}} \href{https://github.com/insait-institute/proofrank}{insait-institute/proofrank} 
\vspace{-9mm}\\ 
\\
}

\begin{document}
\maketitle

\begin{abstract}
Large language models (LLMs) have become capable mathematical problem-solvers, often producing correct proofs for challenging problems. However, correctness alone is not sufficient: mathematical proofs should also be clear, insightful, contain transferable knowledge, and be generally useful to the reader. While \emph{proof quality} is subjective and depends on the reader and context, many of its components can be made concrete. In this work, we identify such components and introduce \bench{}, a benchmark curated from challenging competitions that measures several scalable proxies of proof quality: (i) conciseness -- quantifying whether proofs avoid unnecessary steps; (ii) computational ease -- the extent to which a proof relies on tedious calculations; (iii) cognitive simplicity -- how accessible the used proof techniques are; (iv) diversity -- how varied a model's proofs for a single problem are; and (v) adaptivity -- how well a model can follow a specified technique. We find proof quality varies substantially across models along dimensions that correctness largely fails to capture. Every proof-level proxy can be steered through prompting, but often at a moderate correctness cost. \bench{} thus captures important aspects of mathematical proofs beyond correctness, and indicates the need for models that produce proofs which are not only correct but also valuable.
\end{abstract}

\section{Introduction}

Large language models (LLMs) are increasingly used as mathematical collaborators and are now actively assisting with research on open problems \citep{schmitt2025extremaldescendantintegralsmoduli,ghrist2024latticevaluedbottleneckduality,raz2025euleraiunifyingformulas}. A central capability that enables this is theorem proving: an LLM must not only arrive at correct claims, but also justify them rigorously. As a result, proof-based evaluation has become an important component of mathematical reasoning benchmarks, through both human expert evaluation \citep{prooforbluff,opc,improofbench,brainvbytes} and automated LLM judges \citep{imobench}.

\begin{figure*}[t]
    \centering
    \resizebox{\linewidth}{!}{%
        \input{paper_files/figures/overview}
    }
    \caption{Example of proof quality. Even for the same problem, correct LLM-generated proofs can differ substantially in proof style. This example illustrates all metrics in \bench, including adaptivity to a specific proof technique (fourth LLM-written proof).}
    \vspace{-1.5em}
    \label{fig:overview}
\end{figure*}

\paragraph{Proof quality} 
Existing proof-writing benchmarks have so far almost entirely focused on proof correctness. However, recent work on AI-assisted mathematics argues that proof value extends beyond formal correctness to include understanding, motivation, and broader mathematical context~\citep{klowden2026mathematicalmethods}. Correct proofs can differ substantially in how clearly they communicate the central idea, how much insight they provide, and whether that insight transfers to related problems or suggests new questions. We refer to these alternative dimensions of a good proof as \emph{proof quality}. Current benchmarks do not capture such distinctions and treat all correct proofs as equivalent. 

\paragraph{Scalable proxies for proof quality}
Evaluating proof quality is challenging because it is inherently subjective: mathematicians may disagree about which proof is more elegant, and the appropriate level of detail can depend on the reader's background. This makes it difficult to define a single absolute score for proof quality. Instead, a scalable benchmark should focus on measurable aspects of proofs that serve as proxies for proof quality. For example, an overreliance on computation is a concrete property that can be evaluated automatically and captures a relevant aspect of proof quality.

\paragraph{This work: \bench}
In this work, we introduce \bench, the first benchmark designed specifically to evaluate these scalable proxies of proof quality in LLM-generated proofs. \bench consists of 382 challenging, high-school level, final-answer competition problems spanning algebra, geometry, combinatorics, calculus, and number theory. We specifically choose to use final-answer problems because they make correctness evaluation substantially more scalable and help avoid known biases in full-proof correctness judgments \citep{opc}.

\vspace{-0.25em}
\paragraph{Metrics in \bench}
As shown in \cref{fig:overview}, we consider five complementary metrics capturing different aspects of proof quality. We group these metrics into two types. \emph{Proof-level} metrics evaluate individual proofs for the same problem. In \bench, these are conciseness, computational ease, and cognitive simplicity, which capture whether a proof is compact, avoids unnecessary computation, and is easy to follow. \emph{Problem-level} metrics evaluate a model's behavior across multiple attempts at the same task. In \bench, these are diversity and adaptivity, which measure whether a model can produce distinct valid approaches and whether it can follow a requested technique. Most metrics have no strictly preferable direction, hence we propose possible interpretations for them in \cref{sec:methodology-metrics}.

\vspace{-0.25em}
\paragraph{Pairwise comparisons and correctness}
\begin{wrapfigure}{r}{0.45\linewidth}
    \centering
    \vspace{-4mm}
    \resizebox{\linewidth}{!}{%
        % Auto-generated by paper_files/figures/generate_intro_metric_radar.py
\begin{tikzpicture}[font=\scriptsize, line cap=round, line join=round, inner sep=0.5pt, outer sep=0pt]
\definecolor{radargptfivesix}{HTML}{1F77B4}
\definecolor{radardsvfourpro}{HTML}{FF7F0E}
\definecolor{radargeminipro}{HTML}{D62728}
\definecolor{radarglmfive}{HTML}{2CA02C}
\definecolor{radarqwenthreebig}{HTML}{E377C2}
\draw[black!12, line width=0.3pt] (0.000,0.562) -- (0.487,0.281) -- (0.487,-0.281) -- (0.000,-0.562) -- (-0.487,-0.281) -- (-0.487,0.281) -- cycle;
\draw[black!12, line width=0.3pt] (0.000,1.125) -- (0.974,0.562) -- (0.974,-0.562) -- (0.000,-1.125) -- (-0.974,-0.563) -- (-0.974,0.563) -- cycle;
\draw[black!12, line width=0.3pt] (0.000,1.688) -- (1.461,0.844) -- (1.461,-0.844) -- (0.000,-1.688) -- (-1.461,-0.844) -- (-1.461,0.844) -- cycle;
\draw[black!12, line width=0.3pt] (0.000,2.250) -- (1.949,1.125) -- (1.949,-1.125) -- (0.000,-2.250) -- (-1.949,-1.125) -- (-1.949,1.125) -- cycle;
\draw[black!18, line width=0.3pt] (0,0) -- (0.000,2.250);
\node[align=center, text=black] at (0.000,2.498) {Accuracy};
\draw[black!18, line width=0.3pt] (0,0) -- (1.949,1.125);
\node[align=center, text=black] at (2.163,1.249) {Concise};
\draw[black!18, line width=0.3pt] (0,0) -- (1.949,-1.125);
\node[align=center, text=black] at (2.163,-1.309) {Comp.\ ease};
\draw[black!18, line width=0.3pt] (0,0) -- (0.000,-2.250);
\node[align=center, text=black] at (0.000,-2.498) {Cognitive};
\draw[black!18, line width=0.3pt] (0,0) -- (-1.949,-1.125);
\node[align=center, text=black] at (-2.163,-1.209) {Diversity};
\draw[black!18, line width=0.3pt] (0,0) -- (-1.949,1.125);
\node[align=center, text=black] at (-2.163,1.249) {Adaptivity};
\path[fill=radargptfivesix, fill opacity=0.16] (0.000,2.102) -- (1.949,1.125) -- (1.949,-1.125) -- (0.000,-1.282) -- (-0.816,-0.471) -- (-1.949,1.125) -- cycle;
\draw[radargptfivesix, line width=0.90pt] (0.000,2.102) -- (1.949,1.125) -- (1.949,-1.125) -- (0.000,-1.282) -- (-0.816,-0.471) -- (-1.949,1.125) -- cycle;
\path[fill=radardsvfourpro, fill opacity=0.16] (0.000,1.620) -- (1.949,1.125) -- (1.949,-1.125) -- (0.000,-1.423) -- (-1.086,-0.627) -- (-1.537,0.887) -- cycle;
\draw[radardsvfourpro, line width=0.90pt] (0.000,1.620) -- (1.949,1.125) -- (1.949,-1.125) -- (0.000,-1.423) -- (-1.086,-0.627) -- (-1.537,0.887) -- cycle;
\path[fill=radargeminipro, fill opacity=0.16] (0.000,1.366) -- (0.000,0.000) -- (0.553,-0.319) -- (0.000,-1.038) -- (-1.186,-0.685) -- (-1.130,0.653) -- cycle;
\draw[radargeminipro, line width=0.90pt] (0.000,1.366) -- (0.000,0.000) -- (0.553,-0.319) -- (0.000,-1.038) -- (-1.186,-0.685) -- (-1.130,0.653) -- cycle;
\path[fill=radarglmfive, fill opacity=0.16] (0.000,1.105) -- (0.122,0.070) -- (0.071,-0.041) -- (0.000,-1.074) -- (-0.809,-0.467) -- (-0.826,0.477) -- cycle;
\draw[radarglmfive, line width=0.90pt] (0.000,1.105) -- (0.122,0.070) -- (0.071,-0.041) -- (0.000,-1.074) -- (-0.809,-0.467) -- (-0.826,0.477) -- cycle;
\path[fill=radarqwenthreebig, fill opacity=0.16] (0.000,0.659) -- (0.611,0.353) -- (1.496,-0.863) -- (0.000,-1.167) -- (-0.913,-0.527) -- (-0.514,0.297) -- cycle;
\draw[radarqwenthreebig, line width=0.90pt] (0.000,0.659) -- (0.611,0.353) -- (1.496,-0.863) -- (0.000,-1.167) -- (-0.913,-0.527) -- (-0.514,0.297) -- cycle;
\draw[radargptfivesix, line width=1pt] (-2.20,-2.82) -- (-1.88,-2.82);
\node[anchor=west, text=black] at (-1.80,-2.82) {GPT-5.6-Sol};
\draw[radardsvfourpro, line width=1pt] (0.15,-2.82) -- (0.47,-2.82);
\node[anchor=west, text=black] at (0.55,-2.82) {DeepSeek-V4};
\draw[radargeminipro, line width=1pt] (-2.20,-3.14) -- (-1.88,-3.14);
\node[anchor=west, text=black] at (-1.80,-3.14) {Gemini-3.1};
\draw[radarglmfive, line width=1pt] (0.15,-3.14) -- (0.47,-3.14);
\node[anchor=west, text=black] at (0.55,-3.14) {GLM-5};
\draw[radarqwenthreebig, line width=1pt] (-2.20,-3.46) -- (-1.88,-3.46);
\node[anchor=west, text=black] at (-1.80,-3.46) {Qwen3.5-397B};
\end{tikzpicture}
    }
    \caption{Main results for several models.}
    \label{fig:intro-metric-radar}
    \vspace{-4mm}
\end{wrapfigure}
Absolute scores are hard to calibrate across problems: what counts as concise for one problem may be verbose for another. To address this, we compare model outputs pairwise for each problem and metric, then aggregate these comparisons into model-level ratings. Additionally, we only compare two outputs when both are correct, meaning they reach the correct final answer and provide a complete justification. This is important because we would otherwise reward the use of very simple but wrong proofs.

\vspace{-0.25em}
\paragraph{Experimental results}
As shown in \cref{fig:intro-metric-radar}, we find substantial differences in proof quality across models. The most accurate models also take the highest rating on four of the five quality metrics, where the lead is split: \dsvfourpro{} produces the most concise and least computation-heavy proofs, while \gptfivesix{} is the most adaptive. In contrast, \geminipro{} has the lowest conciseness rating: its solutions are on average $3.5$ times more verbose than necessary. We also find that most quality metrics are only weakly correlated with accuracy, suggesting that they capture distinct aspects of proof quality that are not fully aligned with correctness. For instance, \qwenthreeshort{} scores higher on our metrics than its low accuracy would suggest, while \glmfive{} scores low on most, despite achieving reasonably good accuracy. Finally, we analyze the models' behavior under different prompting strategies aiming to modify the response style, and find that every proof-level metric can be steered, under some potential utility cost, but each model's default proof-style fingerprint is largely retained.

\paragraph{Key contributions} Our key contributions are:
\begin{itemize}[itemsep=0em,leftmargin=1em]
\item Scalable metrics for conciseness, computational ease, cognitive simplicity, diversity, and adaptivity (\cref{sec:methodology-metrics}).
\item \bench, the first benchmark for evaluating LLM-generated proofs beyond correctness (\cref{sec:methodology-bench}).
\item An evaluation of frontier LLMs on \bench, revealing trade-offs in proof quality (\cref{sec:evaluation}).
\item An analysis of how the LLMs' proof style can be steered through prompting (\cref{sec:steerability}).
\end{itemize}
\vspace{-2mm}

\section{Related Work}\label{sec:related}
We discuss related work on LLM mathematical reasoning, focusing on proof generation and quality.

\paragraph{Proof-generation with LLMs} Recent years have seen significant interest in leveraging LLMs for mathematical proof generation, with several works analyzing or benchmarking model performance on this task \citep{prooforbluff,improofbench,brainvbytes,frieder2024,litmus,putnamlike,qedbench,lemmabench}. These studies show rapid progress and indicate that LLMs can now provide meaningful assistance to professional mathematicians \citep{improofbench}. They also highlight persistent limitations, including sycophancy \citep{brokenmath} and overreliance on brittle pattern matching \citep{brainvbytes}. Since manual proof evaluation is costly, several works have investigated LLMs as judges \citep{opc,imobench,mareliablefinegrainedeval,ineqmath}. Related proof-verification benchmarks instead evaluate whether models can grade proofs \citep{opc,hardverify,refgrader,mareliablefinegrainedeval}. These studies show very high agreement with humans, but also reveal systematic biases \citep{opc,mareliablefinegrainedeval}.

\paragraph{Correctness beyond final answers} As LLM-generated proofs become valuable artifacts in practitioners' work, automatically verifying these informal proofs becomes critical for transparency and to prevent false discoveries. While mathematical test suites usually verify for correctness using final-answer verification, multiple works show that the end-to-end proofs may be incomplete or even false~\citep{prooforbluff,opc,brainvbytes}. To address this gap, a separate line of work strengthens the notion of correctness itself, aiming to investigate or validate the intermediate reasoning via step-by-step verification~\citep{oracle,reasoneval,processbench,hardverify}, assigning step-wise rewards~\citep{verifystepbystep,rightisnotenough}, formalizing the proof in a proof assistant~\citep{dtv,tpjudge,logicreward}, or by determining the feasibility of LLM judges~\citep{opc,mareliablefinegrainedeval}. In our work, full-proof correctness is instead a prerequisite, where we compare only proofs that are already correct and complete, and our metrics target properties that are tangential to correctness.  

\paragraph{Proof quality evaluation} While most existing work centers on correctness, recent studies emphasize that correctness alone is not sufficient for LLM-generated mathematics. In particular, \citet{evalmathlm} analyzed the relationship between correctness and perceived helpfulness, finding that these metrics are correlated but can diverge substantially. \citet{frieder2024} also touch on presentation-related irregularities in their human evaluation of LLM-generated proofs. This fits into a broader shift toward evaluating mathematical models beyond raw accuracy, including reliability \citep{reliablemath,abstentionbench}, interaction-based evaluation \citep{interactivebenchmarks}, and structured skill profiles \citep{gauss}. Proof-based benchmarks also often include quantitative error taxonomies \citep{brainvbytes,frieder2024} and qualitative analyses of proof style \citep{prooforbluff,improofbench}, providing insight into issues such as missing structure and hallucinated reasoning. However, these analyses are typically limited in scope and not designed to evaluate multiple dimensions systematically. Our work complements these efforts by quantifying multiple dimensions of proof quality.

\paragraph{LLMs in mathematical education} The impact of LLMs on mathematical education has been a topic of active research \citep{steinbach2025llmhallucinate,pepin2025scoping,turmuzi2026chatgpt,reddig2025generating}. In educational settings, proof quality is often as important as correctness: explanations must be clear and aligned with the learner's expertise. Recent work highlights risks such as hallucinations and misleading feedback \citep{steinbach2025llmhallucinate}, but also demonstrates the potential of LLMs to generate instructional material and guidance \citep{reddig2025generating}.

\paragraph{Formal proof assistants} Another line of work integrates LLMs with formal proof assistants like Lean \citep{lean}, enabling formally verified proofs \citep{hilbert,liu2026numinaleanagentopengeneralagentic,goedelproverv2}. Formal benchmarks range from high-school and undergraduate mathematics \citep{putnambench,proofnet,minif2f} to research-level or real-world formalization tasks \citep{formalproofbench,fate,ravi2026proofbench,sorrydb,taobench}. While formal verification reduces the need to judge correctness in natural language, proof quality remains important. For instance, \citet{liu2026numinaleanagentopengeneralagentic} identify unstructured proofs as a limitation, and \citet{proofoptimizer} develop techniques to reduce the length of formal proofs generated by LLMs. %Our work is complementary and focuses on natural-language proofs.

\paragraph{LLM-as-a-judge} Finally, since many of our metrics use LLMs as judges, we briefly discuss related work in this area. LLM judges are widely applied to open-ended tasks such as instruction-following and question-answering \citep{llmasajudge,wildbench}, as well as mathematical proof generation \citep{imobench,mareliablefinegrainedeval}. These works show that LLM judges can achieve high agreement with human evaluators, but may also introduce systematic biases \citep{justiceprejudice}, including self-preference \citep{selfpreferencebias,selfpreferencebias2}, verbosity bias \citep{verbositybias}, and positional bias \citep{positionbias}. Our metric design explicitly accounts for these biases.

\vspace{-0.5em}
\section{Methodology}
\label{sec:methodology}

\vspace{-0.5em}
In this section, we describe how \bench evaluates proof quality. We first discuss what motivates our use of scalable proxies and pairwise comparisons (\cref{sec:methodology-objective-proxies}). We then define our proof-quality metrics (\cref{sec:methodology-metrics}), and finally describe how these metrics are instantiated in \bench (\cref{sec:methodology-bench}).

\vspace{-0.5em}
\subsection{Measuring Proof Quality Through Scalable Proxies}
\label{sec:methodology-objective-proxies}

\vspace{-1mm}
\paragraph{Proof quality is subjective}
Proof quality is inherently subjective. Mathematicians may disagree about the quality of a proof for many reasons, including differences in background and taste. The appropriate level of detail also depends on the reader: a short solution may be ideal for an expert, but unhelpful for a student trying to learn a new concept. For this reason, there is no universally accepted way to determine, in an absolute sense, which model produces the highest-quality proofs.

\vspace{-0.5em}
\paragraph{Scalable proxies}
Instead, we measure concrete properties that serve as scalable proxies for aspects of proof quality. These proxies are diagnostic rather than prescriptive. For example, a conciseness metric can reveal which model produces shorter proofs, without assuming that every mathematician would always prefer the shortest proof. These measurements are valuable because they are interpretable, scalable, and expose systematic model tendencies. All but one of our metrics have no universally preferred direction, with either end being the right choice for some readers. We therefore intend \bench{} primarily as a tool for matching a model to a use case, and note for each metric below how readers may be served by each of its ends, with an extended discussion given in \cref{app:interpretation}.

\vspace{-0.5em}
\paragraph{Unit of comparison} 
Proof-quality proxies can apply to individual proofs or to the models that produce them. Some valuable metrics are not properties of any single proof, but capture how useful and high-quality a model's proofs are overall. For instance, a model that produces a more diverse set of solutions to the same problem may be more useful because it can provide different insights and make it easier to adapt ideas to related problems. To capture this, we consider two types of metrics: \emph{proof-level} metrics, which compare individual proofs, and \emph{problem-level} metrics, which evaluate a model across multiple solutions to the same problem. Both types are then used to compare models.

\vspace{-0.5em}
\paragraph{Pairwise comparisons}
Even for scalable proxies, assigning an absolute score to a proof can be difficult. One problem may admit a short conceptual solution, while another may inherently require computations. Thus, a proof that appears computationally heavy for one problem may be unavoidable for another. To avoid calibrating scores across problems, we use pairwise comparisons between two models on the same problem. This makes differences easier to interpret: one can compare which of two proofs is more concise or computationally easier without needing an absolute scale across all problems. These comparisons can then be converted into model rankings using rating systems~\citep{llmasajudge}.

\vspace{-0.5em}
\paragraph{Correctness as a prerequisite}
A proof can only be of high quality if it is correct. In our experiments, we found that LLMs often produce elegant-looking but incorrect proofs when they fail to solve a problem. Such solutions provide no mathematical value, but can artificially inflate the quality metrics of weaker models. It is therefore important to separate correctness from proof-quality evaluation, so that quality metrics reflect properties of valid proofs rather than artifacts of incomplete or incorrect solutions. We treat correctness as a prerequisite for proof-quality evaluation: proof quality is compared between two models only on problems where both models produced correct solutions.

\vspace{-0.5em}
\paragraph{Final-answer problems} 
Automatic verification of natural-language proofs is challenging, and LLM judge biases may propagate into downstream evaluation. For instance, prior work \citep{opc} finds that LLMs can approach human performance on correctness classification, but may present weaker performance on their own solutions. Reliable open-ended verification also requires the strongest models as verifiers, which is prohibitively expensive at scale. Thus, we use final-answer problems as proof-writing tasks: models still produce complete written solutions that we verify in full, but the known answer lightens the load for the weaker judge.

\vspace{-0.5em}
\subsection{Scalable Proxies for Proof Quality}
\label{sec:methodology-metrics}
\vspace{-1mm}

\bench evaluates five complementary metrics: conciseness, computational ease, cognitive simplicity, diversity, and adaptivity. The first three are proof-level metrics that compare pairs of correct solutions to the same problem, while the last two are problem-level metrics that compare all solutions produced by different models on the same problem. All prompts are given in \cref{app:prompts}, and we compare our metrics against simpler methodological baselines we initially considered in \cref{app:cheaper-baselines}. 

\paragraph{Conciseness}
Verbose proofs can be cumbersome to read and verify, and may contain redundant steps that obscure the main ideas. However, detail is not inherently negative: an expert may prefer the most compressed argument because they can supply the omitted steps themselves, while a novice benefits from the additional detail filling gaps in their knowledge. We therefore measure where a model falls on this preference axis. To quantify conciseness, absolute proof length is not a perfect measure of conciseness, since some proofs inherently require more steps than others. To address this, we measure the \emph{compressibility} of a proof: how much of the text can be removed without changing the logical argument. Specifically, we ask a strong reasoning model to shorten each proof as much as possible while preserving logical equivalence. We then extract valid English words from the original proof $p$ and its shortened version $\tilde p$, and define the compressibility ratio $C(p) = |p|/|\tilde p|$.

Larger values indicate that more of the original text can be removed without changing the argument. To obtain a pairwise comparison, we compare $C(p_a)$ and $C(p_b)$ for two proofs $p_a$ and $p_b$ of the same problem. If $\frac{|C(p_a) - C(p_b)|}{\max(C(p_a), C(p_b))} > 0.1$, we prefer the more concise proof. Otherwise, to avoid treating harmless variation in wording as meaningful, we consider the two proofs equally concise. We show that this methodology is robust to the choice of compression model in \cref{app:rephraser-verbosity}. 

\paragraph{Computational ease and cognitive simplicity}
Correct proofs can be difficult to verify for two reasons: they may rely on heavy computation, or they may involve difficult logical dependencies and non-obvious conceptual ideas. We separate these two effects. Computational ease measures how well a proof avoids brute-force algebra, enumeration, case checking, or other calculation-heavy reasoning. Computation-heavy proofs are harder to check notationally, but are often more rigorous and less likely to omit small yet important steps. Cognitive simplicity instead measures how easy the argument is to follow, including the complexity of the techniques used and the number of nontrivial connections required. Cognitively simpler proofs are faster to verify and reach a broader audience, but may rely on less novel ideas. For both metrics, we compare pairs of correct solutions to the same problem and ask an LLM judge (\gptoss{}) which proof is better on the relevant dimension. Details on the validation of these metrics are given in \cref{app:human-validation}.

\paragraph{Diversity}
Models may internally have a tendency to use a wider or more narrow range of techniques. A more diverse model provides different insights and makes it easier to adapt ideas to related problems, which is valuable for research use and for presenting alternative solutions, while a less diverse model is more consistent and therefore more reproducible across users. To measure diversity, we sample $k=4$ solutions from each model, filter the correct ones, and cluster them by proof technique.

To perform this clustering, we first ask an LLM (\gptoss{}) to extract a proof summary for each solution, summarizing the main techniques and mathematical tools used. This allows different approaches to be compared without the distraction of heavy computational details or verbose textual information. We then provide all summaries from all models in another LLM call and ask it to cluster them into groups of similar techniques. In particular, we require the LLM to compare the relevant proof style, tools and theorems used, and overall proof structure, while disregarding minor stylistic, notational, and computational differences. The clustering procedure is validated in \cref{app:diversity-consistency}.

We quantify diversity through the entropy of the resulting empirical distribution. If, for a model $m_i$, $P(C_j \mid m_i)$ is its empirical distribution over technique clusters $C_1,\ldots,C_q$, the entropy is:
\begin{equation*}
    H(m_i) = - \sum_{j=1}^{q} P(C_j \mid m_i) \log P(C_j \mid m_i).
\end{equation*}
To obtain a pairwise comparison, we compare $H(m_i)$ and $H(m_{i'})$ for two models $m_i$ and $m_{i'}$ on the same problem, and prefer the model with higher entropy.

\paragraph{Adaptivity}
Adaptivity measures whether a model can produce a correct proof while following a requested technique. This is an important capability: a student may want to learn how to solve a problem using a particular method within their curriculum, while a researcher may want to explore whether a problem can be solved using different approaches. Unlike the other four metrics, adaptivity therefore has an unambiguously preferred direction. To measure it, for each benchmark problem, we first scrape human solutions and use an LLM to summarize these solutions and cluster them by technique, as we do for the diversity metric. Then, for each problem-technique pair $(p,t)$, we prompt the model to solve problem $p$ using the specified technique $t$. An attempt is counted as successful only if the resulting solution is correct and follows the requested technique. To determine whether the solution follows the requested technique, we give an LLM judge the technique description and ask whether the solution uses it in a non-trivial and logically sound way. We validate this judge in \cref{app:technique-verifier}. Finally, to form the pairwise comparisons, we compare all model pairs on each prompt, assigning a victory only if exactly one of the two attempts is successful, and a tie otherwise.
\vspace{-1mm}
\subsection{\bench}
\label{sec:methodology-bench}
\label{sec:methodology-dataset}
\vspace{-1mm}

\bench instantiates these metrics on 382 final-answer problems from established benchmarks, namely \ma \citep{matharena} and \answerbench \citep{imobench}. These tasks cover a wide range of domains, including Algebra, Geometry, Combinatorics, Number Theory, and Calculus. We also collect public solutions for these problems to provide a reference set of human-observed techniques. Although all problems are final-answer problems, most originate from challenging proof-based competitions, and we require a rigorous proof for every evaluated solution. Additional details on problem selection are given in \cref{app:dataset}.

\vspace{-1mm}
\paragraph{Correctness verification}
To instantiate the correctness filter, we use a two-stage heuristic:

\begin{itemize}[itemsep=0em,leftmargin=1em]
    \item \textbf{Final-answer correctness}: Following \citet{imobench}, we use \gptoss{} to determine whether the model's final answer matches the ground truth.
    \item \textbf{Completeness}: We then ask \gptoss{} to verify that the presented solution is self-contained: each step should follow from the problem statement, a previous step, or a standard mathematical fact. This filters out incomplete or lucky solutions.
\end{itemize}

A model's accuracy is the fraction of problems where it produces a solution passing both checks.

To validate this filter, we compare it against \gptfivefour{} judgments of full proof correctness on 200 randomly sampled solutions, obtaining $93.5\%$ accuracy after manual resolution of discrepancies between \gptfivefour{} and our filter. We describe this experiment in more detail in \cref{app:correctness-verifier}.

However, this filter introduces a discrepancy between models: weaker models have fewer correct solutions, so their quality metrics are computed on a smaller subset, generally consisting of easier problems. While absolute scores would be affected significantly by this discrepancy, pairwise comparisons are less affected. Two models are only compared on problems where both produced correct solutions, so differences in the difficulty distribution of solved problems do not substantially affect their relative quality scores. To demonstrate this, we confirmed that filtering for highly-solved problems still produces consistent rankings ($\rho\geq 0.95$) in \cref{app:highly-solved}. In particular, additional comparisons on problems solved only by stronger models primarily affect the relative ratings of those stronger models, rather than giving weaker models credit for problems they did not solve.

\paragraph{Bradley-Terry ranking}
To convert pairwise outcomes into model rankings, we fit a Bradley-Terry rating model for each pairwise metric. If model $m_a$ is compared against model $m_b$, the probability that $m_a$ wins is modeled as
\begin{equation*}
P(m_a \succ m_b) = \frac{1}{1 + e^{-(\beta_a - \beta_b + \gamma)}} ,
\end{equation*}
where $\beta_a$ and $\beta_b$ are model-specific parameters, and $\gamma$ is an intercept that captures systematic preference for the first solution, a known issue for LLM judges called position bias \citep{positionbias}. As in prior work \citep{llmasajudge}, model positions are randomized, and ties are modeled as half a win for each model. We estimate the coefficients using maximum likelihood estimation and report the resulting ratings on an Elo-style scale $r_\alpha = 400\beta_\alpha + 1200$.

\paragraph{Sampling diverse problems and requested techniques}
Both the diversity and adaptivity metrics require problems that can be solved using multiple techniques. To identify such problems, we collect several human solutions for each problem and cluster them by technique using our LLM-based clustering procedure. For diversity, we require each problem to have at least 3 clusters of human solutions, resulting in 119 problems that are used for this metric. For adaptivity, we first extract every problem-technique pair from the human solutions and then exclude pairs that were already naturally produced by \gptoss{} in its initial solution attempt. This ensures that the requested techniques are not trivial for LLMs. However, it requires care when interpreting the results of \gptoss{} on this metric, since \gptoss{} is disadvantaged by construction. The procedure yields 445 techniques across 237 problems.

\vspace{-0.5em}
\section{Experimental Evaluation}\label{sec:evaluation}

\begin{table*}[t!]
    \caption{Ratings across the different \bench metrics, measured on correct solutions, with their accuracy. The weak baselines at the bottom are rated against the frozen ratings of the other models.}
    \label{tab:main_bench}
    \centering
    \resizebox{\linewidth}{!}{
    \begin{tabular}{l c c c c c || c}
        \toprule
        & {\textbf{Conciseness}}
        & {\textbf{Computational Ease}}
        & {\textbf{Cognitive Simplicity}}
        & {\textbf{Diversity}}
        & {\textbf{Adaptivity}}
        & {\textbf{Accuracy}} \\
        \midrule
        \gptfivesix{} & 1944 & 1609 & 1256 & 1135 & \textbf{1619} & $\mathbf{93.4}\%$ \\
        \gptfivefour{} & 1295 & 1566 & \textbf{1338} & 1236 & 1484 & $86.3\%$ \\
        \dsvfourpro{} & \textbf{1973} & \textbf{1636} & 1306 & 1246 & 1431 & $72.0\%$ \\
        \geminiflashseven{} & 463 & 852 & \textit{1096} & 1227 & 1314 & $71.2\%$ \\
        \geminipro{} & \textit{165} & 1027 & 1169 & 1287 & 1264 & $60.7\%$ \\
        \kimi{} & 1463 & 1215 & 1098 & 1253 & 1067 & $53.5\%$ \\
        \glmfive{} & 850 & \textit{829} & 1182 & 1132 & 1139 & $49.1\%$ \\
        \dsvthreetwo{} & 1144 & 988 & 1266 & \textbf{1311} & 1049 & $47.9\%$ \\
        \gptoss{} & 1383 & 1064 & 1098 & 1179 & \textit{999} & $41.1\%$ \\
        \stepfun{} & 1345 & 1151 & 1145 & 1194 & 1168 & $34.3\%$ \\
        \grokfourfast{} & 1407 & 990 & 1184 & 1129 & 1031 & $31.4\%$ \\
        \geminiflash{} & 1117 & 1260 & 1247 & \textit{1094} & 1025 & $29.8\%$ \\
        \qwenthreebig{} & 1051 & 1414 & 1215 & 1175 & 1011 & $\mathit{29.3}\%$ \\
        \midrule
        \midrule
        \geminiflashlow{} & 1217 & 937 & 1196 & 1081 & 975 & $16.5\%$ \\
        \qwenthreeeight{} & 750 & 1209 & 1324 & 1067 & 939 & $8.1\%$ \\
        \bottomrule
    \end{tabular}
    \vspace{-1em}
    }
\end{table*}

\vspace{-0.5em}
We now present our evaluation on \bench{}. We first summarize model rankings in \cref{sec:results-main}. As our analysis uses default instructions, we further investigate how prompting can steer metrics (\cref{sec:steerability}). We then examine distributional differences between LLM and human solutions (\cref{sec:distributional-differences}), and show the necessity of correctness filtering in \cref{sec:results-ablation}. We validate the methodology and report further robustness checks in \cref{app:additional_validation,app:additional-experiments}, and provide a qualitative example-based analysis in \cref{app:qualitative-analysis}.

\vspace{-0.5em}
\paragraph{Evaluated models}
We evaluate a set of 13 frontier models: \gptfivesix{}~\citep{gpt56}, \gptfivefour{}~\citep{gpt54}, \gptoss{}~\citep{gptoss}, \geminipro{}~\citep{gemini31pro}, \geminiflashseven{}~\citep{gemini37flash}, \geminiflash{}~\citep{gemini3flash}, \qwenthreebig{}~\citep{qwen35}, \glmfive{}~\citep{glm5}, \kimi{}~\citep{kimik25}, \grokfourfast{}~\citep{grok41}, \dsvfourpro{}~\citep{deepseekv4}, \dsvthreetwo{}~\citep{deepseekv32}, and \stepfun{}~\citep{step35}. Additionally, we run \geminiflash{} (\texttt{low} reasoning) and \qwenthreeeight{}~\citep{qwen3} as weak baselines, scoring them against the frozen ratings of the other models, as their low accuracy leaves few solutions past the filter, which add noise to the rankings. Details on how we ran each LLM are provided in \cref{app:additional-experimental-details}.

\vspace{-0.5em}
\subsection{Key Takeaways}\label{sec:results-main}
\vspace{-0.5em}

We present the main results on \bench{} in \cref{tab:main_bench}. As only adaptivity has a universally preferred direction, we suggest reading \cref{tab:main_bench} by selecting the metrics relevant to a given use case, with bold and italics marking the two ends of every column, i.e., per the pointers provided in \cref{sec:methodology-metrics} and extended in \cref{app:interpretation}. Absolute scores for the metrics where such scores are measurable, namely conciseness, diversity, and adaptivity, are reported in \cref{app:raw-scores}, where we show that they correlate well, but not perfectly, with rankings. Confidence intervals of the main results are given in \cref{app:main-results-ci}. Below, we discuss the key takeaways. We describe results in terms of high or low scores, but do not imply that one is inherently better than the other, except for adaptivity, where higher is unambiguously better.

\vspace{-0.5em}
\paragraph{The most accurate models also rate highest}
The three most accurate models, \gptfivesix{}, \gptfivefour{}, and \dsvfourpro{}, take the highest rating on four of the five quality metrics, though the lead is split: \dsvfourpro{} writes the most concise and least computation-heavy proofs, \gptfivefour{} writes the most cognitively simple proofs, and \gptfivesix{} tops adaptivity. Other models show more mixed profiles. For example, \dsvthreetwo{} produces the most diverse approaches, while also being very computationally complex.

\paragraph{Stronger Gemini models are extremely verbose}
\geminipro{} clearly produces the least concise proofs, scoring around 1800 Elo points below \dsvfourpro{}, the least verbose model. This means that \dsvfourpro{}'s solutions are more concise than those by \geminipro{} in nearly $100\%$ of comparisons. In absolute terms, \geminipro{}'s solutions are on average $3.5$ times more verbose than necessary, which creates a significant burden for a human reader. \geminiflashseven{} follows closely with an Elo of under $500$. Qualitatively, both models frequently restate the goal, define steps in great detail, and end by restating the result before the final answer.

\vspace{-1mm}
\paragraph{\qwenthreeshort{} and top models have high computational ease}
\qwenthreeshort{} and the top 3 models score the highest in terms of computational ease. Manual inspection suggests that \qwenthreeshort{}'s high score is largely explained by its tendency to pursue more elegant ideas, for instance by avoiding brute-force , computation-heavy approaches to geometry problems, which are common among the other models. However, \qwenthreeshort{} has also the lowest accuracy among all evaluated models. In contrast, the top 3 models achieve a balance between correctness and computational ease, suggesting that they can find computationally simpler proofs without sacrificing accuracy.

\vspace{-1mm}
\paragraph{Accuracy and quality differ substantially}
Overall, the Spearman rank correlation between accuracy and the quality metrics is relatively low, with rankings on quality metrics differing substantially from rankings by accuracy. For instance, \glmfive{} scores low on all quality metrics despite achieving reasonably good accuracy. The highest correlation between accuracy and any quality metric is $0.89$ for adaptivity, which is expected since adaptivity requires correctness as a prerequisite. On the other hand, the correlations of all other metrics range between $0.2$ and $0.4$, suggesting that these metrics capture aspects of proof quality that are not aligned with correctness.

\vspace{-1mm}
\paragraph{Cognitive simplicity and diversity are most uniform}
Cognitive simplicity and diversity are the most uniform metrics across models, with all models scoring relatively close to one another. For instance, on cognitive simplicity, the range of scores is only 242 points, corresponding to a preference rate of only $65\%$ between the two extremes. In contrast, on conciseness, the range is over 1800 points, corresponding to a preference rate of $99\%$ between the two extremes.

\begin{figure*}[t]
\centering
\input{paper_files/figures/steering.tex}
\caption{Steerability of the \bench{} quality metrics. (a)--(c) show Elo rating changes, while (d) reports the average number of distinct methods per problem. Arrows start at the value under the standard prompt and end at the value under a directed prompt, with arrows marking intent. The grey band marks the spread of the standard-prompt values across models. Models are ordered by accuracy.}
\label{fig:steering}
\vspace{-1em}
\end{figure*}

\subsection{Steerability of Proof Quality} \label{sec:steerability}
\vspace{-0.5em}
Our main results measure every model under the same standardized prompt, which captures a uniform notion of behavior under a default neutral prompt. However, this leaves open how much this default fingerprint persists under specific instructions. We therefore rerun \bench{} using prompts that explicitly push a single metric in a chosen direction, which separates what a model does by default from what it can be asked to do.

\vspace{-1mm}
\paragraph{Evaluation methodology} For cognitive simplicity and computational ease we add instructions pushing towards either end of the metrics. For conciseness we use two prompts pushing towards brevity, as proofs can be trivially extended via redundancy. As diversity is defined over independent samples rather than a single proof, we instead ask for several distinct methods within a single call. We exclude adaptivity, which is prompt-driven by construction. All prompts are included in \cref{app:steering_prompts}.

Each prompted variant is scored against the frozen ratings of the original models. \Cref{fig:steering} reports the resulting rating changes. \Cref{tab:steering} summarizes each setting's effect on ratings, accuracy, and ranking preservation, via the Spearman correlation $\rho$ and the number of rank inversions (Inv.). We exclude \grokfourfast{}, deprecated during our research, and \stepfun{}, whose request error rate under the steered prompts exceeds 70\%.

\begin{wraptable}{r}{0.6\linewidth}
    \caption{Aggregate effect of each directed prompt, over the 11 models.}
    \label{tab:steering}
    \centering
    \resizebox{\linewidth}{!}{
    \begin{tabular}{l l r r r r r}
        \toprule
        & \textbf{Prompt} & \textbf{$\Delta$Elo}
        & \multicolumn{2}{c}{\textbf{$\Delta$Acc ($\%$)}}
        & \textbf{Inv.} & \textbf{$\rho$} \\
        \cmidrule(lr){4-5}
        & & median & median & worst & & \\
        \midrule
        \multirow{2}{*}{Cognitive}
          & easier  & $+66$   & $-1.8$  & $-14.7$ & $7$  & $0.88$ \\
          & harder  & $-262$  & $-1.1$  & $-11.8$ & $16$ & $0.57$ \\
        \midrule
        \multirow{2}{*}{Computation}
          & lighter & $+250$  & $-2.3$  & $-13.4$ & $12$ & $0.79$ \\
          & heavier & $-181$  & $+0.5$  & $-13.4$ & $5$  & $0.94$ \\
        \midrule
        \multirow{2}{*}{Conciseness}
          & mild    & $+634$  & $-5.5$  & $-15.6$ & $9$  & $0.85$ \\
          & strong  & $+1114$ & $-10.9$ & $-31.7$ & $10$ & $0.82$ \\
        \bottomrule
    \end{tabular}
    }
\end{wraptable}

\vspace{-1mm}
\paragraph{Every metric is steerable} All metrics respond in the requested direction for nearly every model. The changes are largest for conciseness, with a median gain of over 1,000 Elo in the stronger setting, while asking for several methods at once raises the average number of distinct methods per problem from $1.78$ to $2.39$. Across all model--setting pairs, only three shifts are statistically indistinguishable from zero: \glmfive{} under the easier-proof prompt ($+19$), and \geminiflash{} ($-46$) and \glmfive{} ($+31$) under heavier computation, the last being the only occasion a model moves against its instruction.

\vspace{-1mm}
\paragraph{Rankings are largely preserved under prompting} Under a shared directed prompt, the ranking of models stays close to the standard-prompt ranking, with $\rho \in [0.79, 0.94]$ for five of the six directed prompts and $5$--$12/55$ model pairs inverted. Where inversions do occur, they concentrate among models that were close to begin with: $75\%$ happen at a base gap of $\leq 200$, against $46\%$ of model pairs being that close, as we show in \cref{tab:steering_inversion_gap}. The one substantial reordering is the harder-proof prompt, at $\rho = 0.57$ with $16/55$ pairs inverted. This is expected, since cognitive simplicity is the metric with the least ordering to preserve, given its narrow spread. Overall, while prompting shifts the absolute level of metrics, each model's behavioral fingerprint is largely retained. The exception is diversity, where the single-call approach reorders models substantially ($\rho = 0.07$). However, it closely matches adaptivity ($\rho = 0.90$), as it then measures producing alternative approaches on demand (\cref{app:steering-diversity}).

\vspace{-1mm}
\paragraph{Steering can reduce utility} Prompts that ask only for a different style, rather than a different task, can still cost accuracy. As shown in \cref{tab:steering}, this is worst for conciseness, where shortening leads models to skip critical, non-trivial steps or to assert them without proof, leaving the proof incomplete; the median drop doubles from $5.5$ to $10.9$ points between the mild and the strong prompt. The mild prompt is thus the better trade-off in practice, reaching $57\%$ of the strong prompt's median rating gain at half its accuracy cost. The other proof-level settings are milder, with median changes between $-2.3$ and $+0.5$ points, while asking for several methods at once costs a comparable median of $11.0$ points. The drop in accuracy is also model dependent: every prompt costs its worst-affected model more than $10$ points, while others are barely affected (\cref{tab:steering_acc}).

\begin{table*}[t]
    \caption{Model ratings across the different \bench metrics on all solutions.}
    \label{tab:all_bench}
    \centering
    \newcommand{\ratingdelta}[2]{%
        \makebox[6.0em][c]{%
            \makebox[2.2em][r]{#1}%
            \,\makebox[3.3em][l]{#2}%
        }%
    }
    \resizebox{\linewidth}{!}{
        \begin{tabular}{l c c c c}
            \toprule
            & \textbf{Conciseness}
            & \textbf{Computational Ease}
            & \textbf{Cognitive Simplicity}
            & \textbf{Diversity} \\
            \midrule
            \gptfivesix{} & \ratingdelta{1569}{(-375)} & \ratingdelta{1233}{(-376)} & \ratingdelta{1066}{(-190)} & \ratingdelta{\textit{1002}}{(-133)} \\
            \gptfivefour{} & \ratingdelta{1159}{(-136)} & \ratingdelta{1311}{(-255)} & \ratingdelta{1172}{(-166)} & \ratingdelta{1095}{(-141)} \\
            \dsvfourpro{} & \ratingdelta{\textbf{1662}}{(-311)} & \ratingdelta{1335}{(-301)} & \ratingdelta{1167}{(-139)} & \ratingdelta{1223}{(-23)} \\
            \geminiflashseven{} & \ratingdelta{526}{(+63)} & \ratingdelta{\textit{763}}{(-89)} & \ratingdelta{\textit{986}}{(-110)} & \ratingdelta{1089}{(-138)} \\
            \geminipro{} & \ratingdelta{\textit{454}}{(+289)} & \ratingdelta{1001}{(-26)} & \ratingdelta{1069}{(-100)} & \ratingdelta{1168}{(-119)} \\
            \kimi{} & \ratingdelta{1321}{(-142)} & \ratingdelta{1179}{(-36)} & \ratingdelta{1094}{(-4)} & \ratingdelta{1286}{(+33)} \\
            \glmfive{} & \ratingdelta{817}{(-33)} & \ratingdelta{980}{(+151)} & \ratingdelta{1273}{(+91)} & \ratingdelta{1238}{(+106)} \\
            \dsvthreetwo{} & \ratingdelta{1154}{(+10)} & \ratingdelta{1153}{(+165)} & \ratingdelta{1343}{(+77)} & \ratingdelta{1249}{(-62)} \\
            \gptoss{} & \ratingdelta{1226}{(-157)} & \ratingdelta{1083}{(+19)} & \ratingdelta{1099}{(+1)} & \ratingdelta{1301}{(+122)} \\
            \stepfun{} & \ratingdelta{1648}{(+303)} & \ratingdelta{\textbf{1477}}{(+326)} & \ratingdelta{1287}{(+142)} & \ratingdelta{\textbf{1397}}{(+203)} \\
            \grokfourfast{} & \ratingdelta{1601}{(+194)} & \ratingdelta{1305}{(+315)} & \ratingdelta{1362}{(+178)} & \ratingdelta{1193}{(+64)} \\
            \geminiflash{} & \ratingdelta{1329}{(+212)} & \ratingdelta{1306}{(+46)} & \ratingdelta{1298}{(+51)} & \ratingdelta{1101}{(+7)} \\
            \qwenthreebig{} & \ratingdelta{1132}{(+81)} & \ratingdelta{1473}{(+59)} & \ratingdelta{\textbf{1385}}{(+170)} & \ratingdelta{1258}{(+83)} \\
            \bottomrule
        \end{tabular}
    }
    \vspace{-1em}
\end{table*}

\subsection{Distributional Differences With Human Solutions} \label{sec:distributional-differences}
\vspace{-0.75em}

Beyond model rankings, the clustering pipeline lets us compare LLM-written and human-written solution distributions. We measure the fraction of correct LLM solutions that are not assigned to any human-solution cluster. On average, $24\%$ of correct LLM solutions fall outside the human clusters ($14\%$ to $32\%$ across models), while $31\%$ of human solutions are not matched to any LLM one.

These differences vary by mathematical domain. Many unmatched LLM solutions occur in number theory and geometry. For number theory, models frequently use obscure techniques from the literature that are not present in our collected human-solution set, while for geometry, models often use alternative computational approaches such as complex numbers, trigonometry, or Cartesian coordinates. For example, \geminiflash{} produces unmatched number theory solutions $45.0\%$ of the time, while \gptfivefour{} has a $47.3\%$ unmatched rate in geometry. Importantly, these results should not be interpreted as evidence that LLMs produce new mathematics, since our human solution set is not exhaustive. Rather, they show that LLM-generated solutions often differ from the most common human-written solutions and can reflect different parts of the method distribution.

\vspace{-0.5em}
\subsection{Analysis Without Correctness Prerequisite}\label{sec:results-ablation}
\vspace{-0.75em}

All preceding analyses compare only correct solutions. To test this design choice, we rerun \bench{} without filtering for correctness and include all solutions. We exclude adaptivity, where correctness is required by definition.

\paragraph{Results} As shown in \cref{tab:all_bench}, incorrect solutions have substantial and varying effects across models. The scores of \gptfivesix{} are significantly reduced across all metrics, as its correct solutions are now compared against many incorrect, and potentially very simple, solutions from other models. In contrast, the scores of \stepfun{} improve dramatically. Qualitatively, we find that its incorrect solutions often provide very high-level but incorrect arguments, leading to strong scores across metrics. The pattern is systematic: the five most accurate models lose rating in $18$ of their $20$ model--metric pairs, while the remaining eight gain in $26$ of $32$. Similar trends arise when filtering for a correct answer but allowing incomplete justifications, as we discuss in \cref{app:bench-correct-answers}. This indicates that filtering for correctness is essential when evaluating proof quality. Otherwise, the results are dominated by the simplicity of incorrect solutions.

\vspace{-1mm}
\section{Conclusion} \label{sec:conclusion}
\vspace{-2mm}
In this work, we introduced \bench{}, a benchmark for assessing the quality of LLM-generated mathematical proofs beyond correctness. We defined a set of metrics that scalably measure proxies for proof quality, including conciseness, computational ease, cognitive simplicity, diversity, and technique adaptivity. Our evaluation reveals distinct proof-style profiles among frontier models, as well as nontrivial tradeoffs between accuracy and other proof-quality metrics. We further find that every proof-level metric can be steered by explicit instruction, at times at a cost in accuracy, but the underlying models' relative ranking remains largely consistent. This perspective can help users select models whose proof-writing behavior better matches their mathematical goals, can support future work on steering models toward user-specific proof styles, and lays the groundwork for user studies of how LLM proof quality is perceived by different audiences. We discuss the limitations of our benchmark, as well as the directions it opens up, in \cref{app:limitations-future}.

\section*{Acknowledgements}

This research was partially funded by the Ministry of Education and Science of Bulgaria (support for INSAIT, part of the Bulgarian National Roadmap for Research Infrastructure). This project was supported with computational resources provided by Google Cloud Platform (GCP).

We thank Nikolay Drenchev for contributing the independent annotations reported in \cref{app:human-validation}.

\bibliography{references}
\bibliographystyle{unsrtnat}

%%%%%%%%%%%%%%%%%%%%%%%%%%%%%%%%%%%%%%%%%%%%%%%%%%%%%%%%%%%%%%%%%%%%%%%%%%%%%%%
%%%%%%%%%%%%%%%%%%%%%%%%%%%%%%%%%%%%%%%%%%%%%%%%%%%%%%%%%%%%%%%%%%%%%%%%%%%%%%%
% APPENDIX
%%%%%%%%%%%%%%%%%%%%%%%%%%%%%%%%%%%%%%%%%%%%%%%%%%%%%%%%%%%%%%%%%%%%%%%%%%%%%%%
%%%%%%%%%%%%%%%%%%%%%%%%%%%%%%%%%%%%%%%%%%%%%%%%%%%%%%%%%%%%%%%%%%%%%%%%%%%%%%%
\clearpage
\appendix
% cleveref's \appendix detection breaks with recent hyperref; set the types explicitly
\crefalias{section}{appendix}
\crefalias{subsection}{subappendix}
\crefalias{subsubsection}{subsubappendix}
\onecolumn
\section{Limitations and Future Work} \label{app:limitations-future}

\subsection{Limitations} \label{sec:limitations}

We acknowledge several limitations of our benchmark. First, we do not directly measure proof quality, since there is no objective notion of proof quality that can be easily quantified. Human evaluation is possible, but it is expensive, time-consuming, and may be inconsistent across evaluators. We include a small human evaluation in \cref{app:human-validation}, covering both annotators from within and outside the author team, to validate our LLM-as-a-judge components, but we rely primarily on scalable proxies that can be evaluated automatically. The independent portion of that study is small, at $22$ pairs per dimension from a single annotator, so the agreement it reports carries substantial sampling error and should be read as a check that author agreement is not inflated rather than as a precise estimate.

Second, LLM judges are imperfect and may not always align with human judgments of proof quality. They may also exhibit biases or become less reliable as future models develop new proof-writing styles. This is difficult to avoid in scalable evaluations: any automatic proof-quality metric must rely on a heuristic or proxy. We mitigate this issue as much as possible through extensive validation experiments (see \cref{app:additional-experiments}), which we use to iteratively refine our evaluation components and prompts to make them as reliable and robust as possible. Through these iterations, we also rejected several candidate evaluation methodologies, which we discuss in \cref{app:challenges}.

Third, all ratings are measured under a single standardized prompt and therefore describe default model behavior rather than the full range of behavior a model can be steered into. \Cref{sec:steerability} quantifies how far each metric moves under directed prompting, but the ratings we report should not be read as an inability of a model to behave differently when asked.

Finally, our benchmark focuses on final-answer competition problems, which are a subset of mathematical problems. This allows us to evaluate correctness more easily, but it may limit the generality of our findings to other types of problems, such as proof-writing tasks that require more open-ended reasoning or research-level problems.

\subsection{Future Work} \label{app:future-work}

A direct consequence of our final limitation is to extend this work to research-level problems, to which all of our approaches can be applied directly. However, additional challenges may arise in validating our components at that level. This includes how well the LLM judges align with expert human judgment, and whether our automated validation findings transfer to the new domains. Furthermore, it remains unclear whether research-level proofs can be verified at scale without significant costs.

A more human-centered extension of this work is to measure how proof quality is perceived by separate audiences the benchmark is meant to serve. Our metrics are defined in terms that are deliberately reader-independent, but what counts as cognitively simple, or as an appropriate level of detail, plainly depends on who is reading. We see our contribution as the groundwork that makes such a study possible, in that it provides metrics with explicit definitions, a documented measurement procedure, and controls for confounders such as problem difficulty.

A natural design would run comparisons separately within several audiences, from school students and undergraduates to instructors, competition coaches, and research mathematicians, and then compare the resulting ratings across groups. Covariates such as level of study and mathematical tradition would have to be recorded alongside the judgments, since they shape which notation is considered familiar and how much detail a proof is expected to include.

\section{Impact Statement} \label{app:impact}

\bench{} can support better evaluation of LLMs used for mathematical reasoning by broadening the scope of evaluation beyond correctness to include other important proof-quality dimensions such as conciseness and computational ease. This is useful in almost all mathematical applications, since good proofs should help users understand the underlying mathematical ideas and build intuition. By providing a more nuanced view of model performance, \bench{} can help guide users in selecting models that are accurate and produce high-quality proofs.

\section{Interpreting \bench{} Metrics} \label{app:interpretation}

In this section, we expand on the discussion in \cref{sec:methodology-metrics} and describe in more detail how we suggest the \bench{} metrics be interpreted by practitioners.

We see two takeaways from our results. The first is that models diverge substantially along dimensions that correctness-only evaluation cannot see, and that these dimensions are measurable and reproducible. The second, and more actionable one, is that because most of our dimensions have no universally preferred direction, \bench{} is best read as a tool for matching a model to a use case, rather than for declaring a single winner. Below, we instantiate this advice for each dimension.

\paragraph{Conciseness} Verbosity is not inherently negative. An expert reader may prefer the most compressed argument, since they can supply the missing steps themselves, whereas a novice may benefit from a model that spells out more details, filling gaps in their knowledge and lowering the cognitive load. Furthermore, when used for autoformalization efforts or for confirming a new result, having as many details as possible is valuable, while also coming at a cost for the reader to parse and verify the proof. A low conciseness rating is therefore a warning sign primarily when reading time is the scarce resource, and a feature when every step has to be independently checkable.

\paragraph{Computational ease} Computation-heavy proofs can be harder to read and to check notationally, but they are typically more rigorous and less likely to omit small yet important steps. Models on the computation-heavy end of this metric are thus well suited to settings where a mechanical but complete argument is preferable to a slick one, while models on the other end are preferable when the goal is to communicate the idea behind the proof.

\paragraph{Cognitive simplicity} Cognitively simpler solutions take less time to verify, which matters when reviewer or reader time is the bottleneck. They are also more approachable for a broader audience, but may, however, present less novel or interesting approaches. Models that score low on this metric may still be the right choice for a reader who is specifically looking for less standard ideas.

\paragraph{Diversity} A more diverse model explores more ideas, which is valuable for research use and for generating alternative solutions, e.g., when preparing teaching material that presents a problem from several angles. A less diverse model is more consistent, and can therefore be more reproducible across users, which is desirable when the same query is expected to yield the same answer.

\paragraph{Adaptivity} This is the one dimension where higher is unambiguously better, and the most directly actionable. It matters most in education, where a solution must be reachable from the student's current knowledge, e.g., one would not expect a sixth-grader to follow a geometry proof via complex numbers, and in exam preparation, where a specific technique is being tested and the model must be able to produce a proof using it. Since adaptivity requires correctness as a prerequisite and is measured under a technique-constrained prompt, it is also the metric most closely aligned with accuracy.

\paragraph{Default behavior and prompt malleability} All ratings reported in \bench{} are measured under a single standardized prompt, and therefore describe a model's default behavior rather than the full range of behavior it can be steered into. This is the relevant notion for the many downstream settings in which the user does not specify a proof style, but it does mean that a low rating on a metric should not be read as an inability of the model to behave differently when asked. Our steerability analysis in \cref{sec:steerability} makes this precise: prompts that explicitly ask for the behavior a metric measures do move the metric in the intended direction for essentially every model, sometimes at a noticeable cost in accuracy. At the same time, when all models are steered in the same direction, their relative ordering is largely preserved, so the guidance above remains applicable to prompted usage as well. The main exception is diversity, where explicitly asking for several distinct approaches reorders models strongly, so default diversity predicts prompted diversity less well than for other metrics.

\section{Additional Experimental Details} \label{app:additional-experimental-details}

We provide additional experimental details that are not included in the main text to ensure full transparency and reproducibility.

\subsection{Model Details and Inference} \label{app:model-inference-details}

\paragraph{Hyperparameters} All models were run using the hyperparameters recommended by their respective providers. Exact per-model configurations are provided in our code. Unless otherwise specified, models were run with their maximum available reasoning budget. We retried a generation only when the corresponding request exceeded a timeout of 150 minutes. For \gptfivefour{} and \gptfivesix{}, we used the \texttt{background=True} parameter, which was needed to avoid connection failures on long requests. All evaluated models are run with reasoning enabled, including those for which reasoning is optional, such as \dsvthreetwo{}, so that each model is measured in its strongest configuration. Reasoning traces are themselves never scored, and all our metrics are computed on the final response alone, since this is what a user actually sees and the only representation available for closed models. 

\paragraph{Parsing and Format Adherence} For prompts requiring a structured output format, such as diversity clustering or core-idea generation, we implemented task-specific parsers to extract the required fields from model responses. These parsers were also used to identify formatting errors and internal inconsistencies in the outputs. When a response did not conform to the expected format, we retried the prompt up to 10 times. If all retry attempts failed, the corresponding instance was marked as a formatting failure and excluded from the affected dataset. In practice, such failures were rare, occurring in less than $0.5\%$ of runs in each experiment.

\subsection{Computational Requirements}

For larger and closed-source LLMs, we used the providers' API services, as well as the OpenRouter API. To improve throughput, we issued 32 to 128 concurrent API requests, depending on provider rate limits and model availability. Our budget for API inference was roughly \$5,000, most of which was spent on the steerability experiments, where solvers were rerun on the full benchmark. 

All local inference experiments, including those using \gptoss{} and \qwenthreebig{}, were run on a single node with 8 NVIDIA H200 GPUs, each with 144 GB of VRAM. These runs used at most 144 GB of system RAM and 32 CPU cores. In practice, most experiments could be run on 1 to 4 H200 GPUs, although with reduced throughput. In total, local LLM inference took roughly 120 GPU hours, including solver inference and LLM judging.
\section{Validation of Methodology and Metrics} \label{app:additional_validation}

Because our evaluation targets subjective and nuanced aspects of mathematical proof generation, we perform additional validation experiments to assess the reliability of our methodology and the robustness of our metrics. In particular, we focus on components that use an LLM-as-a-judge: rephrasing-based conciseness measurement, pairwise preference judgment, clustering consistency, and correctness and technique verification.

\subsection{Effect of Rephraser on Conciseness Measurement} \label{app:rephraser-verbosity}

Since our conciseness metric relies on an LLM rephraser to measure the compressibility of a solution, we test whether these measurements are robust to the choice of rephraser. Here, we compare our main verifier, \gptoss{}, with \grokfourfast{}. 

\begin{figure*}[t]
    \centering
    \includegraphics[width=\textwidth]{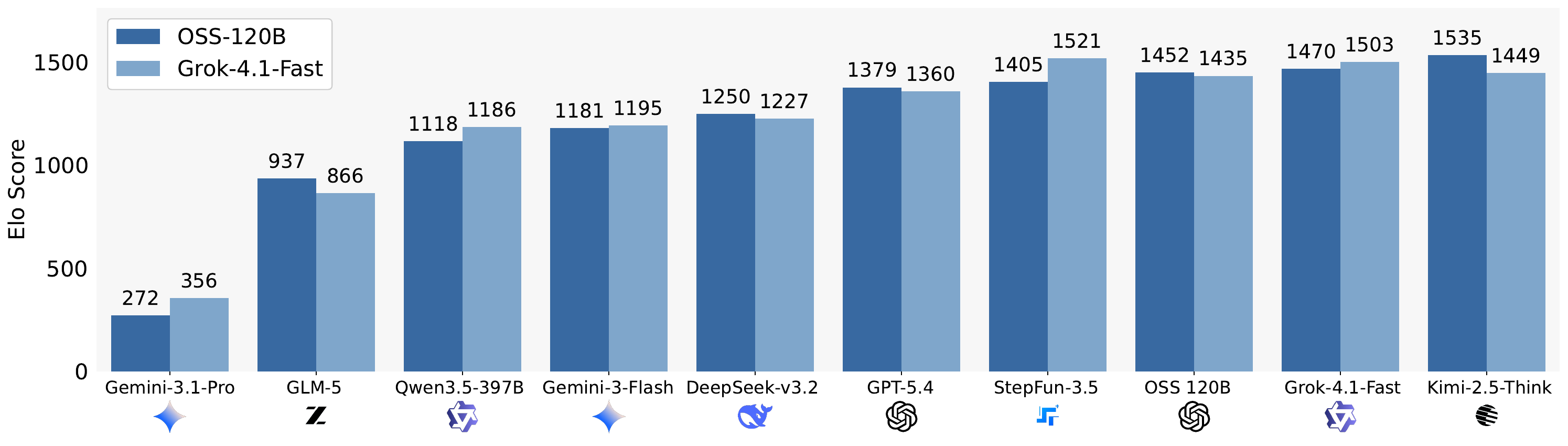}
    \caption{Conciseness Elo changes when \grokfourfast{} replaces \gptoss{} as the rephraser.}
    \label{fig:rephraser_elo}
\end{figure*}

We find that \grokfourfast{} tends to produce shorter rephrasings, and carries over this capability to this task by compressing each solution more aggressively. However, the relative comparisons between models remain consistent. As shown in \cref{fig:rephraser_elo}, the average shift in Elo ratings is only $4.1\%$. Although some scores change, the overall ranking is preserved, with few rank inversions.

We use \gptoss{} in our main experiments because it is a more reliable rephraser, producing semantically valid rephrasings in $94.1\%$ of cases. In contrast, \grokfourfast{} achieves only $88.1\%$. We describe this experiment in more detail in \cref{app:rephrasing_validity}.

\subsection{Human-driven Validation} \label{app:human-validation}

To assess the reliability of our methodology, we asked three authors with sufficient mathematical background to compare pairs of solutions along three primary metrics: \emph{conciseness}, \emph{computational ease}, and \emph{cognitive simplicity}. We also considered two additional qualitative dimensions: \emph{insightfulness} and \emph{elegance}. Insightfulness measures whether a solution introduces transferable ideas or techniques, while elegance captures whether the proof uses a simple, creative, or unexpectedly efficient argument. To check that our validation is not confined to the author team, we additionally recruited an external annotator with IMO-level experience who was not involved in the development of this work. The instructions given to all annotators are provided in \cref{app:human-validation-instructions}.

We collected $112$ pairwise judgments for each dimension. $40$ of these judgments were graded twice by annotators from the author team, and a further $22$ pairs per dimension were independently annotated by the external annotator. To summarize inter-annotator consistency, we use a graded agreement score: $100\%$ for identical preferences, $50\%$ when one annotator expressed no preference and the other preferred one solution, and $0\%$ for opposing preferences. We report agreement among the author annotators, between the external and author annotators, and pooled over all doubly annotated comparisons in \cref{tab:human_agreement}. Overall, the results show good agreement on 4 of the 5 dimensions.

\begin{table}[t]
    \centering
    \caption{Average inter-annotator agreement from our human annotation study, reported among the author annotators, between the external annotator and the author annotators, and pooled over all doubly annotated comparisons.}
    \label{tab:human_agreement}
    \centering
    \resizebox{\linewidth}{!}{
    \begin{tabular}{lccccc}
        \toprule
        \textbf{Comparison} & \textbf{Conciseness} & \textbf{Computational Ease} & \textbf{Cognitive Simplicity} & \textbf{Insightfulness} & \textbf{Elegance} \\
        \midrule
        Author to author & $57.6\%$ & $83.9\%$ & $80.1\%$ & $87.0\%$ & $89.0\%$ \\
        External to author & $59.1\%$ & $86.4\%$ & $81.8\%$ & $81.8\%$ & $72.7\%$ \\
        \midrule
        Overall & $58.2\%$ & $84.8\%$ & $80.7\%$ & $85.1\%$ & $83.0\%$ \\
        \bottomrule
    \end{tabular}
    }
\end{table}

\paragraph{Inter-annotator agreement} Agreement exceeds $80\%$ on every dimension except conciseness, and is highest for \emph{insightfulness} ($85.1\%$) and \emph{computational ease} ($84.8\%$). Although insightfulness and elegance are highly subjective, the high level of agreement suggests that annotators often share similar preferences, at least within the context of our examples. Agreement above $80\%$ for \emph{computational ease} and \emph{cognitive simplicity} further suggests that these criteria can be applied consistently in pairwise comparisons. We illustrate both regimes in \cref{app:examples-clearcut,app:examples-contentious}, which demonstrate pairs on which the annotators reached the same decisive verdicts and pairs on which they split.

The main exception is \emph{conciseness}, where agreement remains below $60\%$ both among the authors and against the external annotator. Annotators noted that different solutions were often verbose in different ways, so judgments depended on which redundant parts they emphasized. After annotation, the annotators agreed that they had often tried to approximate our compressibility metric. However, doing so reliably was difficult without a systematic method for identifying which parts of the solution were redundant. This observation motivates our automatic conciseness metric: instead of relying on local perceptions of verbosity, it uses an operational definition based on how much text can be removed while preserving the mathematical argument. We work through a pair on which the two annotators split in exactly this way in \cref{app:example-conciseness}.

\paragraph{External annotator} Since all three author annotators know each other and discussed this project together, their judgments may be more aligned than those of independent annotators. The external annotations let us quantify this effect. Agreement is stable on the three dimensions that \bench{} reports: \emph{conciseness}, \emph{computational ease}, and \emph{cognitive simplicity} are all marginally higher against the external annotator than among the authors. The drop is instead confined to \emph{insightfulness} and \emph{elegance}, the two qualitative dimensions that we collect for analysis but that do not feed into any benchmark rating, and may be attributed to the low sample size. Thus, the alignment we would expect from a closely collaborating annotator pool does not significantly impact agreement on the dimensions we use for model comparison.

\paragraph{Metric correlation} We further measure the Pearson correlations between the different dimensions. Overall, we find that computational ease is moderately correlated with elegance ($r=0.51$), and cognitive simplicity is weakly correlated with insight ($r=0.28$). Conciseness, likely due to its low agreement, is essentially orthogonal to the other dimensions.

\paragraph{Validation on computational ease and cognitive simplicity}

Using the $90$ judgments collected from the author annotators, we compare \gptoss{} pairwise preferences against human preferences. \gptoss{} agrees with human judgments at rates of $85.3\%$ for \emph{cognitive simplicity} and $82.1\%$ for \emph{computational ease}. These agreement rates are comparable to inter-annotator agreement. This provides evidence that the LLM judge applies these criteria consistently. One possible explanation for the higher scores is that the doubly annotated subset contains many pairs of similar solutions, where fine-grained distinctions may be more difficult for humans than for automated systems. \Cref{app:example-judge} presents one such pair, on which both annotators record a tie while the judge still returns a preference.

\subsection{Clustering Consistency Validation} \label{app:diversity-consistency}

Our diversity and adaptivity metrics use an LLM-based clustering procedure to group solutions by their underlying mathematical approach. We validate consistency as multiple valid clusterings may exist, as whether two proofs count as ``the same approach'' depends on the granularity at which one reads them, and such similarity comparisons can be subjective~\citep{sabermath}. Because the procedure may depend on the set of solutions shown to the model and on the representation of each solution, we evaluate its consistency under subsampling and compare several input representations.

In the main experiments, the clustering model receives short summaries of all solutions. We compare this design with two alternatives:
\begin{itemize}
    \item \textbf{Ours}: short summaries of all solutions.
    \item \textbf{Full}: full solution texts.
    \item \textbf{Fingerprint}: extracted keywords describing the backbone techniques and proof strategies used in each solution.
\end{itemize}

To measure consistency, we first cluster the full set of solutions for a problem. We sample up to five subsets per problem, choosing each subset size uniformly at random. For each subset, we restrict the full-set clustering to the same solutions and compute the Rand Index between the restricted full-set clustering and the subset clustering. We average the Rand Index across subsets and problems. Results are in \cref{tab:clustering_consistency}, separately for human-only solutions and the combined human + LLM pool.

\begin{table}[t]
    \centering
    \caption{Mean Rand Index between clustering on the full set of solutions and on smaller subsets. We measure this for human-only solutions and the combined human + LLM solution pool.}
    \label{tab:clustering_consistency}
    \begin{tabular}{lcc}
        \toprule
        {\textbf{Method}} & {\textbf{Human}} & {\textbf{Human + LLM}}\\
        \midrule
        \textbf{Ours} & $87.8\%$ & $88.6\%$ \\
        \textbf{Full} & $82.3\%$ & $79.4\%$\\
        \textbf{Fingerprint} & $79.1\%$ & $80.0\%$ \\
        \bottomrule
    \end{tabular}
    \vspace{4mm}
\end{table}

The short-summary representation is the most consistent across both settings. It achieves a mean Rand Index of $87.8\%$ on human-only solutions and $88.6\%$ on the combined human + LLM pool. The full-solution representation is less stable, especially in the larger combined pool, likely because long solutions introduce irrelevant details that distract from the core technique. This is particularly problematic for larger solution sets, where the context size may overwhelm the model. Conversely, the fingerprint representation appears too sparse: a small number of keywords often fails to capture the substantive relationship between two approaches, leading to inconsistent cluster assignments. Consistency alone could in principle be achieved by a degenerate clustering, such as one that places every solution in a single group. We ruled those out manually by inspecting the produced clusters, which we find to be sensible and mathematically meaningful separations of approach. These results motivate our choice of short summaries as a balance between informativeness and robustness.

\subsection{Correctness Verifier Ablation} \label{app:correctness-verifier}

After validating quality judgments and clustering, we validate the correctness filter from \cref{sec:methodology-dataset} which consists of an answer checker plus a completeness verifier. The first checks the final answer and the second checks whether the reasoning justifies it. To test both components, we measure agreement with \gptfivefour{}, a strong standalone judge, using the full verification prompt in \cref{app:solution-verification-prompt}.

\begin{table}[t]
    \centering
    \caption{Agreement with \gptfivefour{} reference judgments for correctness-verification components.}
    \label{tab:correctness_ablation}
    \begin{tabular}{lc}
        \toprule
        {\textbf{Method}} & {\textbf{Agreement}}\\
        \midrule
        Answer + Completeness & $91.5\%$ \\
        Completeness & $88.0\%$ \\
        Answer & $73.0\%$ \\
        \bottomrule
    \end{tabular}
\end{table}

The results in \cref{tab:correctness_ablation} show that the combination of the two components is significantly better aligned than either component in isolation. These results are computed before manually resolving discrepancies between the verifier outputs and the LLM-based validation. 

In the 17 samples where the verifier and \gptfivefour{} disagreed, we found through manual inspection that our verifier was correct in 4 cases, while \gptfivefour{} was correct in the remaining 13. These errors are classified into 10 false positives and 3 false negatives, which fall into the following categories:
\begin{itemize}
    \item In \textbf{10 solutions}, the completeness verifier was too permissive and accepted solutions that omitted an important reasoning step. In most of these cases, however, the solution followed a technique that could plausibly lead to a correct proof if completed.
    \item In \textbf{2 solutions}, the completeness verifier was too strict and rejected arguments that reused an analogous reasoning pattern multiple times. Although this is a valid proof strategy, the verifier expected a more explicit statement of the analogy.
    \item In \textbf{1 solution}, a geometry problem from \answerbench{} contained a confusing rephrasing. The solution correctly identified that one point had not been translated correctly from the original statement. However, because recovering the intended statement required an additional assumption, the completeness verifier rejected the solution.
\end{itemize}

Overall, these errors affect only a small fraction of the evaluated samples. Moreover, many false positives correspond to incomplete but directionally correct solution strategies. We therefore expect the verifier errors to have limited impact on the main results.

Given this level of agreement, we keep \gptoss{} as our verification backbone rather than using \gptfivefour{} directly, primarily for cost reasons, as the verifications required by the full benchmark would cost roughly \$$4{,}800$, and over \$$12{,}000$ including all ablations, whereas \gptoss{} runs on compute already available to us and completes the same batch in under an hour. We retain \gptfivefour{} as the reference judge in this experiment precisely so that the reliability of the cheaper verifier can be quantified rather than assumed.

\subsection{Technique Verification Ablation} \label{app:technique-verifier}

The adaptivity metric relies on a second judgment on top of correctness, namely whether a solution actually follows the requested technique. We validate this component in the same way as the correctness filter. On $100$ randomly sampled problem-technique pairs, which were already filtered for correctness, we compare the \gptoss{} technique verifier against \gptfivefour{} using the prompt in \cref{app:technique_verification}, and manually resolve every discrepancy.

The two models agree on $86\%$ of samples. Of the $14$ disagreements, manual inspection resolves $5$ in favor of \gptoss{}, for an estimated accuracy of $91\%$. The discrepancies were as follows:
\begin{itemize}
    \item In \textbf{9 solutions}, \gptoss{} was given a correct solution but misjudged the relevance of the technique: the requested technique was attempted, led to a dead end, and the proof was obtained by another approach, or it was merely a minor step in the end solution.
    \item In \textbf{5 solutions}, \gptfivefour{} was overly pedantic about what counts as an application of the technique, either because the technique was not referenced by name or because the technique description was itself somewhat vague.
\end{itemize}

As with correctness verification, the cheaper verifier is thus reliable enough for this metric.

\subsection{Robustness to the Correctness Filter} \label{app:highly-solved}

As noted in \cref{sec:methodology-dataset}, the correctness filter rates each model on the problems it solves, so weaker models are rated on a smaller, generally easier subset. To check that this does not drive our results, we repeat the evaluation on problems nearly all models solve, where the per-model subsets almost coincide.

Concretely, we restrict the evaluation to problems solved by at least $9$ of the $13$ evaluated models, i.e., two thirds of the field. Of the $382$ problems in \bench, $131$ meet this criterion. We then refit the Bradley-Terry model of \cref{sec:methodology-dataset} for each metric on the restricted set and compare the resulting ranking against the corresponding ranking of \cref{tab:main_bench}, using the Spearman rank correlation $\rho$.

The rankings are essentially preserved, as all five metrics attain $\rho \geq 0.95$ (\cref{tab:highly_solved}). We conclude that the difficulty shift from the correctness filter does not meaningfully alter the model ordering.

\begin{table}[t]
    \centering
    \caption{Agreement between the main rankings of \cref{tab:main_bench} and the rankings obtained when the evaluation is restricted to problems solved by at least $9$ of the $13$ evaluated models. Problem counts include only problems contributing at least one pairwise comparison.}
    \label{tab:highly_solved}
    \resizebox{\linewidth}{!}{
    \begin{tabular}{lccccc}
        \toprule
        & \textbf{Conciseness} & \textbf{Computational Ease} & \textbf{Cognitive Simplicity} & \textbf{Diversity} & \textbf{Adaptivity} \\
        \midrule
        Problems & $382 \rightarrow 131$ & $382 \rightarrow 131$ & $382 \rightarrow 131$ & $115 \rightarrow 67$ & $445 \rightarrow 77$ \\
        $\rho$ & $0.995$ & $0.989$ & $0.956$ & $0.995$ & $0.984$ \\
        \bottomrule
    \end{tabular}
    }
\end{table}

\subsection{Comparison with Cheaper Alternatives} \label{app:cheaper-baselines}

Several of our components admit simpler and cheaper implementations, and it is natural to ask whether the additional cost we incur buys anything. We therefore compare each component against the most obvious cheaper alternative, measured on the same data used to validate the component.

For conciseness, the cheapest alternative is the raw output length: we count the English words of each solution and prefer the shorter one, using the same $10\%$ threshold for ties as in \cref{sec:methodology-metrics}. This requires no model call at all, but agrees with the human annotations of \cref{app:human-validation} only $42.3\%$ of the time, i.e.\ less often than not. Our compressibility metric reaches $59.9\%$, which is low in absolute terms but is the expected range given the $57.6\%$ agreement between the annotators themselves. Raw length is thus likely uninformative, which is consistent with our motivation for measuring compressibility instead, as a proof can be long because the approach is hard rather than because it is verbose.

For computational ease and cognitive simplicity, the cheaper alternative is an ordinal judge that scores each solution from $0$ to $10$ against a rubric, which needs one call per solution rather than one per pair and therefore scales linearly rather than quadratically in the number of models. We describe this design and why we abandoned it in \cref{app:ordinal-ranking}. It agrees with the human annotations $66.7\%$ and $66.1\%$ of the time, against $82.1\%$ and $85.3\%$ for the pairwise judges we use.

For correctness verification, we compare against \opcrone{} \citep{opc}, a model trained specifically for this task. Both it and our two-stage filter are scored against ground-truth verdicts from manually resolving every disagreement in \cref{app:correctness-verifier}. It agrees in $71.5\%$ of cases, against $93.5\%$ for our filter. This is below the accuracy reported by \citet{opc}, which we attribute to our dataset being harder and our evaluated models stronger, so that their incorrect solutions are more convincing. It is a reasonable baseline, but only marginally cheaper, as both require one or two judge calls per attempt.

The two remaining metrics do not admit a comparison of this form. For diversity, no single ground-truth clustering exists, which is why we validate the consistency of the procedure instead (\cref{app:diversity-consistency}). For adaptivity, the judge already makes a single call per solution, and symbolic methods cannot establish whether a technique is used substantively rather than merely mentioned, so there is no cheaper alternative to compare against.

\section{Additional Results} \label{app:additional-experiments}

In this section, we provide additional experimental results that complement the evaluations in \cref{sec:evaluation}.

\subsection{Absolute Metric Scores} \label{app:raw-scores}

For conciseness, diversity, and adaptivity, the underlying metric is a scalar, reported in \cref{tab:raw_scores}.

\begin{wraptable}{r}{0.6\linewidth}
    \caption{Raw scores across the different metrics corresponding to the main results.}
    \label{tab:raw_scores}
    \centering
    \resizebox{\linewidth}{!}{
    \begin{tabular}{l c c c}
        \toprule
        & {\textbf{Compressibility}} 
        & {\textbf{Diversity}} 
        & {\textbf{Adaptivity}} \\
        \midrule
        \gptfivesix{} & 1.46 & 0.25 & \textbf{90.8\%} \\
        \gptfivefour{} & 1.93 & 0.37 & 77.7\% \\
        \dsvfourpro{} & \textbf{1.43} & 0.36 & 75.2\% \\
        \geminiflashseven{} & 2.90 & 0.33 & 58.0\% \\
        \geminipro{} & \textit{3.51} & 0.40 & 52.1\% \\
        \kimi{} & 1.92 & \textbf{0.41} & 29.2\% \\
        \glmfive{} & 2.78 & 0.21 & 39.6\% \\
        \dsvthreetwo{} & 2.39 & \textbf{0.41} & 27.2\% \\
        \gptoss{} & 1.98 & 0.24 & \textit{21.6\%} \\
        \stepfun{} & 2.01 & 0.26 & 40.9\% \\
        \grokfourfast{} & 2.09 & 0.21 & 25.2\% \\
        \geminiflash{} & 2.39 & \textit{0.17} & 24.5\% \\
        \qwenthreebig{} & 2.64 & 0.27 & 22.9\% \\
        \bottomrule
    \end{tabular}
    }
\end{wraptable}

\paragraph{Differences between absolute and relative metrics} The raw scores in \cref{tab:raw_scores} correlate strongly with the Elo ratings, with Spearman rank correlations of $-0.95$, $0.96$, and $1$ for conciseness, diversity, and adaptivity, respectively. The small discrepancy for conciseness arises because models occasionally produce outlier solutions with extremely high compressibility. For instance, \grokfourfast{} sometimes duplicates its entire solution. These outliers explain the swap in the ranking of \grokfourfast{} and \gptfivefour{}, which does not have this issue. Similarly, whenever \kimi{} produces more than one solution, it tends to produce many different ones. This makes its diversity more variable across samples, and explains the swap in the ranking of \kimi{} and \geminipro{}, which is more consistent across samples.

\subsection{Further Directed Prompting Results} \label{app:steering}

We extend the results reported in \cref{sec:steerability}. This includes further analysis of the directed prompting experiments, and a deeper dive into the systematics of the setup. We first include per-model results in \cref{tab:steering_elo}, which shows the difference each prompted variant makes relative to the standard prompt, with frozen ratings for all base models. This is complemented by \cref{tab:steering_acc}, which contrasts the steerability against the models' utility decline. Alongside these, we also fit an independent rating over comparisons between prompted variants only, reported in \cref{tab:steering_c}; the $\rho$ and inversion counts of \cref{tab:steering} are computed from this fit.
\begin{table*}[t]
    \caption{Per-model effect of each directed prompt, rated against the standard prompt (Std.).}
    \label{tab:steering_elo}
    \centering
    \resizebox{\linewidth}{!}{
    \begin{tabular}{l r c c r c c r c c}
        \toprule
        & \multicolumn{3}{c}{\textbf{Cognitive Simplicity}}
        & \multicolumn{3}{c}{\textbf{Computational Ease}}
        & \multicolumn{3}{c}{\textbf{Conciseness}} \\
        \cmidrule(lr){2-4} \cmidrule(lr){5-7} \cmidrule(lr){8-10}
        & Std. & easier & harder
        & Std. & lighter & heavier
        & Std. & mild & strong \\
        \midrule
        \gptfivesix{}
          & 1257 & $+103^{+31}_{-29}$ & $-325^{+33}_{-39}$
          & 1610 & $+444^{+58}_{-57}$ & $-222^{+48}_{-48}$
          & 1946 & $+962^{+79}_{-76}$ & $+1807^{+169}_{-159}$ \\
        \gptfivefour{}
          & 1339 & $+41^{+35}_{-35}$  & $-357^{+43}_{-41}$
          & 1566 & $+319^{+59}_{-55}$ & $-374^{+46}_{-53}$
          & 1296 & $+1170^{+65}_{-58}$ & $+1791^{+97}_{-91}$ \\
        \dsvfourpro{}
          & 1306 & $+40^{+35}_{-34}$  & $-269^{+40}_{-40}$
          & 1638 & $+239^{+58}_{-57}$ & $-256^{+50}_{-46}$
          & 1974 & $+505^{+62}_{-64}$ & $+881^{+79}_{-79}$ \\
        \geminiflashseven{}
          & 1096 & $+126^{+38}_{-33}$ & $-316^{+41}_{-43}$
          & 852  & $+426^{+56}_{-51}$ & $-179^{+57}_{-54}$
          & 465  & $+742^{+52}_{-52}$ & $+1755^{+70}_{-62}$ \\
        \geminipro{}
          & 1170 & $+103^{+40}_{-35}$ & $-368^{+47}_{-43}$
          & 1028 & $+250^{+52}_{-49}$ & $-181^{+56}_{-57}$
          & 167  & $+424^{+71}_{-65}$ & $+1265^{+74}_{-64}$ \\
        \kimi{}
          & 1099 & $+113^{+34}_{-44}$ & $-262^{+43}_{-47}$
          & 1216 & $+364^{+56}_{-55}$ & $-229^{+57}_{-58}$
          & 1464 & $+317^{+56}_{-54}$ & $+1090^{+81}_{-82}$ \\
        \glmfive{}
          & 1181 & $+19^{+45}_{-44}$  & $-181^{+40}_{-46}$
          & 826  & $+411^{+61}_{-69}$ & $+31^{+60}_{-52}$
          & 843  & $+765^{+57}_{-57}$ & $+1033^{+69}_{-65}$ \\
        \dsvthreetwo{}
          & 1266 & $+66^{+47}_{-45}$  & $-108^{+46}_{-44}$
          & 988  & $+188^{+69}_{-68}$ & $-93^{+61}_{-64}$
          & 1141 & $+891^{+74}_{-82}$ & $+1226^{+107}_{-83}$ \\
        \gptoss{}
          & 1098 & $+50^{+43}_{-38}$  & $-176^{+50}_{-45}$
          & 1063 & $+190^{+63}_{-63}$ & $-182^{+62}_{-66}$
          & 1382 & $+634^{+73}_{-62}$ & $+1114^{+84}_{-94}$ \\
        \geminiflash{}
          & 1247 & $+73^{+44}_{-42}$  & $-150^{+46}_{-49}$
          & 1260 & $+162^{+63}_{-67}$ & $-46^{+61}_{-72}$
          & 1118 & $+368^{+67}_{-62}$ & $+854^{+69}_{-69}$ \\
        \qwenthreebig{}
          & 1215 & $+54^{+43}_{-43}$  & $-196^{+46}_{-48}$
          & 1414 & $+103^{+62}_{-69}$ & $-148^{+62}_{-70}$
          & 1053 & $+250^{+56}_{-60}$ & $+888^{+84}_{-75}$ \\
        \bottomrule
    \end{tabular}
    }
\end{table*}

\begin{table}[t]
    \caption{Per-model accuracy change (percentage points) under each directed prompt.}
    \label{tab:steering_acc}
    \centering
    \resizebox{0.75\linewidth}{!}{
    \begin{tabular}{l r r r r r r r}
        \toprule
        & \multicolumn{2}{c}{\textbf{Cognitive}}
        & \multicolumn{2}{c}{\textbf{Computation}}
        & \multicolumn{2}{c}{\textbf{Conciseness}}
        & \\
        \cmidrule(lr){2-3} \cmidrule(lr){4-5} \cmidrule(lr){6-7}
        & easier & harder & lighter & heavier & mild & strong & \textbf{median} \\
        \midrule
        \gptfivesix{}       & $0.0$   & $-0.2$   & $-2.3$   & $+0.5$   & $-5.5$   & $-31.7$  & $-1.3$ \\
        \gptfivefour{}      & $+2.4$  & $-1.1$   & $+0.4$   & $+2.3$   & $-0.1$   & $-9.4$   & $+0.2$ \\
        \dsvfourpro{}       & $-3.6$  & $-2.0$   & $-10.9$  & $-5.8$   & $-7.9$   & $-14.1$  & $-6.9$ \\
        \geminiflashseven{} & $-2.1$  & $-2.9$   & $-7.1$   & $-1.0$   & $-7.9$   & $-19.9$  & $-5.0$ \\
        \geminipro{}        & $-0.3$  & $0.0$    & $+1.3$   & $+3.7$   & $+1.0$   & $-6.8$   & $+0.5$ \\
        \kimi{}             & $-5.1$  & $-4.9$   & $-3.0$   & $-2.8$   & $-15.6$  & $-19.0$  & $-5.0$ \\
        \glmfive{}          & $-6.2$  & $+0.1$   & $+1.2$   & $+3.0$   & $+1.3$   & $-10.9$  & $+0.7$ \\
        \dsvthreetwo{}      & $-14.7$ & $-11.8$  & $-13.4$  & $-13.4$  & $-11.5$  & $-17.6$  & $-13.4$ \\
        \gptoss{}           & $-1.8$  & $-2.4$   & $-3.7$   & $-2.4$   & $-7.6$   & $-9.2$   & $-3.1$ \\
        \geminiflash{}      & $+1.6$  & $+1.8$   & $+3.1$   & $+1.0$   & $+3.9$   & $+5.0$   & $+2.5$ \\
        \qwenthreebig{}     & $+5.6$  & $+1.9$   & $+2.9$   & $+4.0$   & $+2.7$   & $-0.2$   & $+2.8$ \\
        \midrule
        \textbf{median}     & $-1.8$  & $-1.1$   & $-2.3$   & $+0.5$   & $-5.5$   & $-10.9$  & $-2.2$ \\
        \bottomrule
    \end{tabular}
    }
\end{table}

\begin{table}[t]
    \centering
    \caption{Independent variants-only ratings, fitted on comparisons between prompted variants.}
    \label{tab:steering_c}
    \setlength{\tabcolsep}{10pt}
    \begin{tabular}{l r r r r r r}
        \toprule
        & \multicolumn{2}{c}{\textbf{Cognitive}}
        & \multicolumn{2}{c}{\textbf{Computation}}
        & \multicolumn{2}{c}{\textbf{Conciseness}} \\
        \cmidrule(lr){2-3} \cmidrule(lr){4-5} \cmidrule(lr){6-7}
        & easier & harder & lighter & heavier & mild & strong \\
        \midrule
        \gptfivesix{}       & 1284 & 1157 & \textbf{1768} & 1483 & \textbf{2099} & \textbf{2167} \\
        \gptfivefour{}      & \textbf{1317} & 1212 & 1539 & 1343 & 1764 & 1635 \\
        \dsvfourpro{}       & 1276 & 1245 & 1527 & \textbf{1489} & 1760 & 1515 \\
        \geminiflashseven{} & 1149 & \textit{954}  & 967  & \textit{837}  & 620  & 1075 \\
        \geminipro{}        & 1173 & 1042 & 968  & 1024 & \textit{25}   & \textit{301} \\
        \kimi{}             & 1136 & 1098 & 1274 & 1143 & 1241 & 1293 \\
        \glmfive{}          & 1112 & 1257 & 975  & 999  & 1126 & 830 \\
        \dsvthreetwo{}      & 1245 & \textbf{1413} & 938  & 1039 & 1446 & 1459 \\
        \gptoss{}           & \textit{1087} & 1200 & \textit{901}  & 1031 & 1372 & 1211 \\
        \geminiflash{}      & 1235 & 1329 & 1090 & 1390 & 969  & 853 \\
        \qwenthreebig{}     & 1187 & 1293 & 1253 & 1423 & 779  & 862 \\
        \bottomrule
    \end{tabular}
\end{table}

\begin{table}[t]
    \caption{Cumulative distribution of rank inversions between the main and prompt-restricted rankings. ``All pairs'' denotes what fraction of the model pairs are considered in each block.}
    \label{tab:steering_inversion_gap}
    \centering
    \resizebox{0.75\linewidth}{!}{
    \begin{tabular}{l l r r r r r}
        \toprule
        & \textbf{Prompt} & \textbf{Inv.}
        & \multicolumn{4}{c}{\textbf{Cumulative share at base Elo gap}} \\
        \cmidrule(lr){4-7}
        & & & $\leq 50$ & $\leq 100$ & $\leq 200$ & $\leq 400$ \\
        \midrule
        \multirow{3}{*}{Cognitive}
          & easier   & $7$ & $71\%$ & $100\%$ & $100\%$ & $100\%$ \\
          & harder   & $16$ & $31\%$ & $75\%$ & $100\%$ & $100\%$ \\
          & \emph{all pairs} & -- & \emph{$25\%$} & \emph{$58\%$} & \emph{$89\%$} & \emph{$100\%$} \\
        \midrule
        \multirow{3}{*}{Computation}
          & lighter  & $12$ & $33\%$ & $50\%$ & $75\%$ & $100\%$ \\
          & heavier  & $5$ & $40\%$ & $60\%$ & $80\%$ & $100\%$ \\
          & \emph{all pairs} & -- & \emph{$11\%$} & \emph{$15\%$} & \emph{$33\%$} & \emph{$62\%$} \\
        \midrule
        \multirow{3}{*}{Conciseness}
          & mild     & $9$ & $11\%$ & $33\%$ & $44\%$ & $89\%$ \\
          & strong   & $10$ & $10\%$ & $30\%$ & $40\%$ & $70\%$ \\
          & \emph{all pairs} & -- & \emph{$4\%$} & \emph{$11\%$} & \emph{$16\%$} & \emph{$36\%$} \\
        \midrule
        & all prompts & $59$ & $31\%$ & $58\%$ & $75\%$ & $93\%$ \\
        & \emph{all pairs} & -- & \emph{$13\%$} & \emph{$28\%$} & \emph{$46\%$} & \emph{$66\%$} \\
        \bottomrule
    \end{tabular}
    }
\end{table}

\paragraph{Where the inversions occur} We also investigate how the discordant pairs appear. \Cref{tab:steering_inversion_gap} shows the cumulative distribution of swaps across different base model Elo gaps, showing that they concentrate at small gaps. Namely, $58\%$ of the $59$ inversions occur between models at most $100$ Elo apart, although only $28\%$ of the $330$ pairs are that close. This difference implies that such swaps are highly expected, and more likely a result of sample noise than a significant shift in rankings.

\begin{table*}[t]
    \caption{Diversity under a single call asking for several distinct methods, vs. the standard setting of four independent calls (Std.). Elo columns use the same cluster-entropy score as \cref{tab:main_bench}, with standard-prompt models held at their leaderboard ratings, so the Std. columns match \cref{tab:main_bench,tab:all_bench}.}
    \label{tab:diversity_prompt}
    \centering
    \newcommand{\ratingdelta}[2]{%
        \makebox[5.6em][c]{%
            \makebox[2.2em][r]{#1}%
            \,\makebox[2.9em][l]{#2}%
        }%
    }
    \resizebox{\linewidth}{!}{
    \begin{tabular}{l r c r c c c}
        \toprule
        & \multicolumn{2}{c}{\textbf{Elo, all solutions}}
        & \multicolumn{2}{c}{\textbf{Elo, correct only}}
        & \textbf{Distinct methods}
        & \textbf{\#Correct of 4} \\
        \cmidrule(lr){2-3} \cmidrule(lr){4-5} \cmidrule(lr){6-6} \cmidrule(lr){7-7}
        & Std. & single call & Std. & single call
        & Std.\ $\to$ single & Std.\ $\to$ single \\
        \midrule
        \gptfivesix{}       & 1002 & \ratingdelta{1713}{(+711)} & 1135 & \ratingdelta{2027}{(+892)} & $1.42 \to 3.05$ & $3.80 \to 3.46$ \\
        \gptfivefour{}      & 1095 & \ratingdelta{1425}{(+330)} & 1236 & \ratingdelta{1631}{(+395)} & $1.58 \to 2.31$ & $3.51 \to 3.34$ \\
        \dsvfourpro{}       & 1223 & \ratingdelta{1396}{(+173)} & 1246 & \ratingdelta{1556}{(+310)} & $1.68 \to 2.20$ & $2.86 \to 3.07$ \\
        \geminiflashseven{} & 1089 & \ratingdelta{1363}{(+274)} & 1227 & \ratingdelta{1438}{(+211)} & $1.55 \to 2.20$ & $2.94 \to 2.12$ \\
        \geminipro{}        & 1168 & \ratingdelta{1486}{(+318)} & 1287 & \ratingdelta{1496}{(+209)} & $1.70 \to 2.46$ & $2.74 \to 1.72$ \\
        \kimi{}             & 1286 & \ratingdelta{1472}{(+186)} & 1253 & \ratingdelta{1407}{(+154)} & $2.17 \to 2.46$ & $1.79 \to 1.12$ \\
        \glmfive{}          & 1238 & \ratingdelta{1419}{(+181)} & 1132 & \ratingdelta{1459}{(+327)} & $1.71 \to 2.34$ & $1.24 \to 1.14$ \\
        \dsvthreetwo{}      & 1249 & \ratingdelta{1417}{(+168)} & 1311 & \ratingdelta{1304}{(-7)} & $2.26 \to 2.32$ & $1.30 \to 0.91$ \\
        \gptoss{}           & 1301 & \ratingdelta{1573}{(+272)} & 1179 & \ratingdelta{1320}{(+141)} & $1.97 \to 2.66$ & $1.55 \to 0.65$ \\
        \geminiflash{}      & 1101 & \ratingdelta{1301}{(+200)} & 1094 & \ratingdelta{1288}{(+194)} & $1.58 \to 2.02$ & $1.30 \to 0.86$ \\
        \qwenthreebig{}     & 1258 & \ratingdelta{1435}{(+177)} & 1175 & \ratingdelta{1356}{(+181)} & $1.92 \to 2.30$ & $1.19 \to 0.59$ \\
        \bottomrule
    \end{tabular}
    }
\end{table*}

\paragraph{Diversity responds differently to prompting} \label{app:steering-diversity} Since diversity is defined over independent samples, we cannot steer it by appending an instruction to the generation prompt. Instead, we ask for several distinct methods within a single call (\cref{app:steering_prompts}), to distinguish between natural and instructed diversity. As shown in \cref{tab:diversity_prompt}, this raises diversity for nearly every model, with the average number of distinct method clusters per problem growing from $1.78$ to $2.39$ and a median gain of $209$ Elo; the only exception is \dsvthreetwo{}, which is flat at $-7$ Elo. At the same time, the number of correct solutions out of four falls for 10 of the 11 models, from $2.20$ to $1.73$ on average, a median drop of $11.0$ accuracy points. Diversity is also the setting in which the ranking shifts the most: the single-call ranking attains $\rho = 0.07$ against the standard-prompt ranking, with $27$ of the $55$ pairs inverted. This is expected, as the intervention replaces the generation protocol rather than appending an instruction to it, so default diversity is a weaker predictor of prompted diversity than is the case for the other metrics. Instead, prompted diversity closely tracks adaptivity, with $\rho = 0.90$ against its ranking, compared to $0.13$ for default diversity, as it measures the ability to produce alternative approaches on demand rather than the spread of a model's natural ones.

\subsection{\bench{} on Correct-Answer Solutions} \label{app:bench-correct-answers}

We repeat the experiments of \cref{sec:results-ablation}, but now restrict the evaluation to solutions verified to have a correct final answer, rather than removing the correctness dependence entirely. As shown in \cref{tab:ca_bench}, the rating differences under this restriction follow a trend similar to those without any correctness filtering. Of the $52$ model-metric differences, $50$ point in the same direction as in \cref{tab:all_bench}, and $39$ are smaller in magnitude, so answer-only filtering only  recovers part of the effect of the full correctness filter. The two sign changes both occur on differences below $30$ Elo, within sampling noise.

Two models stand out. \stepfun{} remains at approximately the same ratings as in the main results (\cref{sec:results-main}), whereas dropping the filter entirely moves it by $142$ to $326$ Elo. This suggests that its solutions with an incorrect final answer are substantially different in style, being less verbose and overall simpler, which leads to a major ranking shift when they are included. Conversely, \geminiflash{} is the one model whose gains are uniformly larger here than without any filter, indicating that the solutions it recovers under the weaker filter, which reach the correct answer but do not justify it, are the ones that flatter it most.

\begin{table*}[t]
    \caption{Model ratings across the different \bench metrics (excluding Adaptivity) on solutions with correct answers (CA). We also show the differences in Elo with respect to the main results.}
    \label{tab:ca_bench}
    \centering
    \resizebox{\linewidth}{!}{
    \begin{tabular}{l c c c c}
        \toprule
        & \textbf{Conciseness} 
        & \textbf{Computational Ease}
        & \textbf{Cognitive Simplicity} 
        & \textbf{Diversity} \\
        \midrule
        \gptfivesix{} & 1681 (-263) & 1354 (-255) & 1139 (-117) & \textit{1052} (-83) \\
        \gptfivefour{} & 1167 (-128) & 1386 (-180) & 1244 (-94) & 1142 (-94) \\
        \dsvfourpro{} & \textbf{1756} (-217) & 1416 (-220) & 1213 (-93) & 1178 (-68) \\
        \geminiflashseven{} & 485 (+22) & \textit{795} (-57) & \textit{1028} (-68) & 1165 (-62) \\
        \geminipro{} & \textit{363} (+198) & 1032 (+5) & 1116 (-53) & 1210 (-77) \\
        \kimi{} & 1381 (-82) & 1186 (-29) & 1084 (-14) & \textbf{1316} (+63) \\
        \glmfive{} & 828 (-22) & 937 (+108) & 1213 (+31) & 1188 (+56) \\
        \dsvthreetwo{} & 1162 (+18) & 1119 (+131) & 1303 (+37) & 1232 (-79) \\
        \gptoss{} & 1263 (-120) & 1067 (+3) & 1091 (-7) & 1282 (+103) \\
        \stepfun{} & 1390 (+45) & 1190 (+39) & 1184 (+39) & 1256 (+62) \\
        \grokfourfast{} & 1604 (+197) & 1249 (+259) & 1325 (+141) & 1160 (+31) \\
        \geminiflash{} & 1360 (+243) & 1356 (+96) & 1309 (+62) & 1148 (+54) \\
        \qwenthreebig{} & 1161 (+110) & \textbf{1514} (+100) & \textbf{1352} (+137) & 1270 (+95) \\
        \bottomrule
    \end{tabular}
    }
\end{table*}

\subsection{Impact of Rephrasing Validity on Conciseness} \label{app:rephrasing_validity}

As we explain in \cref{sec:methodology-metrics}, we use a strong reasoning LLM to shorten solutions while preserving their completeness and correctness, which allows us to compute our conciseness metric. However, this introduces the risk that the rephrasing process may add or omit important details. To assess this, we use an LLM to automatically determine whether the original and rephrased solutions are fully equivalent, disregarding minor stylistic differences. 

\begin{figure*}[t]
    \centering
    \includegraphics[width=\textwidth]{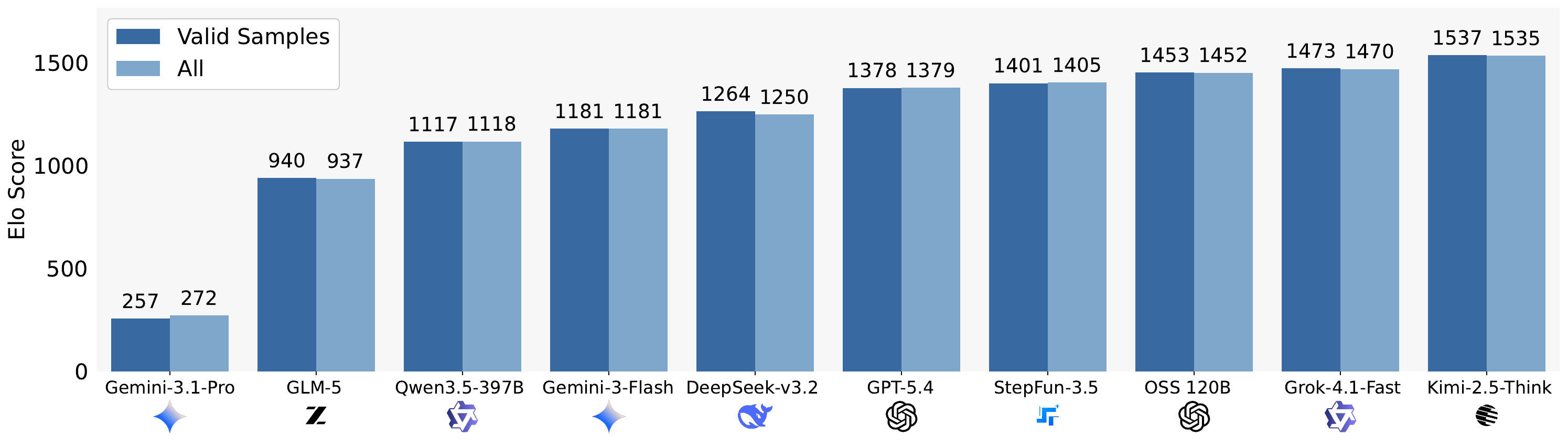}
    \caption{Elo changes when only considering valid rephrasings for the conciseness metric.}
    \label{fig:conciseness-validity}
\end{figure*}

We find that $94.1\%$ of the rephrasings are fully valid, with the remaining $5.9\%$ potentially affecting the reliability of the conciseness metric. We examine the resulting Elo changes when only considering valid rephrasings in \cref{fig:conciseness-validity}, which shows that the Elo differences are minimal for all models, with no changes larger than 10 to 15 points. This change is much smaller than the sampling error (\cref{app:main-results-ci}), suggesting that our conclusions persist under both settings.

\subsection{Confidence Intervals for Main Results} \label{app:main-results-ci}

To assess the statistical significance of our main results, we compute $95\%$ confidence intervals for all Elo scores and accuracy rates in \cref{tab:main_bench_ci}. We do so by 1000-fold bootstrapping on the pairwise judgments for the quality metrics and by assuming a binomial distribution for the accuracy rates.

\begin{table*}[t]
    \caption{Main model ratings across the different \bench metrics, measured on correct solutions only, alongside their accuracy rate. We include 95\% confidence intervals for all metrics in the superscripts. The Elo intervals are $95\%$ percentile bootstrap intervals over 1000 resamples, each refitted through the same procedure as the point estimate, while the accuracy intervals are binomial intervals over the 382 problems.}
    \label{tab:main_bench_ci}
    \centering
    \resizebox{\linewidth}{!}{
    \begin{tabular}{l c c c c c || c}
        \toprule
        & {\textbf{Conciseness}}
        & {\textbf{Computational Ease}}
        & {\textbf{Cognitive Simplicity}}
        & {\textbf{Diversity}}
        & {\textbf{Adaptivity}}
        & {\textbf{Accuracy}} \\
        \midrule
        \gptfivesix{} & $1944^{+42}_{-38}$ & $1609^{+36}_{-40}$ & $1256^{+25}_{-24}$ & $1135^{+30}_{-31}$ & $\mathbf{1619^{+14}_{-13}}$ & $\mathbf{93.4^{\pm 2.5}}\%$ \\
        \gptfivefour{} & $1295^{+33}_{-35}$ & $1566^{+35}_{-34}$ & $\mathbf{1338^{+27}_{-25}}$ & $1236^{+34}_{-35}$ & $1484^{+12}_{-12}$ & $86.3^{\pm 3.4}\%$ \\
        \dsvfourpro{} & $\mathbf{1973^{+42}_{-38}}$ & $\mathbf{1636^{+40}_{-37}}$ & $1306^{+24}_{-25}$ & $1246^{+35}_{-34}$ & $1431^{+13}_{-14}$ & $72.0^{\pm 4.5}\%$ \\
        \geminiflashseven{} & $463^{+41}_{-42}$ & $852^{+40}_{-38}$ & $\mathit{1096^{+28}_{-26}}$ & $1227^{+36}_{-31}$ & $1314^{+12}_{-11}$ & $71.2^{\pm 4.5}\%$ \\
        \geminipro{} & $\mathit{165^{+49}_{-57}}$ & $1027^{+37}_{-36}$ & $1169^{+25}_{-28}$ & $1287^{+37}_{-36}$ & $1264^{+10}_{-11}$ & $60.7^{\pm 4.9}\%$ \\
        \kimi{} & $1463^{+44}_{-40}$ & $1215^{+38}_{-39}$ & $1098^{+29}_{-31}$ & $1253^{+45}_{-44}$ & $1067^{+10}_{-12}$ & $53.5^{\pm 5.0}\%$ \\
        \glmfive{} & $850^{+44}_{-45}$ & $\mathit{829^{+41}_{-46}}$ & $1182^{+29}_{-29}$ & $1132^{+35}_{-35}$ & $1139^{+11}_{-12}$ & $49.1^{\pm 5.0}\%$ \\
        \dsvthreetwo{} & $1144^{+51}_{-51}$ & $988^{+51}_{-51}$ & $1266^{+34}_{-31}$ & $\mathbf{1311^{+52}_{-50}}$ & $1049^{+11}_{-11}$ & $47.9^{\pm 5.0}\%$ \\
        \gptoss{} & $1383^{+39}_{-39}$ & $1064^{+41}_{-42}$ & $1098^{+29}_{-29}$ & $1179^{+34}_{-36}$ & $\mathit{999^{+10}_{-11}}$ & $41.1^{\pm 4.9}\%$ \\
        \stepfun{} & $1345^{+41}_{-43}$ & $1151^{+46}_{-48}$ & $1145^{+29}_{-33}$ & $1194^{+35}_{-33}$ & $1168^{+11}_{-11}$ & $34.3^{\pm 4.8}\%$ \\
        \grokfourfast{} & $1407^{+50}_{-45}$ & $990^{+43}_{-48}$ & $1184^{+32}_{-31}$ & $1129^{+39}_{-43}$ & $1031^{+10}_{-11}$ & $31.4^{\pm 4.7}\%$ \\
        \geminiflash{} & $1117^{+42}_{-46}$ & $1260^{+46}_{-45}$ & $1247^{+29}_{-30}$ & $\mathit{1094^{+37}_{-37}}$ & $1025^{+11}_{-11}$ & $29.8^{\pm 4.6}\%$ \\
        \qwenthreebig{} & $1051^{+47}_{-47}$ & $1414^{+49}_{-47}$ & $1215^{+31}_{-31}$ & $1175^{+37}_{-39}$ & $1011^{+10}_{-11}$ & $\mathit{29.3^{\pm 4.6}}\%$ \\
        \midrule
        \midrule
        \geminiflashlow{} & $1217^{+51}_{-51}$ & $937^{+59}_{-63}$ & $1196^{+25}_{-29}$ & $1081^{+29}_{-32}$ & $975^{+8}_{-7}$ & $16.5^{\pm 3.7}\%$ \\
        \qwenthreeeight{} & $750^{+74}_{-85}$ & $1209^{+81}_{-80}$ & $1324^{+34}_{-33}$ & $1067^{+39}_{-44}$ & $939^{+8}_{-8}$ & $8.1^{\pm 2.7}\%$ \\
        \bottomrule
    \end{tabular}
    }
\end{table*}

\section{Methodological Challenges and Failed Attempts} \label{app:challenges}

Throughout our work, we explored a variety of methodologies commonly used in the LLM evaluation literature. However, we found that many of these approaches were ineffective for evaluating aspects that are either too subjective or too nuanced for LLMs to assess consistently and accurately. In this section, we describe several of these attempted approaches and explain why they failed, both to contextualize our final design choices and to provide guidance for future work in this area.

\subsection{Ordinal Ranking of Cognitive Simplicity and Computational Ease} \label{app:ordinal-ranking}

Before adopting our pairwise comparison-based evaluation for cognitive simplicity and computational ease, we experimented with asking LLMs to assign scores to solutions on an ordinal scale. To do so, we designed a rubric for each proof aspect. For example, our cognitive simplicity rubric identified the most challenging proof concept in a solution and ranked it by the mathematical background required: for example, middle-school, IMO-level, or advanced research mathematics.

We found such rubrics difficult to use in our setting for two main reasons:
\begin{itemize}[itemsep=0em,leftmargin=2em]
    \item LLMs are not necessarily familiar with the background assumptions required by the rubric, and can therefore struggle to apply it consistently.
    \item Designing a comprehensive evaluation rubric is inherently difficult, as many nuances and edge cases cannot be captured by fixed rules, which increases noise and introduces potential bias.
\end{itemize}

We encountered similar challenges in the human evaluation of \cref{app:human-validation}, where we initially attempted to rate each aspect on a scale from 0 to 5. Not only was it hard to design an appropriate rubric for such ratings, but the subjective nature of the task also introduced substantial annotation noise, making it difficult to draw meaningful conclusions. We quantify the cost of this design in \cref{app:cheaper-baselines}. The ordinal judge is cheaper, as it requires one call per solution rather than one per pair, but agrees with our human annotations roughly $16$ to $19$ percentage points less often than our pairwise judges.

\subsection{Evaluation on Proof-based Problems} \label{app:proof-based-evaluation}

In an initial stage of our work, we considered constructing \bench{} from proof-based problems rather than restricting ourselves to final-answer problems. However, validating the correctness of proof-based solutions in a scalable manner is extremely challenging. Although \citet{opc} find that frontier LLMs perform well as correctness judges, these models are often expensive to use, making it financially infeasible to scale such evaluation to a benchmark of our size.

We also attempted to use smaller models as verifiers, such as \opcrone{}, \grokfourfast{}, and \gptoss{}, which are either open-weight or more affordable. However, we found that these models were either insufficiently accurate or significantly biased in favor of certain models over others. Such biases would substantially skew our results, so we decided against using LLMs as correctness judges.

Nevertheless, we still wanted to account for proof correctness in our evaluation, since final-answer solutions that do not present a complete proof are of limited value to the mathematical community. We therefore adopted the two-step correctness verification framework described in \cref{sec:methodology-dataset}.
\section{Additional Dataset Information} \label{app:dataset}

In this section, we describe the procedure used to construct the problem set for \bench{}. For competitions imported from MathArena, we exclude problems that are likely to be too straightforward to elicit meaningfully distinguishable model solutions. As a heuristic, we remove the first five problems from each AIME competition, treating AIME I and AIME II separately, and the first three problems from each HMMT category, with problems 1 to 10 corresponding to Algebra and Number Theory, 11 to 20 to Combinatorics, and 21 to 30 to Geometry. For \answerbench{} and the Apex/Apex Shortlist problems, we retain all problems except for four drawn from AIME 2025 or HMMT 2025, which are already included through the corresponding competition sources.

\begin{figure}[t]
    \centering
    \includegraphics[width=0.63\linewidth]{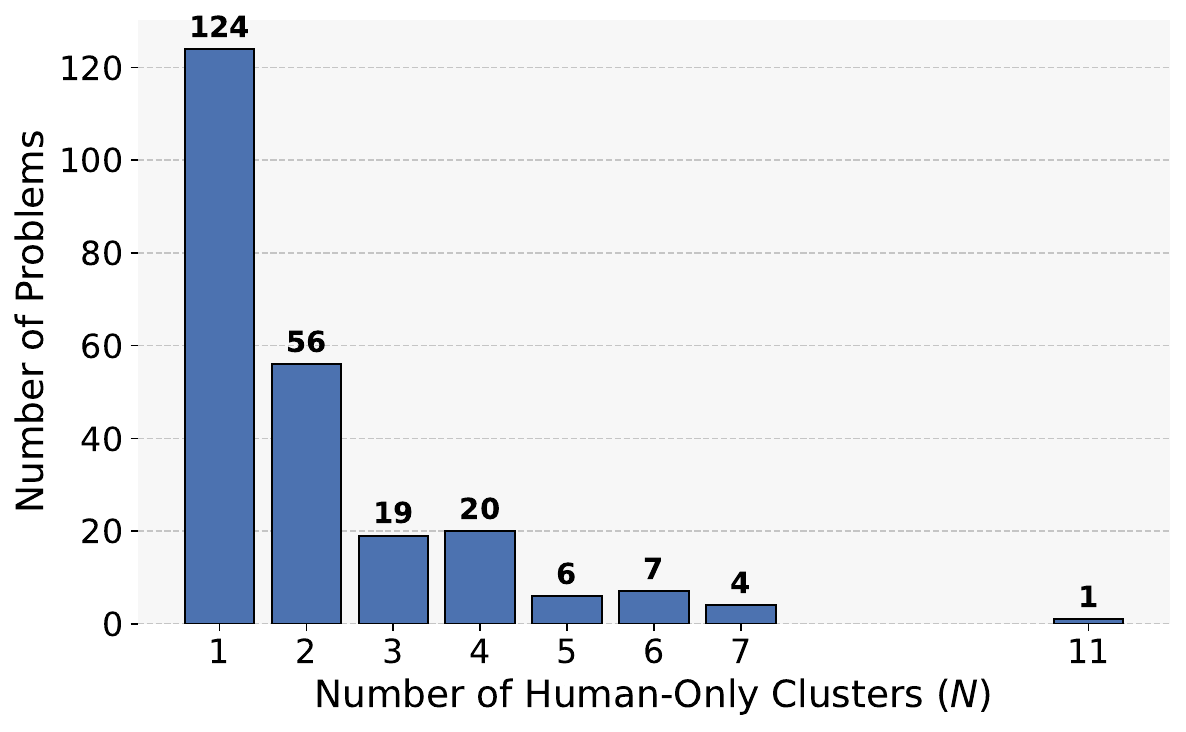}
    \caption{Distribution of the number of solutions per problem in our dataset.}
    \label{fig:solutions_per_problem}
\end{figure}

For each problem, we collect human-written solution attempts from public Art of Problem Solving (AoPS) posts or the official competition sources. We discard problems with no public solution. This filtering removed about 100 problems from the original \answerbench{} dataset of 400 problems. In addition, some \answerbench{} problems are substantially rephrased relative to their source problems. We remove about 60 such problems when the original solution strategy does not trivially transfer to the rephrased version. After these filtering steps, 241 \answerbench{} problems remain. The full dataset breakdown is reported in \cref{tab:dataset_distribution}.

\begin{table}[t]
    \centering
    \caption{A list of sources for the problems in \bench.}
    \resizebox{\linewidth}{!}{
        \begin{tabular}{
            llcc
            }
            \toprule
            \textbf{Competition} & \textbf{Description} & \textbf{Problems Included} & \textbf{Source}\\

\midrule
\midrule
\multicolumn{4}{c}{\textbf{MathArena}}\\
\midrule
\midrule
    AIME 2025 & Answer-based competition, serving as a qualifier for the USAMO & 6-15,21-30 & \href{https://matharena.ai/}{Public} \\
    HMMT February 2025 & Answer-based competition hosted by Harvard and MIT & 4-10,14-20,24-30 & \href{https://matharena.ai/}{Public} \\
    AIME 2026 & Answer-based competition, serving as a qualifier for the USAMO & 6-15,21-30 & \href{https://matharena.ai/}{Public} \\
    HMMT February 2026 & Answer-based competition hosted by Harvard and MIT & 4-10,14-20,24-33 & \href{https://matharena.ai/}{Public} \\
    Apex 2025 & Selected final-answer problems, which LLMs struggle to solve & All (12) & \href{https://matharena.ai/}{Public} \\
    Apex Shortlist 2025 & Selected final-answer problems, which LLMs find more challenging & 1-25,30-48 & \href{https://matharena.ai/}{Public} \\
\midrule
\midrule
\multicolumn{4}{c}{\textbf{IMO-AnswerBench}}\\
\midrule
\midrule
    Various & High-school-level competitions, from which problems were rephrased. & 241 problems & \href{https://github.com/google-deepmind/superhuman/blob/main/imobench/answerbench_v2.csv}{Public}\\
    \bottomrule
        \end{tabular}
    }
    \label{tab:dataset_distribution}
\end{table}

For adaptivity, we use these public solutions to sample up to 4 candidate techniques per problem, as described in \cref{sec:methodology}. This limits inference costs and prevents problems with unusually many available solution strategies from dominating the results. As shown in \cref{fig:solutions_per_problem}, the number of extracted techniques varies substantially across problems. The cap also mitigates the overrepresentation of geometry problems, which account for a large fraction of high-technique-count problems because many synthetic and algebraic solutions are possible.

\section{Qualitative Analysis} \label{app:qualitative-analysis}

The metrics we report in \bench{} aggregate hundreds of pairwise judgments per model, so no single comparison determines a rating. However, our human-led annotation process uncovered some particular representative model behaviours, as well as several cases, which clearly highlight the challenge of evaluating subjective and nuanced differences between solutions. In this section we walk through eight comparisons drawn from the doubly annotated subset described in \cref{app:human-validation}. In each case we show the problem, both model solutions, and the verdicts recorded by the annotators. \Cref{app:examples-clearcut} covers four pairs on which the annotators independently reached the same decisive verdicts, and \cref{app:examples-contentious} covers four on which they did not, or on which the LLM judge diverged from them. We further describe the importance of having a set of nuanced comparisons in \cref{app:examples-nuance}.

Throughout, we present verdicts in the direction in which they were elicited (see \cref{app:human-validation-instructions}): annotators were asked which solution is \emph{more} verbose, \emph{more} elegant, \emph{more} insightful, \emph{more} conceptually complex, and \emph{more} computationally demanding.

\subsection{Clear-Cut Comparisons} \label{app:examples-clearcut}

On a substantial fraction of pairs, both annotators independently reach the same decisive verdict, and on several they agree on every dimension. Three patterns stood out when reviewing these cases. First, annotators regularly prefer one solution on some dimensions and the other on the rest, so a unanimous verdict on one axis does not drag the remaining axes with it. Second, unanimous verdicts sometimes run against what superficial statistics would suggest, which is direct evidence that annotators read the solutions rather than count words. Third, when two solutions genuinely are similar on some dimension, both annotators independently record a tie there, so ties mostly indicate closeness rather than indecision. The examples below each illustrate one of these patterns.

\subsubsection{Unanimous Verdicts in Both Directions} \label{app:example-shortlist}

In this example with solutions, produced by \geminiflash{} and \stepfun{}, both annotators agree on all five dimensions, with no ties. This includes conciseness, even though the two solutions are of comparable length ($526$ versus $653$ words). The same problem returns in \cref{app:example-cognitive} with a different pair of models, where the annotators disagree.

\begin{exproblem}{Problem \texttt{shortlist\_2025\_6}}
Jerry places at most one rook in each cell of a $2025 \times 2025$ grid of cells. A rook attacks another rook if the two rooks are in the same row or column and there are no other rooks between them. Determine the maximum number of rooks Jerry can place on the grid such that no rook attacks $4$ other rooks.
\end{exproblem}

\begin{excompare}{gemini}{Solution 1 --- \geminiflash{}}{stepfun}{Solution 2 --- \stepfun{}}
To find the maximum number of rooks Jerry can place on a $2025 \times 2025$ grid such that no rook attacks $4$ other rooks, let $n = 2025$. A rook $r$ at cell $(i, j)$ attacks another rook if they are in the same row or column and no other rooks are between them. Each rook can attack at most $2$ rooks in its row (one to its left and one to its right) and at most $2$ rooks in its column (one above and one below). Let $a_H(r)$ and $a_V(r)$ be the number of rooks $r$ attacks in its row and column, respectively. Then $a(r) = a_H(r) + a_V(r)$, with $a_H(r), a_V(r) \in \{0, 1, 2\}$.

The condition $a(r) \le 3$ for all rooks $r$ means that no rook can simultaneously attack $2$ rooks in its row and $2$ rooks in its column. That is, $a_H(r) = 2$ and $a_V(r) = 2$ is impossible. Let $I_H$ be the set of ``interior'' rooks in their rows ($a_H(r) = 2$) and $I_V$ be the set of ``interior'' rooks in their columns ($a_V(r) = 2$). The condition is $I_H \cap I_V = \emptyset$.

Let $R$ be the set of all rooks placed, so $N = |R|$. Let $E_H$ be the set of ``edge'' rooks in their rows, which are rooks that are not interior rooks ($E_H = R \setminus I_H$). Similarly, $E_V = R \setminus I_V$. The condition $I_H \cap I_V = \emptyset$ is equivalent to $R = E_H \cup E_V$. Thus, by the principle of inclusion-exclusion:
\[
N = |E_H \cup E_V| = |E_H| + |E_V| - |E_H \cap E_V|
\]
Let $r_i$ be the number of rooks in row $i$ and $c_j$ be the number of rooks in column $j$. The number of horizontal edge rooks in row $i$ is $\min(r_i, 2)$. Thus, $|E_H| = \sum_{i=1}^n \min(r_i, 2) \le 2n$. Similarly, $|E_V| = \sum_{j=1}^n \min(c_j, 2) \le 2n$.

Let $i_{\min}$ and $i_{\max}$ be the first and last non-empty rows. Any rook $(i_{\min}, j)$ is the first rook in column $j$ because there are no rooks in rows $i < i_{\min}$. Thus, all $r_{i_{\min}}$ rooks in the first row are vertical edge rooks. The horizontal edge rooks in this row (of which there are $\min(r_{i_{\min}}, 2)$) are therefore in $E_H \cap E_V$. Similarly, the $\min(r_{i_{\max}}, 2)$ horizontal edge rooks in the last row are in $E_H \cap E_V$. If $i_{\min} < i_{\max}$, these two sets are disjoint, so $|E_H \cap E_V| \ge \min(r_{i_{\min}}, 2) + \min(r_{i_{\max}}, 2)$.

Substituting this into our equation for $N$:
\[
\begin{aligned}
N &\le \sum_{i=1}^n \min(r_i, 2) + 2n \\
  &\quad - (\min(r_{i_{\min}}, 2) + \min(r_{i_{\max}}, 2)) \\
  &= \sum_{i \neq i_{\min}, i_{\max}} \min(r_i, 2) + 2n
\end{aligned}
\]
There are at most $n-2$ terms in the sum, each at most $2$. Thus:
\[
N \le 2(n-2) + 2n = 4n - 4
\]
For $n = 2025$, this gives $N \le 4(2025) - 4 = 8100 - 4 = 8096$. Jerry can achieve this by placing rooks along the entire boundary of the grid. For such a configuration, each corner rook attacks $2$ rooks, and each other boundary rook attacks $3$ rooks.

The maximum number of rooks Jerry can place is $8096$.
\[\boxed{8096}\]

\exswitch

The maximum number of rooks Jerry can place on a $2025 \times 2025$ grid such that no rook attacks four other rooks is $4 \cdot 2025 - 4 = 8096$.

\textbf{Proof:}
A rook attacks another if they share a row or column with no rook between them. Thus a rook attacks a neighbor in each of the four directions (left, right, up, down) if such a neighbor exists. Hence a rook attacks four others exactly when there is at least one rook to its left, right, above, and below. Equivalently, the rook is \textbf{not} the leftmost or rightmost in its row, and \textbf{not} the topmost or bottommost in its column. Therefore, the condition ``no rook attacks $4$ other rooks'' means every rook is either leftmost or rightmost in its row, or topmost or bottommost in its column.

Define \textbf{border cells} as those with row index $1$ or $n$ or column index $1$ or $n$; there are $4n-4$ such cells. All other cells are \textbf{interior}.

\textbf{Claim 1.} If a rook is placed in an interior cell $(i,j)$ (with $2 \le i \le n-1$ and $2 \le j \le n-1$), then at least one of the border cells $(i,1)$, $(i,n)$, $(1,j)$, $(n,j)$ must be empty.

\emph{Proof.} The rook must be extreme in its row or column.
\begin{itemize}[itemsep=0em,leftmargin=2em,topsep=0.2em]
    \item If it is leftmost in row $i$, then no rook exists in row $i$ with column $< j$; in particular $(i,1)$ is empty.
    \item If it is rightmost, then $(i,n)$ is empty.
    \item If it is topmost in column $j$, then $(1,j)$ is empty.
    \item If it is bottommost, then $(n,j)$ is empty. $\blacksquare$
\end{itemize}

\textbf{Claim 2.} Two different interior rooks cannot force the same border cell to be empty.

\emph{Proof.} Consider a border cell $B$.
\begin{itemize}[itemsep=0em,leftmargin=2em,topsep=0.2em]
    \item If $B = (i,1)$, it can be forced empty only by an interior rook that is leftmost in row $i$; there is at most one such rook.
    \item If $B = (i,n)$, only a rightmost in row $i$.
    \item If $B = (1,j)$, only a topmost in column $j$.
    \item If $B = (n,j)$, only a bottommost in column $j$.
\end{itemize}
Thus at most one interior rook can be responsible for a given border cell. $\blacksquare$

Let $I$ be the number of interior rooks and $B$ the number of rooks on border cells. By Claim 1, each interior rook forces at least one border cell to be empty. By Claim 2, the forced border cells are distinct, so the number of empty border cells is at least $I$. The total number of border cells is $4n-4$, so the number of empty border cells equals $(4n-4) - B$. Hence $(4n-4) - B \ge I$, i.e., $B + I \le 4n-4$. Since the total number of rooks is $N = B + I$, we have $N \le 4n-4$.

\textbf{Construction achieving $4n-4$:} Place a rook on every border cell. This gives exactly $4n-4$ rooks.
\begin{itemize}[itemsep=0em,leftmargin=2em,topsep=0.2em]
    \item For a rook on row $1$ at column $j$: if $j=1$ or $j=n$, it is leftmost/rightmost in its row and also topmost/bottommost in its column. If $2 \le j \le n-1$, row $1$ is fully filled so the rook is not leftmost/rightmost, but column $j$ contains only rooks at rows $1$ and $n$ (since interior rows have no rook in column $j$), so it is topmost.
    \item For a rook on row $n$: analogous.
    \item For a rook on column $1$ at row $i$ with $2 \le i \le n-1$: row $i$ contains only rooks at columns $1$ and $n$ (since interior columns are empty), so the rook is leftmost. Column $1$ is fully filled, so it is not topmost/bottommost, but row-extreme suffices.
    \item For a rook on column $n$: analogous.
\end{itemize}
Thus every rook is extreme in its row or column, so the condition holds.

Therefore the maximum is exactly $4n-4$. For $n = 2025$, this is $4 \cdot 2025 - 4 = 8100 - 4 = 8096$.
\[\boxed{8096}\]
\end{excompare}

\begin{exverdict}{Annotator verdicts --- \texttt{shortlist\_2025\_6}}
\centering
\begin{tabular}{lcc}
    \toprule
    \textbf{Which solution is \dots} & \textbf{Annotator 1} & \textbf{Annotator 2} \\
    \midrule
    more verbose                   & Solution 2 & Solution 2 \\
    more elegant                   & Solution 2 & Solution 2 \\
    more insightful                & Solution 2 & Solution 2 \\
    more conceptually complex      & Solution 1 & Solution 1 \\
    more computationally demanding & Solution 1 & Solution 1 \\
    \bottomrule
\end{tabular}
\end{exverdict}

All ten judgments are decisive and all five dimensions agree. The longer solution, Solution 2, is unanimously the more verbose one, but it is also found to be the more elegant and the more insightful. In particular, its two claims, that each interior rook forces a border cell to be empty and that the forced cells are distinct, carry the bound in a much cleaner way. The shorter Solution 1 is unanimously judged both more conceptually complex and more computationally demanding, since it defines edge and interior rook sets, applies inclusion-exclusion, and then does additional bookkeeping over the extremal rows. Hence, judging verbosity does not require an extreme length gap, and this verdict does not bleed into the other four dimensions.

\subsubsection{Length does not Imply Verbosity} \label{app:example-aime2026}

Similar to the previous example, here the two solutions are within $60$ words of each other ($539$ versus $479$), yet the annotators are unanimous and decisive on four of the five dimensions.

\begin{exproblem}{Problem \texttt{aime\_2026\_9}}
Joanne has a blank fair six-sided die and six stickers each displaying a different integer from $1$ to $6$. Joanne rolls the die and then places the sticker labeled $1$ on the top face of the die. She then rolls the die again, places the sticker labeled $2$ on the top face, and continues this process to place the rest of the stickers in order. If the die ever lands with a sticker already on its top face, the new sticker is placed to cover the old sticker. Let $p$ be the conditional probability that at the end of the process exactly one face has been left blank, given that all the even-numbered stickers are visible on faces of the die. Then $p$ can be written as $\tfrac{m}{n}$, where $m$ and $n$ are relatively prime positive integers. Find $m + n$.
\end{exproblem}

\begin{excompare}{qwen}{Solution 1 --- \qwenthreeshort{}}{oai}{Solution 2 --- \gptfivefour{}}
To find the conditional probability $p$, we model the process using random variables $X_k \in \{1, 2, \dots, 6\}$ representing the face of the die that lands on top during the $k$-th roll, where sticker $k$ is placed. Since the die is fair, each $X_k$ is independently and uniformly distributed over the set of faces.

\textbf{1. Analyzing the condition ``even-numbered stickers are visible''.}
A sticker $k$ is visible if it is not covered by any sticker $j > k$. This occurs if and only if the face $X_k$ is not rolled again in any subsequent step.
\begin{itemize}[itemsep=0em,leftmargin=2em,topsep=0.2em]
    \item \textbf{Sticker 6:} Always visible as it is the last sticker placed.
    \item \textbf{Sticker 4:} Visible if $X_4 \neq X_5$ and $X_4 \neq X_6$.
    \item \textbf{Sticker 2:} Visible if $X_2 \neq X_3$, $X_2 \neq X_4$, $X_2 \neq X_5$, and $X_2 \neq X_6$.
\end{itemize}
Let $E$ be the event that stickers $2$, $4$, and $6$ are visible. The constraints on the sequence $(X_1, \dots, X_6)$ are:
\[ \begin{gathered} X_4 \neq X_5, \quad X_4 \neq X_6, \quad X_2 \neq X_3, \\ X_2 \neq X_4, \quad X_2 \neq X_5, \quad X_2 \neq X_6 \end{gathered} \]

\textbf{2. Calculating the size of event $E$.}
We count the number of sequences satisfying these constraints:
\begin{itemize}[itemsep=0em,leftmargin=2em,topsep=0.2em]
    \item $X_1$: $6$ choices.
    \item $X_2$: $6$ choices.
    \item $X_3$: $5$ choices ($X_3 \neq X_2$).
    \item $X_4$: $5$ choices ($X_4 \neq X_2$).
    \item $X_5$: $4$ choices ($X_5 \neq X_2$ and $X_5 \neq X_4$).
    \item $X_6$: $4$ choices ($X_6 \neq X_2$ and $X_6 \neq X_4$).
\end{itemize}
Total sequences in $E$:
\[ |E| = 6 \times 6 \times 5 \times 5 \times 4 \times 4 = 14400 \]

\textbf{3. Analyzing the condition ``exactly one face is blank''.}
Let $B$ be the event that exactly one face is left blank. A face is blank if it never lands on top. For exactly one face to be blank, the set of values $\{X_1, \dots, X_6\}$ must have cardinality $5$. This implies the sequence consists of $5$ distinct values, where one value $R$ appears twice and the other four values $y_1, y_2, y_3, y_4$ appear once.

\textbf{4. Calculating the size of $B \cap E$.}
To satisfy both $B$ and $E$, we choose a missing value $M$ ($6$ choices) and a repeated value $R$ ($5$ choices). For each pair $(M, R)$, we must place the two occurrences of $R$ at indices $(i, j)$ such that no constraint $x_i \neq x_j$ is violated. Since the other values are distinct, violations only occur if both occurrences of $R$ are placed at indices that are constrained to be different.

The forbidden pairs of indices for $R$ are:
\[ \{(2, 3), (2, 4), (2, 5), (2, 6), (4, 5), (4, 6)\} \]
There are $6$ forbidden pairs out of $\binom{6}{2} = 15$ total pairs. Thus, there are $15 - 6 = 9$ allowed pairs of positions for $R$. For each allowed pair, the remaining $4$ distinct values can be arranged in $4! = 24$ ways.
\[ |B \cap E| = (6 \times 5) \times 9 \times 24 = 30 \times 216 = 6480 \]

\textbf{5. Final probability.}
The conditional probability $p$ is:
\[ p = \frac{|B \cap E|}{|E|} = \frac{6480}{14400} = \frac{648}{1440} = \frac{9}{20} \]
Since $m=9$ and $n=20$ are relatively prime, we find:
\[ m + n = 9 + 20 = 29 \]
\[\boxed{29}\]

\exswitch

\[\boxed{29}\]

Let $X_i$ be the face on top on the $i$-th roll, for $i=1,2,\dots,6$. Since the die is fair and the rolls are independent, all $6^6$ sequences $(X_1,X_2,\dots,X_6)$ are equally likely.

A face is blank at the end exactly when it was never on top in any of the six rolls. So ``exactly one face has been left blank'' means that exactly $5$ distinct faces appear among
\[ X_1,X_2,X_3,X_4,X_5,X_6. \]

Now let $E$ be the event that all even-numbered stickers $2,4,6$ are visible at the end.
\begin{itemize}[itemsep=0em,leftmargin=2em,topsep=0.2em]
    \item Sticker $2$ is visible iff no later sticker covers it, so none of rolls $3,4,5,6$ may land on face $X_2$.
    \item Sticker $4$ is visible iff rolls $5,6$ do not land on face $X_4$.
    \item Sticker $6$ is always visible, since no sticker comes after it.
\end{itemize}
Thus $E$ is equivalent to:
\begin{itemize}[itemsep=0em,leftmargin=2em,topsep=0.2em]
    \item $X_2,X_4,X_6$ are pairwise distinct,
    \item $X_3\neq X_2$,
    \item $X_5\neq X_2$ and $X_5\neq X_4$.
\end{itemize}
No other restrictions are needed.

So the number of sequences satisfying $E$ is
\[ 6\cdot 5\cdot 4\cdot 5\cdot 4\cdot 6. \]
Indeed:
\begin{itemize}[itemsep=0em,leftmargin=2em,topsep=0.2em]
    \item $X_2$: $6$ choices,
    \item $X_4$: $5$ choices (not $X_2$),
    \item $X_6$: $4$ choices (not $X_2,X_4$),
    \item $X_3$: $5$ choices (anything but $X_2$),
    \item $X_5$: $4$ choices (anything but $X_2,X_4$),
    \item $X_1$: $6$ choices.
\end{itemize}

Now we count the sequences satisfying both $E$ and the condition ``exactly one face blank''.

First choose $X_2,X_4,X_6$. There are
\[ 6\cdot 5\cdot 4 \]
ways to do this, and these are three distinct faces. Call them $a,b,c$. Let the remaining three faces of the die be $d,e,f$.

Under $E$,
\begin{itemize}[itemsep=0em,leftmargin=2em,topsep=0.2em]
    \item $X_3$ can be any face except $a$, so $X_3\in\{b,c,d,e,f\}$,
    \item $X_5$ can be any face except $a,b$, so $X_5\in\{c,d,e,f\}$,
    \item $X_1$ can be any face.
\end{itemize}

Since $a,b,c$ are already used by rolls $2,4,6$, having exactly one blank face overall means: among the remaining faces $d,e,f$, exactly two are used by $X_1,X_3,X_5$, and one is never used.

Choose which one of $d,e,f$ stays blank: $3$ choices.

Suppose $f$ is the blank one. Then $X_1,X_3,X_5$ may only use faces from $\{a,b,c,d,e\}$, and both $d$ and $e$ must appear at least once.

Count these:
\begin{itemize}[itemsep=0em,leftmargin=2em,topsep=0.2em]
    \item Total choices avoiding $f$: $4\cdot 3\cdot 5=60$, since $X_3\in\{b,c,d,e\}$, $X_5\in\{c,d,e\}$, $X_1\in\{a,b,c,d,e\}$.
    \item Choices with no $d$: $3\cdot 2\cdot 4=24$.
    \item Choices with no $e$: $3\cdot 2\cdot 4=24$.
    \item Choices with neither $d$ nor $e$: $2\cdot 1\cdot 3=6$.
\end{itemize}
So the number using both $d$ and $e$ is
\[ 60-24-24+6=18. \]
Since there are $3$ choices for which of $d,e,f$ is blank, for each fixed choice of $a,b,c$ we get
\[ 3\cdot 18=54 \]
valid triples $(X_1,X_3,X_5)$.

Therefore the number of sequences satisfying both conditions is
\[ 6\cdot 5\cdot 4\cdot 54. \]
Hence
\[ p=\frac{6\cdot 5\cdot 4\cdot 54}{6\cdot 5\cdot 4\cdot 5\cdot 4\cdot 6} =\frac{54}{120} =\frac{9}{20}. \]
\end{excompare}

\begin{exverdict}{Annotator verdicts --- \texttt{aime\_2026\_9}}
\centering
\begin{tabular}{lcc}
    \toprule
    \textbf{Which solution is \dots} & \textbf{Annotator 1} & \textbf{Annotator 2} \\
    \midrule
    more verbose                   & Solution 2 & Solution 2 \\
    more elegant                   & Solution 2 & Solution 2 \\
    more insightful                & Solution 2 & Solution 2 \\
    more conceptually complex      & Solution 1 & tie \\
    more computationally demanding & Solution 2 & Solution 2 \\
    \bottomrule
\end{tabular}
\end{exverdict}

Four of the five dimensions receive identical decisive verdicts, and the single mismatch is a preference against a tie rather than a reversal. Regarding verbosity, Solution 2 spells out routine counting in itemized sub-steps and gives display equations for trivial arithmetic, and it opens with framing that is never used again. Solution 1, on the other hand, packs denser content into each sentence. The annotators therefore call Solution 2 more verbose despite the two solutions being nearly equal in length, which confirms that their notion of verbosity is text spent relative to content delivered rather than raw word count, as intended in our benchmark design.

\subsubsection{Verbosity as the Only Separating Metric} \label{app:example-ties}

Our third example is a pair of solutions to a geometry problem, produced by \geminiflash{} and \geminipro{}. The two follow the same computational route and differ almost entirely in how much text carries it. Since the second solution is twice the length of the first ($987$ versus $491$ words), we stack the two here rather than setting them side by side.

\begin{exproblem}{Problem \texttt{aime\_2025\_27}}
Let $A_1 A_2 A_3 \ldots A_{11}$ be an $11$-sided non-convex simple polygon such that, for every integer $2 \le i \le 10$, the area of $\triangle A_i A_{1} A_{i+1}$ is equal to $1$ and $\cos(\angle A_i A_{1} A_{i+1}) = \frac{12}{13}$, and the perimeter of the $11$-gon is equal to $20$. Then $A_1 A_2 + A_1 A_{11} = \frac{m \sqrt{n} - p}{q}$ where $m, n, p$, and $q$ are positive integers, $n$ is not divisible by the square of any prime, and no prime divides all of $m, p$, and $q$. Find $m + n + p + q$.
\end{exproblem}

\begin{exsolution}{gemini}{Solution 1 --- \geminiflash{}}
To solve the problem, let the lengths of the segments $A_1 A_i$ be $r_i$ for $i=2, 3, \dots, 11$. For each $i \in \{2, 3, \dots, 10\}$, we are given the area of $\triangle A_i A_1 A_{i+1}$ is $1$ and $\cos(\angle A_i A_1 A_{i+1}) = 12/13$. Let $\theta_i = \angle A_i A_1 A_{i+1}$. Since $0 < \theta_i < \pi$ and $\cos \theta_i = 12/13$, we have:
\[
\sin \theta_i = \sqrt{1 - \cos^2 \theta_i} = \sqrt{1 - \left(\frac{12}{13}\right)^2} = \sqrt{\frac{169-144}{169}} = \frac{5}{13}.
\]
The area of $\triangle A_i A_1 A_{i+1}$ is given by $\frac{1}{2} r_i r_{i+1} \sin \theta_i = 1$. Substituting $\sin \theta_i = 5/13$, we get:
\[
\frac{1}{2} r_i r_{i+1} \cdot \frac{5}{13} = 1 \implies r_i r_{i+1} = \frac{26}{5}.
\]
Let $k = 26/5$. For $i=2, 3, \dots, 10$, we have $r_i r_{i+1} = k$. This implies: $r_2 r_3 = k$, $r_3 r_4 = k \implies r_4 = r_2$, $r_4 r_5 = k \implies r_5 = r_3$, and so on. Thus, $r_{2j} = r_2$ and $r_{2j+1} = r_3 = k/r_2$. Let $x = r_2$ and $y = r_{11} = k/r_2$.

The sides of the $11$-gon are $s_1, s_2, \dots, s_{11}$, where $s_1 = A_1 A_2 = r_2 = x$ and $s_{11} = A_{11} A_1 = r_{11} = y$. For $i \in \{2, 3, \dots, 10\}$, the length $s_i = A_i A_{i+1}$ is found using the law of cosines in $\triangle A_i A_1 A_{i+1}$:
\[
s_i^2 = r_i^2 + r_{i+1}^2 - 2 r_i r_{i+1} \cos \theta_i = x^2 + y^2 - 2k \cdot \frac{12}{13}.
\]
Substituting $k = 26/5$ and $\cos \theta_i = 12/13$:
\[
s_i^2 = x^2 + y^2 - 2 \cdot \frac{26}{5} \cdot \frac{12}{13} = x^2 + y^2 - \frac{48}{5}.
\]
Let $s = s_i$ for $i=2, \dots, 10$. The perimeter is $x + 9s + y = 20$. Let $S = x + y$. Then $x^2 + y^2 = S^2 - 2xy = S^2 - 52/5$. Substituting this into the equation for $s^2$:
\[
s^2 = S^2 - \frac{52}{5} - \frac{48}{5} = S^2 - 20.
\]
The perimeter equation becomes $S + 9\sqrt{S^2 - 20} = 20$. Squaring both sides:
\[
81(S^2 - 20) = (20 - S)^2 \implies 81S^2 - 1620 = 400 - 40S + S^2 \implies 80S^2 + 40S - 2020 = 0.
\]
Dividing by $20$, we get $4S^2 + 2S - 101 = 0$. Solving for $S > 0$ using the quadratic formula:
\[
S = \frac{-2 + \sqrt{4 - 4(4)(-101)}}{8} = \frac{-2 + \sqrt{1620}}{8} = \frac{-2 + 18\sqrt{5}}{8} = \frac{9\sqrt{5} - 1}{4}.
\]
The expression for $A_1 A_2 + A_1 A_{11}$ is $x + y = S = \frac{9\sqrt{5} - 1}{4}$. The values of $m, n, p, q$ are $m=9, n=5, p=1, q=4$. We check: $n=5$ is square-free, $m, n, p, q$ are positive integers, and $\gcd(9, 1, 4) = 1$. The sum $m + n + p + q = 9 + 5 + 1 + 4 = 19$.

Final Answer: $\boxed{19}$
\end{exsolution}

\begin{exsolution}{gemini}{Solution 2 --- \geminipro{}}
To find the sum $A_1 A_2 + A_1 A_{11}$, we will first use the given information about the triangles formed by the vertices $A_1, A_i,$ and $A_{i+1}$ to find the lengths of the edges of the polygon.

\textbf{Step 1: Finding the lengths of the segments from $A_1$.}
Let $r_i = A_1 A_i$ denote the distance from $A_1$ to the vertex $A_i$ for $2 \le i \le 11$. For each integer $i$ from 2 to 10, consider the triangle $\triangle A_i A_1 A_{i+1}$. The area of a triangle can be expressed as:
\[
\text{Area} = \frac{1}{2} A_1 A_i \cdot A_1 A_{i+1} \sin(\angle A_i A_1 A_{i+1})
\]
We are given that $\cos(\angle A_i A_1 A_{i+1}) = \frac{12}{13}$. Because the internal angle of a triangle must be strictly between $0^\circ$ and $180^\circ$, its sine must be positive. Thus, we can find the sine using the Pythagorean identity:
\[
\begin{aligned}
\sin(\angle A_i A_1 A_{i+1}) &= \sqrt{1 - \cos^2(\angle A_i A_1 A_{i+1})} \\
&= \sqrt{1 - \left(\frac{12}{13}\right)^2} = \sqrt{\frac{169 - 144}{169}} = \frac{5}{13}
\end{aligned}
\]
We are also given that the area of each of these 9 triangles is $1$. Substituting the known values into the area formula gives:
\[
1 = \frac{1}{2} r_i r_{i+1} \left(\frac{5}{13}\right)
\]
Multiplying both sides by $\frac{26}{5}$ yields:
\[
r_i r_{i+1} = \frac{26}{5} \quad \text{for } i = 2, 3, \ldots, 10
\]
This equation gives us a system:
\[
r_2 r_3 = \frac{26}{5}, \quad r_3 r_4 = \frac{26}{5}, \quad r_4 r_5 = \frac{26}{5}, \dots
\]
From $r_2 r_3 = r_3 r_4$, and knowing $r_3 > 0$ (since lengths are positive), we can divide by $r_3$ to conclude that $r_2 = r_4$. By continuing this reasoning ($r_{i} r_{i+1} = r_{i+1} r_{i+2} \implies r_i = r_{i+2}$), the lengths alternate. Therefore, we have:
\[
r_2 = r_4 = r_6 = r_8 = r_{10} = x, \qquad r_3 = r_5 = r_7 = r_9 = r_{11} = y
\]
for some positive real numbers $x$ and $y$ such that $xy = \frac{26}{5}$. Notice that $A_1 A_2 = x$ and $A_1 A_{11} = y$. The quantity we want to find is $A_1 A_2 + A_1 A_{11} = x + y$. Let $S = x + y$.

\textbf{Step 2: Finding the lengths of the edges $A_i A_{i+1}$.}
For $2 \le i \le 10$, we can find the side length $A_i A_{i+1}$ using the Law of Cosines on $\triangle A_i A_1 A_{i+1}$:
\[
A_i A_{i+1}^2 = A_1 A_i^2 + A_1 A_{i+1}^2 - 2(A_1 A_i)(A_1 A_{i+1}) \cos(\angle A_i A_1 A_{i+1})
\]
Since one of $\{A_1 A_i, A_1 A_{i+1}\}$ is $x$ and the other is $y$, we have $A_1 A_i^2 + A_1 A_{i+1}^2 = x^2 + y^2$. Furthermore, their product is $xy = \frac{26}{5}$, and we are given the cosine of the angle is $\frac{12}{13}$. Thus:
\[
A_i A_{i+1}^2 = x^2 + y^2 - 2 \left(\frac{26}{5}\right) \left(\frac{12}{13}\right) = x^2 + y^2 - \frac{48}{5}
\]
We can express $x^2 + y^2$ in terms of $S = x + y$:
\[
x^2 + y^2 = (x+y)^2 - 2xy = S^2 - 2\left(\frac{26}{5}\right) = S^2 - \frac{52}{5}
\]
Substituting this back into our expression for the squared side length:
\[
A_i A_{i+1}^2 = \left(S^2 - \frac{52}{5}\right) - \frac{48}{5} = S^2 - \frac{100}{5} = S^2 - 20
\]
So, the length of each of these 9 edges is $d = \sqrt{S^2 - 20}$.

\textbf{Step 3: Utilizing the perimeter to solve for $S$.}
The perimeter of the 11-gon $A_1 A_2 \ldots A_{11}$ is the sum of the lengths of all 11 of its edges:
\[
\text{Perimeter} = A_1 A_2 + \left( \sum_{i=2}^{10} A_i A_{i+1} \right) + A_{11} A_1
\]
Substituting our defined variables ($A_1 A_2 = x$, $A_{11} A_1 = y$, and $A_i A_{i+1} = \sqrt{S^2 - 20}$):
\[
\text{Perimeter} = x + 9\sqrt{S^2 - 20} + y = (x + y) + 9\sqrt{S^2 - 20} = S + 9\sqrt{S^2 - 20}
\]
We are given that the perimeter is equal to 20. Thus, we establish the equation:
\[
S + 9\sqrt{S^2 - 20} = 20
\]
Isolating the square root gives $9\sqrt{S^2 - 20} = 20 - S$. Squaring both sides (which requires that $20 - S \ge 0$ for a valid solution):
\[
81(S^2 - 20) = (20 - S)^2 \implies 81S^2 - 1620 = 400 - 40S + S^2
\]
Subtracting the right side from the left to set the equation to zero:
\[
80S^2 + 40S - 2020 = 0
\]
Dividing by 20 leaves a simpler quadratic equation:
\[
4S^2 + 2S - 101 = 0
\]
Using the quadratic formula to solve for $S$:
\[
S = \frac{-2 \pm \sqrt{2^2 - 4(4)(-101)}}{2(4)} = \frac{-2 \pm \sqrt{4 + 1616}}{8} = \frac{-2 \pm \sqrt{1620}}{8}
\]
We simplify the radical, noting that $1620 = 324 \times 5 = 18^2 \times 5$, so $\sqrt{1620} = 18\sqrt{5}$:
\[
S = \frac{-2 \pm 18\sqrt{5}}{8} = \frac{-1 \pm 9\sqrt{5}}{4}
\]
Because $S = x + y$ must represent the sum of two positive lengths, $S$ must be strictly positive. Since $9\sqrt{5} > 1$, we take the positive root:
\[
S = \frac{9\sqrt{5} - 1}{4}
\]
\emph{(Note: $\sqrt{5} \approx 2.236 \implies S \approx 4.781$, confirming that $20 - S \ge 0$ is indeed satisfied, ensuring no extraneous solutions were generated by squaring).}

\textbf{Step 4: Formatting the solution.}
We have found that $A_1 A_2 + A_1 A_{11} = S = \frac{9\sqrt{5} - 1}{4}$. This is exactly in the required format $\frac{m\sqrt{n} - p}{q}$, giving:
\[
m = 9, \quad n = 5, \quad p = 1, \quad q = 4
\]
We verify the given conditions: $m, n, p, q$ are positive integers, $n=5$ is square-free, and no prime divides all of $m, p,$ and $q$ (since $p=1$). Finally, finding the required sum:
\[
m + n + p + q = 9 + 5 + 1 + 4 = 19
\]
\[\boxed{19}\]
\end{exsolution}

\begin{exverdict}{Annotator verdicts --- \texttt{aime\_2025\_27} (\geminiflash{} vs.\ \geminipro{})}
\centering
\begin{tabular}{lcc}
    \toprule
    \textbf{Which solution is \dots} & \textbf{Annotator 1} & \textbf{Annotator 2} \\
    \midrule
    more verbose                   & Solution 2 & Solution 2 \\
    more elegant                   & tie & tie \\
    more insightful                & tie & tie \\
    more conceptually complex      & tie & tie \\
    more computationally demanding & tie & tie \\
    \bottomrule
\end{tabular}
\end{exverdict}

Both solutions follow the same computational route: the area formula gives the constant product $r_i r_{i+1} = 26/5$, the products force the lengths to alternate, the Law of Cosines makes all nine remaining sides equal, and the perimeter condition yields the same quadratic in $S$. However, both annotators judge the longer solution more verbose, since it also supplies running commentary on the algebra it has just performed, and both return ties on all four remaining dimensions.

\subsubsection{Decisive Verdicts Alongside Agreed Ties} \label{app:example-geometry}

Finally, this example pairs \glmfive{} against \qwenthreeshort{} on a geometry problem, where both solutions belong to the same method family, coordinate bashing, and differ mainly in the coordinate system they choose, which makes a difference. The annotators agree on all five dimensions, with three decisive verdicts and two agreed ties.

\begin{exproblem}{Problem \texttt{imo-bench\_geometry-013}}
In triangle $XYZ$, we have $XY = YZ$ and $\angle Y = 20^\circ$. Point $P$ on $XZ$ is such that $XP : PZ = 1 : 2$, and point $Q$ is the projection of $Z$ onto $YP$. Find the angle $\angle XQY$ in degrees.
\end{exproblem}

\begin{excompare}{glm}{Solution 1 --- \glmfive{}}{qwen}{Solution 2 --- \qwenthreeshort{}}
\textbf{Proof.} We place the triangle in a coordinate system. Because angles are invariant under scaling, rotation and translation, we may set
\[
Y = (0,0),\; X = (1,0),\; Z = (\cos 20^\circ, \sin 20^\circ),
\]
so that $XY = YZ = 1$ and $\angle XYZ = 20^\circ$.

\textbf{1. Point $P$ on $XZ$.} Given $XP : PZ = 1 : 2$, the section formula gives
\[
P = \frac{2X + Z}{3} = \left( \frac{2 + \cos 20^\circ}{3},\; \frac{\sin 20^\circ}{3} \right).
\]

\textbf{2. Point $Q$, foot from $Z$ to $YP$.} Since $Y$ is the origin, line $YP$ is the line through the origin with direction vector $P$. The orthogonal projection of $Z$ onto this line is
\[
Q = \frac{Z \cdot P}{P \cdot P}\, P.
\]
Compute the dot products:
\[
\begin{aligned}
Z \cdot P &= \cos 20^\circ\cdot\frac{2+\cos 20^\circ}{3} + \sin 20^\circ\cdot\frac{\sin 20^\circ}{3} \\
          &= \frac{1}{3}\bigl(2\cos 20^\circ + \cos^2 20^\circ + \sin^2 20^\circ\bigr) \\
          &= \frac{1 + 2\cos 20^\circ}{3},
\end{aligned}
\]
\[
\begin{aligned}
P \cdot P &= \left(\frac{2+\cos 20^\circ}{3}\right)^2 + \left(\frac{\sin 20^\circ}{3}\right)^2 \\
          &= \frac{1}{9}\bigl((2+\cos 20^\circ)^2 + \sin^2 20^\circ\bigr) \\
          &= \frac{5 + 4\cos 20^\circ}{9}.
\end{aligned}
\]
Hence
\[
\begin{aligned} t = \frac{Z \cdot P}{P \cdot P} &= \frac{(1+2\cos 20^\circ)/3}{(5+4\cos 20^\circ)/9} \\ &= \frac{3(1+2\cos 20^\circ)}{5+4\cos 20^\circ}. \end{aligned}
\]
Thus
\[
\begin{aligned} Q = tP = \Bigl( &\frac{(1+2\cos 20^\circ)(2+\cos 20^\circ)}{5+4\cos 20^\circ}, \\ &\frac{(1+2\cos 20^\circ)\sin 20^\circ}{5+4\cos 20^\circ} \Bigr). \end{aligned}
\]
Denote $D = 5 + 4\cos 20^\circ$ for brevity.

\textbf{3. Cosine of $\angle XQY$.} Vectors from $Q$:
\[
\overrightarrow{QX} = (1 - Q_x,\; -Q_y), \qquad \overrightarrow{QY} = (-Q_x,\; -Q_y).
\]
Their dot product and lengths:
\[
\begin{aligned}
\overrightarrow{QX} \cdot \overrightarrow{QY} &= (1 - Q_x)(-Q_x) + (-Q_y)(-Q_y) \\
&= -Q_x + Q_x^2 + Q_y^2 = |Q|^2 - Q_x,
\end{aligned}
\]
and $|\overrightarrow{QY}| = |Q|$. Therefore
\[
\cos \angle XQY = \frac{|Q|^2 - Q_x}{|Q| \cdot |\overrightarrow{QX}|}.
\]
We compute $|Q|^2$ and $|\overrightarrow{QX}|^2$. First,
\[
\begin{aligned} |Q|^2 &= Q_x^2 + Q_y^2 \\ &= \frac{(1+2\cos 20^\circ)^2}{D^2} \\ &\quad \times \bigl((2+\cos 20^\circ)^2 + \sin^2 20^\circ\bigr). \end{aligned}
\]
But $(2+\cos 20^\circ)^2 + \sin^2 20^\circ = 5 + 4\cos 20^\circ = D$, so
\[
|Q|^2 = \frac{(1+2\cos 20^\circ)^2}{D}.
\]
Next, $|\overrightarrow{QX}|^2 = (1 - Q_x)^2 + Q_y^2 = 1 - 2Q_x + |Q|^2$. Substituting the expressions for $Q_x$ and $|Q|^2$:
\[
\begin{aligned}
&1 - 2Q_x + |Q|^2 \\
&= 1 - \frac{2(1+2\cos 20^\circ)(2+\cos 20^\circ)}{D} \\
&\quad + \frac{(1+2\cos 20^\circ)^2}{D}.
\end{aligned}
\]
The numerator simplifies:
\[
\begin{aligned}
&(1+2\cos 20^\circ)^2 - 2(1+2\cos 20^\circ)(2+\cos 20^\circ) \\
&= (1+2\cos 20^\circ) \\
&\quad \times \bigl[(1+2\cos 20^\circ) - 2(2+\cos 20^\circ)\bigr] \\
&= (1+2\cos 20^\circ)(-3).
\end{aligned}
\]
Hence
\[
\begin{aligned} 1 - 2Q_x + |Q|^2 &= 1 - \frac{3(1+2\cos 20^\circ)}{D} \\ &= \frac{D - 3(1+2\cos 20^\circ)}{D}. \end{aligned}
\]
With $D = 5 + 4\cos 20^\circ$,
\[
\begin{aligned}
D - 3(1+2\cos 20^\circ) &= (5+4\cos 20^\circ) \\
&\quad - 3 - 6\cos 20^\circ \\
&= 2(1-\cos 20^\circ).
\end{aligned}
\]
Therefore $|\overrightarrow{QX}|^2 = \frac{2(1-\cos 20^\circ)}{D}$.

\textbf{4. Simplification of $\cos \angle XQY$.} For the numerator,
\[
\begin{aligned}
|Q|^2 - Q_x &= \frac{(1+2\cos 20^\circ)^2}{D} \\
&\quad - \frac{(1+2\cos 20^\circ)(2+\cos 20^\circ)}{D} \\
&= \frac{1+2\cos 20^\circ}{D}\,(\cos 20^\circ - 1).
\end{aligned}
\]
For the denominator,
\[
\begin{aligned}
|Q| \cdot |\overrightarrow{QX}| &= \frac{1+2\cos 20^\circ}{\sqrt{D}} \cdot \sqrt{\frac{2(1-\cos 20^\circ)}{D}} \\
&= \frac{(1+2\cos 20^\circ)\sqrt{2(1-\cos 20^\circ)}}{D}.
\end{aligned}
\]
Thus
\[
\cos \angle XQY = \frac{\cos 20^\circ - 1}{\sqrt{2(1-\cos 20^\circ)}}.
\]
Because $\cos 20^\circ - 1 = -(1-\cos 20^\circ)$,
\[
\begin{aligned} \cos \angle XQY &= -\frac{1-\cos 20^\circ}{\sqrt{2(1-\cos 20^\circ)}} \\ &= -\sqrt{\frac{1-\cos 20^\circ}{2}}. \end{aligned}
\]

\textbf{5. Trigonometric identity.} Using $1 - \cos \theta = 2\sin^2(\theta/2)$ with $\theta = 20^\circ$:
\[
\sqrt{\frac{1-\cos 20^\circ}{2}} = \sqrt{\frac{2\sin^2 10^\circ}{2}} = \sin 10^\circ.
\]
Hence $\cos \angle XQY = -\sin 10^\circ$.

\textbf{6. Determination of the angle.} Since $\sin 10^\circ = \cos 80^\circ$, we have
\[
\begin{aligned} \cos \angle XQY &= -\cos 80^\circ \\ &= \cos(180^\circ - 80^\circ) = \cos 100^\circ. \end{aligned}
\]
The angle $\angle XQY$ is a geometric angle formed by two segments, so it lies in $(0^\circ, 180^\circ)$. Its cosine is negative, therefore the angle must be obtuse and uniquely
\[\boxed{\angle XQY = 100^\circ}\]

\exswitch

To find the measure of $\angle XQY$, we use a Cartesian coordinate system.

\textbf{1. Coordinate setup.} Let $M$ be the midpoint of $XZ$. Since $XY = YZ$, $YM$ is the perpendicular bisector of $XZ$. We place $M$ at the origin $(0,0)$ and $XZ$ along the $x$-axis.
\begin{itemize}[itemsep=0em,leftmargin=1.2em,topsep=0.2em]
    \item Let $X = (-a, 0)$ and $Z = (a, 0)$.
    \item Since $\triangle XYZ$ is isosceles with $\angle Y = 20^\circ$, the base angles are $\angle X = \angle Z = \frac{180^\circ - 20^\circ}{2} = 80^\circ$.
    \item The height $YM = a \tan 80^\circ = a \cot 10^\circ$. Thus, $Y = (0, a \cot 10^\circ)$.
    \item Point $P$ lies on $XZ$ such that $XP : PZ = 1 : 2$. Since $XZ = 2a$, $XP = \frac{2}{3}a$.
    \item The coordinate of $P$ is $x_P = -a + \frac{2}{3}a = -\frac{a}{3}$. So, $P = (-\frac{a}{3}, 0)$.
\end{itemize}

\textbf{2. Equations of lines $YP$ and $ZQ$.} The slope of line $YP$ is
\[
m_{YP} = \frac{a \cot 10^\circ - 0}{0 - (-a/3)} = 3 \cot 10^\circ.
\]
Point $Q$ is the projection of $Z$ onto $YP$, meaning $ZQ \perp YP$. The slope of $ZQ$ is
\[
m_{ZQ} = -\frac{1}{m_{YP}} = -\frac{\tan 10^\circ}{3}.
\]
The equation of line $ZQ$ passing through $Z(a, 0)$ is $y = -\frac{\tan 10^\circ}{3}(x - a)$, and the equation of line $YP$ is
\[
y = 3 \cot 10^\circ\left(x + \tfrac{a}{3}\right) = 3 \cot 10^\circ x + a \cot 10^\circ.
\]

\textbf{3. Coordinates of $Q$.} To find $Q(x_Q, y_Q)$, we solve the system:
\[
3 \cot 10^\circ x + a \cot 10^\circ = -\tfrac{\tan 10^\circ}{3}(x - a).
\]
Multiplying by $3 \tan 10^\circ$:
\[
\begin{gathered}
9x + 3a = -x \tan^2 10^\circ + a \tan^2 10^\circ \\
\implies x_Q(9 + \tan^2 10^\circ) = a(\tan^2 10^\circ - 3),
\end{gathered}
\]
so that
\[
x_Q = a \frac{\tan^2 10^\circ - 3}{9 + \tan^2 10^\circ}, \quad y_Q = \frac{4a \tan 10^\circ}{9 + \tan^2 10^\circ}.
\]

\textbf{4. Slopes and angle calculation.} We calculate the slopes of vectors $\vec{QX}$ and $\vec{QY}$. For the slope of $QX$, written $m_1$:
\[
\begin{aligned}
x_X - x_Q &= -a - a \frac{\tan^2 10^\circ - 3}{9 + \tan^2 10^\circ} \\
&= \frac{-2a(\tan^2 10^\circ + 3)}{9 + \tan^2 10^\circ},
\end{aligned}
\]
\[
\begin{aligned} m_1 = \frac{0 - y_Q}{x_X - x_Q} &= \frac{-4a \tan 10^\circ}{-2a(\tan^2 10^\circ + 3)} \\ &= \frac{2 \tan 10^\circ}{\tan^2 10^\circ + 3}. \end{aligned}
\]
For the slope of $QY$, written $m_2$: since $Q$ lies on $YP$, $m_2 = m_{YP} = 3 \cot 10^\circ$.

The tangent of the angle $\theta$ between the lines is
\[
\tan \theta = \left| \frac{m_2 - m_1}{1 + m_2 m_1} \right|.
\]
Substituting the values:
\[
\begin{aligned}
m_2 - m_1 &= \frac{3}{\tan 10^\circ} - \frac{2 \tan 10^\circ}{\tan^2 10^\circ + 3} \\
&= \frac{3(\tan^2 10^\circ + 3) - 2 \tan^2 10^\circ}{\tan 10^\circ(\tan^2 10^\circ + 3)} \\
&= \frac{\tan^2 10^\circ + 9}{\tan 10^\circ(\tan^2 10^\circ + 3)},
\end{aligned}
\]
\[
\begin{aligned}
1 + m_2 m_1 &= 1 + \frac{3}{\tan 10^\circ} \cdot \frac{2 \tan 10^\circ}{\tan^2 10^\circ + 3} \\
&= \frac{\tan^2 10^\circ + 9}{\tan^2 10^\circ + 3},
\end{aligned}
\]
so that
\[
\tan \theta = \frac{1}{\tan 10^\circ} = \cot 10^\circ = \tan 80^\circ.
\]
The acute angle between the lines is $80^\circ$. To determine the interior angle $\angle XQY$, we check the vector directions. $\vec{QY}$ points in the first quadrant (up-right), while $\vec{QX}$ points in the third quadrant (down-left). The angle between these directions is $180^\circ - 80^\circ = 100^\circ$.
\[\boxed{100^\circ}\]
\end{excompare}

\begin{exverdict}{Annotator verdicts --- \texttt{imo-bench\_geometry-013}}
\centering
\begin{tabular}{lcc}
    \toprule
    \textbf{Which solution is \dots} & \textbf{Annotator 1} & \textbf{Annotator 2} \\
    \midrule
    more verbose                   & Solution 1 & Solution 1 \\
    more elegant                   & Solution 2 & Solution 2 \\
    more insightful                & tie & tie \\
    more conceptually complex      & tie & tie \\
    more computationally demanding & Solution 1 & Solution 1 \\
    \bottomrule
\end{tabular}
\end{exverdict}

We observe $5/5$ agreement with three decisive verdicts and two agreed ties. Both annotators independently record that the two solutions are comparable on insightfulness and conceptual complexity, which reflects the fact that both take a heavily computational route and neither contributes a genuine idea beyond it. They also agree that Solution 1 is the more verbose and the more computationally demanding, and that Solution 2 is the more elegant because it spares some of the heavy computation. All three decisive verdicts trace back to the choice of coordinate system. Placing $Y$ at the origin, as Solution 1 does, forces further computation and explanation downstream, whereas Solution 2's use of the midpoint of $XZ$ exploits the symmetry of the isosceles triangle from the start. This highlights that our annotators were able to recognize that the dimensions that depend on the argument tie, and the dimensions that depend on execution should be separate.

\subsection{Contentious Comparisons} \label{app:examples-contentious}

However, some solution pairs cannot be easily separated. We now present four comparisons on which agreement breaks down. The first three split the two human annotators, one on each of the three dimensions that \bench{} reports, and the fourth is a pair on which the annotators agree on a tie but the LLM judge does not. These are the cases that motivate our design choices, and we include them to make concrete the failure modes summarized in \cref{app:human-validation} and \cref{app:challenges}.

\subsubsection{Conciseness: Verbose in Different Ways} \label{app:example-conciseness}

Conciseness is the one dimension where inter-annotator disagreement is most pronounced (\cref{tab:human_agreement}). The following pair, solved by \glmfive{} and \geminipro{}, illustrates how, when the two solutions take genuinely different routes to the same answer and each is redundant in a different place, which one is ``more verbose'' depends on which kind of redundancy the reader weighs more heavily.

\begin{exproblem}{Problem \texttt{imo-bench\_number\_theory-042}}
Find all positive integers $k$ such that the number of positive divisors of $2k^3$ is $2k$.
\end{exproblem}

\begin{excompare}{glm}{Solution 1 --- \glmfive{}}{gemini}{Solution 2 --- \geminipro{}}
We are to find all positive integers $k$ such that the number of positive divisors of $2k^3$ equals $2k$. Let $d(n)$ denote the number of positive divisors of $n$. If $n=\prod p_i^{e_i}$ then $d(n)=\prod (e_i+1)$.

\textbf{Case I: $k$ odd.}
If $k=1$, then $2k^3=2$ and $d(2)=2=2k$, so $k=1$ works.

Assume $k>1$ odd. Write its prime factorization
\[
k=\prod_{i=1}^{r} p_i^{a_i},\qquad p_i\;\text{odd primes},\;a_i\ge 1.
\]
Then
\[
k^3=\prod p_i^{3a_i},\qquad 2k^3=2\prod p_i^{3a_i},
\]
so
\[
d(2k^3)=(1+1)\prod (3a_i+1)=2\prod (3a_i+1).
\]
Setting $d(2k^3)=2k$ gives
\[
\begin{gathered} 2\prod (3a_i+1)=2k \\ \Longrightarrow\; \prod (3a_i+1)=k=\prod p_i^{a_i}. \end{gathered} \tag{1}
\]

We show that (1) has no solution with $k>1$.

For each $i$ define $r_i=(3a_i+1)/p_i^{a_i}$. Their product is $1$, so at least one $r_i\ge 1$ (otherwise product $<1$). Moreover $r_i=1$ would imply $3a_i+1=p_i^{a_i}$, which is impossible: for $a_i=1$ we get $4=p_i$ (contradiction because $p_i$ is odd $\ge 3$); for $a_i\ge 2$ the left side is at most $3a_i+1$ while the right side is at least $p_i^2\ge 9$, again impossible. Hence some $r_i>1$.

Now we prove a useful inequality.

\textbf{Lemma.} For any odd prime $p$ and integer $a\ge 1$,
\[
3a+1 \le \frac{4}{3}\,p^a,
\]
with equality iff $p=3$ and $a=1$.

\emph{Proof.} If $p=3$ the claim is $3(3a+1)\le 4\cdot 3^a$. For $a=1$ we have equality. For $a\ge 2$ the right side grows exponentially while the left side grows linearly, so the inequality holds. If $p\ge 5$ then $p^a\ge 5^a>3^a$, so
\[
\frac{3a+1}{p^a}<\frac{3a+1}{3^a}\le\frac{4}{3},
\]
where the last inequality follows from the $p=3$ case. $\blacksquare$

From the lemma, the only way to have $r_i>1$ is $p_i=3$ and $a_i=1$, giving $r_i=4/3$. Therefore any solution of (1) with $k>1$ must contain the prime $3$ with exponent exactly $1$. Then the factor $(3\cdot1+1)=4$ appears in the left product, making $\prod(3a_i+1)$ even. But the right-hand side $k$ is odd (product of odd primes). This contradiction shows that no odd $k>1$ satisfies (1). Hence the only odd solution is $k=1$.

\textbf{Case II: $k$ even.}
Write $k=2^a\cdot m$ with $a\ge 1$ and $m$ odd. Let
\[
m=\prod_{i=1}^{s} q_i^{b_i},\qquad q_i\;\text{odd primes},\;b_i\ge 1.
\]
Then
\[
k^3=2^{3a}\prod q_i^{3b_i},\qquad 2k^3=2^{3a+1}\prod q_i^{3b_i},
\]
so
\[
d(2k^3)=(3a+2)\prod (3b_i+1).
\]
Setting equal to $2k=2^{a+1}\prod q_i^{b_i}$ yields
\[
(3a+2)\prod (3b_i+1)=2^{a+1}\prod q_i^{b_i}. \tag{2}
\]
Rewrite (2) as
\[
\prod_{i=1}^{s}\frac{3b_i+1}{q_i^{b_i}}=\frac{2^{a+1}}{3a+2}. \tag{3}
\]
Denote the left side by $P$. By the lemma, each factor satisfies
\[
\frac{3b_i+1}{q_i^{b_i}}\le\frac{4}{3},
\]
with equality only for $(q_i,b_i)=(3,1)$. Moreover, the only factor that can exceed $1$ is the one for $(3,1)$ (which equals $4/3$); all other factors are $<1$. Consequently $P\le\frac{4}{3}$, with equality possible only if the set of odd primes consists exactly of a single factor $3$ with exponent $1$. If there are no odd primes ($m=1$), then $P=1$.

Now examine the right side of (3). If $a\ge 3$, then
\[
\frac{2^{a+1}}{3a+2}\ge\frac{2^{4}}{11}=\frac{16}{11}>\frac{4}{3}.
\]
(For larger $a$ the fraction increases because the numerator grows exponentially while the denominator grows linearly.) Hence we would need $P>4/3$, impossible. Therefore $a$ cannot be $\ge 3$. Thus only $a=1$ and $a=2$ remain.

\textbf{Subcase $a=2$.}
Then $3a+2=8$ and $2^{a+1}=8$, so (2) simplifies to
\[
\prod (3b_i+1)=\prod q_i^{b_i}=m.
\]
This is exactly equation (1) for the odd integer $m$. By the same reasoning as in Case I (or by the lemma applied to the product), the only possibility is $m=1$. Hence $k=2^2\cdot1=4$. Check: $2\cdot4^3=128$, $d(128)=8=2\cdot4$.

\textbf{Subcase $a=1$.}
Now $3a+2=5$ and $2^{a+1}=4$. Equation (2) becomes
\[
5\prod (3b_i+1)=4\prod q_i^{b_i}. \tag{4}
\]
Since the left side is divisible by $5$, the right side must be divisible by $5$. Because $\gcd(5,4)=1$, the product $\prod q_i^{b_i}$ must contain the prime $5$. Let the exponent of $5$ in $m$ be $e\ge 1$. Write $m=5^e\cdot m'$ with $5\nmid m'$. Then (4) rewrites as
\[
5\cdot(3e+1)\prod_{q_i\neq5}(3b_i+1)=4\cdot5^e\prod_{q_i\neq5}q_i^{b_i}.
\]
Dividing by $5$ gives
\[
(3e+1)\prod_{q_i\neq5}(3b_i+1)=4\cdot5^{e-1}\prod_{q_i\neq5}q_i^{b_i}. \tag{5}
\]
Take the product of ratios for the primes different from $5$:
\[
Q:=\prod_{q_i\neq5}\frac{3b_i+1}{q_i^{b_i}}=\frac{4\cdot5^{e-1}}{3e+1}. \tag{6}
\]
Again, by the lemma each factor in $Q$ is at most $4/3$, and as argued earlier, $Q\le 4/3$. Thus
\[
\frac{4\cdot5^{e-1}}{3e+1}\le\frac{4}{3}\quad\Longrightarrow\quad 5^{e-1}\le\frac{3e+1}{3}. \tag{7}
\]
For $e=1$: $5^{0}=1\le\frac{4}{3}$ holds. For $e=2$: $5\le\frac{7}{3}\approx2.33$ is false. For $e\ge3$ the left side grows exponentially while the right side grows linearly, so (7) fails. Hence $e=1$ is forced.

With $e=1$, (6) gives $Q=\frac{4\cdot5^{0}}{3\cdot1+1}=1$. Thus
\[
\begin{gathered} \prod_{q_i\neq5}\frac{3b_i+1}{q_i^{b_i}}=1 \\ \Longrightarrow\; \prod_{q_i\neq5}(3b_i+1)=\prod_{q_i\neq5}q_i^{b_i}. \end{gathered}
\]
This is again an equation of type (1) for the odd integer $m'$ (the part of $m$ not containing $5$). As before, the only possibility is that there are no such primes, i.e. $m'=1$. Therefore $m=5^1=5$ and $k=2^1\cdot5=10$. Check: $2\cdot10^3=2000=2^4\cdot5^3$, $d(2000)=5\cdot4=20=2\cdot10$.

All possibilities have been examined. The positive integers $k$ satisfying $d(2k^3)=2k$ are exactly
\[\boxed{1,\ 4,\ 10}\]

\exswitch

To find all positive integers $k$ such that the number of positive divisors of $2k^3$ is $2k$, we start by using the Fundamental Theorem of Arithmetic. We can express the prime factorization of $k$ as
\[ k = 2^a p_1^{a_1} p_2^{a_2} \cdots p_m^{a_m} \]
where $a \ge 0$ is a non-negative integer, $m \ge 0$ is the number of distinct odd prime factors, $p_1, p_2, \dots, p_m$ are distinct odd primes, and $a_1, a_2, \dots, a_m \ge 1$ are positive integers.

Using this representation, we can write $2k^3$ as:
\[ 2k^3 = 2 \left( 2^a p_1^{a_1} \cdots p_m^{a_m} \right)^3 = 2^{3a+1} p_1^{3a_1} \cdots p_m^{3a_m} \]
The number of positive divisors of a positive integer $N = p_1^{e_1} \cdots p_k^{e_k}$ is given by the well-known divisor function formula $d(N) = (e_1+1) \cdots (e_k+1)$. Applying this to $2k^3$, we get:
\[ d(2k^3) = (3a+2)(3a_1+1)(3a_2+1) \cdots (3a_m+1) \]
We are given that $d(2k^3) = 2k$. Substituting $k$ and $d(2k^3)$ into this equation gives:
\[ \begin{gathered} (3a+2) \prod_{i=1}^m (3a_i+1) = 2^{a+1} \prod_{i=1}^m p_i^{a_i} \\ \text{--- (Equation 1)} \end{gathered} \]

\textbf{Step 1: Exclusion of the prime factor 3.}
Assume for the sake of contradiction that one of the odd prime factors of $k$ is $3$ (say, $p_1 = 3$). Because $a_1 \ge 1$, the right-hand side of Equation 1 is divisible by $3^{a_1}$, making it a multiple of $3$. This implies that the left-hand side, $(3a+2) \prod_{i=1}^m (3a_i+1)$, must also be a multiple of $3$. However, let us examine the factors on the left-hand side modulo $3$:
\begin{itemize}[itemsep=0em,leftmargin=2em,topsep=0.2em]
    \item $3a+2 \equiv 2 \pmod 3$, so it is not divisible by $3$.
    \item For each $i$ such that $1 \le i \le m$, $3a_i+1 \equiv 1 \pmod 3$, so none of these terms are divisible by $3$.
\end{itemize}
Since $3$ is a prime number, a product of integers that are not divisible by $3$ cannot be divisible by $3$. Therefore, the left-hand side is not divisible by $3$, establishing a direct contradiction. Our assumption must be false, meaning $k$ cannot have $3$ as a prime factor. Thus, all odd prime factors of $k$ must be strictly greater than $3$ (i.e., $p_i \ge 5$ for all $1 \le i \le m$).

\textbf{Step 2: Bounding the components.}
We can rearrange Equation 1 by grouping the factors corresponding to the powers of $2$ and the odd primes:
\[ \frac{3a+2}{2^{a+1}} \prod_{i=1}^m \frac{3a_i+1}{p_i^{a_i}} = 1 \]
Let us define $g(a) = \frac{3a+2}{2^{a+1}}$ for $a \ge 0$, and $f(p, x) = \frac{3x+1}{p^x}$ for primes $p \ge 5$ and integers $x \ge 1$. The equation becomes:
\[ g(a) \prod_{i=1}^m f(p_i, a_i) = 1 \quad \text{--- (Equation 2)} \]

Let us find the absolute maximum value of $f(p, x)$ for $p \ge 5$ and $x \ge 1$. Since $p \ge 5$, we know $p^x \ge 5^x$, which gives $f(p, x) \le \frac{3x+1}{5^x}$.
\begin{itemize}[itemsep=0em,leftmargin=2em,topsep=0.2em]
    \item For $x=1$, $\frac{3(1)+1}{5^1} = \frac{4}{5} = 0.8$.
    \item To see that this is the maximum, we check the ratio of consecutive terms for $x \ge 1$:
    \[ \frac{\frac{3(x+1)+1}{5^{x+1}}}{\frac{3x+1}{5^x}} = \frac{3x+4}{5(3x+1)} = \frac{3x+4}{15x+5} \]
    For $x \ge 1$, it is evident that $15x+5 > 3x+4$, meaning the ratio is strictly less than $1$. Thus, the sequence strictly decreases, and the maximum is securely achieved at $x=1$.
\end{itemize}
Therefore, $f(p, x) \le \frac{4}{5}$, with equality if and only if $p=5$ and $x=1$.

Next, let us analyze $g(a)$ for integers $a \ge 0$:
\[ \begin{gathered} g(0) = \frac{2}{2} = 1, \quad g(1) = \frac{5}{4} = 1.25, \\ g(2) = \frac{8}{8} = 1. \end{gathered} \]
For $a \ge 1$, let us look at the difference between consecutive terms:
\[ \begin{aligned} g(a+1) - g(a) &= \frac{3(a+1)+2}{2^{a+2}} - \frac{3a+2}{2^{a+1}} \\ &= \frac{3a+5 - 2(3a+2)}{2^{a+2}} = \frac{1 - 3a}{2^{a+2}} \end{aligned} \]
Because $1 - 3a < 0$ for all $a \ge 1$, $g(a)$ strictly decreases for $a \ge 1$. Thus, the absolute maximum of $g(a)$ is $g(1) = \frac{5}{4}$.

\textbf{Step 3: Case analysis to find all solutions.}
We evaluate Equation 2 based on $m$, the number of distinct odd prime factors of $k$.

\emph{Case 1: $m \ge 2$.} Using our maximum bounds, the product of the sequence terms is at most:
\[ \begin{aligned} g(a) \prod_{i=1}^m f(p_i, a_i) &\le \frac{5}{4} \left( \frac{4}{5} \right)^m \\ &= \frac{5}{4} \cdot \frac{16}{25} \left( \frac{4}{5} \right)^{m-2} \\ &= \frac{4}{5} \left( \frac{4}{5} \right)^{m-2} \end{aligned} \]
Because $m \ge 2$, the maximum possible value is $\frac{4}{5} < 1$. So no solution can exist for $m \ge 2$.

\emph{Case 2: $m = 1$.} Equation 2 simplifies to $g(a) f(p_1, a_1) = 1$. Since $f(p_1, a_1) \le \frac{4}{5}$, we must have $g(a) \ge \frac{5}{4}$. The only evaluated integer $a$ fulfilling this condition is $a=1$, dictating exactly $g(1) = \frac{5}{4}$. To perfectly satisfy the equation, we are forced to have $f(p_1, a_1) = \frac{4}{5}$. As established earlier, this equality holds true if and only if $p_1=5$ and $a_1=1$. This provides the solution $k = 2^1 \cdot 5^1 = 10$.

\emph{Case 3: $m = 0$.} This reflects $k$ having no odd prime factors, so $k = 2^a$. The empty product defaults to $1$, leaving Equation 2 as $g(a) = 1$. Looking at the behavior of $g(a)$: $g(0) = 1 \implies a = 0 \implies k = 2^0 = 1$; $g(1) = 1.25 \neq 1$; $g(2) = 1 \implies a = 2 \implies k = 2^2 = 4$. Since $g(a)$ is strictly decreasing for $a \ge 2$, $g(a) < 1$ for all $a \ge 3$, leaving no other solutions. This provides the solutions $k = 1$ and $k = 4$.

\textbf{Verification:}
\begin{itemize}[itemsep=0em,leftmargin=2em,topsep=0.2em]
    \item For $k=1$: $2k^3 = 2$. Divisors of $2$ are $1, 2$, giving $d(2) = 2$. And $2k = 2(1) = 2$.
    \item For $k=4$: $2k^3 = 2(64) = 128 = 2^7$. The number of divisors of $2^7$ is $7+1 = 8$. And $2k = 2(4) = 8$.
    \item For $k=10$: $2k^3 = 2(1000) = 2000 = 2^4 \cdot 5^3$. The number of divisors is $(4+1)(3+1) = 20$. And $2k = 2(10) = 20$.
\end{itemize}
All solutions hold perfectly without error.
\[\boxed{1, 4, 10}\]
\end{excompare}

\begin{exverdict}{Annotator verdicts --- \texttt{imo-bench\_number\_theory-042}}
\centering
\begin{tabular}{lcc}
    \toprule
    \textbf{Which solution is \dots} & \textbf{Annotator 1} & \textbf{Annotator 2} \\
    \midrule
    more verbose                   & Solution 1 & Solution 2 \\
    more elegant                   & Solution 2 & Solution 2 \\
    more insightful                & tie        & Solution 2 \\
    more conceptually complex      & tie        & Solution 2 \\
    more computationally demanding & Solution 1 & Solution 1 \\
    \bottomrule
\end{tabular}
\end{exverdict}

\paragraph{Where the disagreement lies.} The annotators split most on verbosity. Solution 1 is clear and offers few obvious places to cut, but it restates the problem, carries commentary that adds nothing to the argument (``Their product is $1$, so at least one $r_i \ge 1$ (otherwise product $<1$). Moreover $r_i=1$ would imply \dots{} Hence some $r_i>1$.''), and renames the same quantity $\prod (3\cdot+1)/\text{prime}^{\cdot}$ three times, as $r_i$ in Case I, as $P$ in Case II, and as $Q$ in the subcase $a=1$. In Solution 2 the redundancy is instead concentrated in framing. The same appeal to unique factorization plus the divisor-count formula that Solution 1 disposes of in two lines is surrounded by filler sentences (``Let us find the absolute maximum value \dots{}'', ``the maximum is securely achieved'', ``All solutions hold perfectly without error''). Which of these weighs more is a subjective call, and is why our automatic conciseness metric replaces this judgment with an operational definition, measuring how much text can be cut while preserving the mathematical argument, rather than asking a reader where the text feels padded.

\subsubsection{Cognitive Simplicity: Formalism Against Abstraction} \label{app:example-cognitive}

The second contentious pair is on the same problem as \cref{app:example-shortlist}, but with \glmfive{} and \gptfivefour{} in place of the earlier models. Where that pair separated on every dimension, this one splits annotators on cognitive simplicity, as the two solutions make their argument hard in different ways.

\begin{exproblem}{Problem \texttt{shortlist\_2025\_6}}
Jerry places at most one rook in each cell of a $2025 \times 2025$ grid of cells. A rook attacks another rook if the two rooks are in the same row or column and there are no other rooks between them. Determine the maximum number of rooks Jerry can place on the grid such that no rook attacks $4$ other rooks.
\end{exproblem}

\begin{excompare}{glm}{Solution 1 --- \glmfive{}}{oai}{Solution 2 --- \gptfivefour{}}
We are given a $2025 \times 2025$ grid. A rook attacks another rook if they are in the same row or column and there is no other rook between them. For a rook, the number of rooks it attacks is the sum of the number of nearest rooks to the left and right in its row, plus the number of nearest rooks above and below in its column. Thus a rook can attack at most four others. The problem asks for the maximum number of rooks that can be placed (at most one per cell) such that no rook attacks four others, i.e., every rook attacks at most three others.

\textbf{Reformulation.} For a given placement, consider a rook at $(i,j)$.
\begin{itemize}[itemsep=0em,leftmargin=1.2em,topsep=0.2em]
    \item In its row $i$, let the rooks be at columns $c_1<c_2<\dots<c_m$. The rook at column $c_k$ attacks the rooks at $c_{k-1}$ (if $k>1$) and $c_{k+1}$ (if $k<m$). Thus it attacks two rooks in its row exactly when it is not the leftmost nor the rightmost in that row.
    \item Similarly, in its column $j$, it attacks two rooks exactly when it is not the topmost nor the bottommost in that column.
\end{itemize}
Consequently, a rook attacks four others precisely when it is \textbf{not} the leftmost nor the rightmost in its row \textbf{and} not the topmost nor the bottommost in its column. The condition ``no rook attacks four others'' is therefore equivalent to: \emph{every rook is either a leftmost or rightmost rook in its row, or a topmost or bottommost rook in its column (or both)}. We call such rooks \emph{protected} in at least one direction.

\textbf{Notation.} Let $n=2025\ge 2$. Define the set of \emph{border cells} $B$ as those cells with row $=1$ or row $=n$ or column $=1$ or column $=n$. The number of border cells is $|B|=2n+2(n-2)=4n-4$. Define the set of \emph{interior cells} $I$ as the remaining cells: rows $2,\dots,n-1$ and columns $2,\dots,n-1$. $|I|=(n-2)^2$.

For a placement $S$ (set of occupied cells), write $X=S\cap I$ (interior rooks) and $Y=S\cap B$ (border rooks). We aim to prove $|S|\le 4n-4$.

\textbf{Key Lemma.} In any admissible placement, there is an injective mapping from $X$ to $B\setminus S$ (empty border cells). Consequently $|X|\le |B\setminus S|$.

\emph{Proof of Lemma.} Take any interior rook $r=(i,j)\in X$ ($2\le i\le n-1$, $2\le j\le n-1$). By the condition, $r$ is either a row-extreme (leftmost or rightmost in its row) or a column-extreme (topmost or bottommost in its column). We assign an empty border cell $f(r)\in B\setminus S$ according to the following rule:
\begin{itemize}[itemsep=0em,leftmargin=1.2em,topsep=0.2em]
    \item If $r$ is a row-extreme (i.e., leftmost or rightmost in row $i$), then: if it is the leftmost rook in row $i$, let $f(r)=(i,1)$; if it is the rightmost rook in row $i$, let $f(r)=(i,n)$. (If it is the only rook in its row, it is both leftmost and rightmost; we may choose e.g.\ the leftmost assignment.)

    Because $j\ge 2$, $j\neq 1$; if $r$ is leftmost, there is no rook in row $i$ with column $<j$, hence $(i,1)$ is empty. Similarly, if $r$ is rightmost, $(i,n)$ is empty.
    \item Otherwise, $r$ is not a row-extreme, so it must be a column-extreme (topmost or bottommost in column $j$). Then: if it is the topmost rook in column $j$, let $f(r)=(1,j)$; if it is the bottommost rook in column $j$, let $f(r)=(n,j)$. (If it is the only rook in its column, we may choose the topmost assignment.)

    Because $i\ge 2$, $i\neq 1$; if $r$ is topmost, there is no rook in column $j$ with row $<i$, so $(1,j)$ is empty. Similarly, bottommost forces $(n,j)$ empty.
\end{itemize}
Now we show that $f$ is injective.
\begin{itemize}[itemsep=0em,leftmargin=1.2em,topsep=0.2em]
    \item If two different interior rooks both fall under the row-extreme case, their images are of the form $(i,1)$ or $(i,n)$ with $i$ being their respective rows. Since rows are distinct for rooks in different rows, and for two rooks in the same row one would be assigned $(i,1)$ and the other $(i,n)$, the images are distinct.
    \item If both fall under the column-extreme case, their images are $(1,j)$ or $(n,j)$ with $j$ being their columns. Columns are distinct for rooks in different columns, and for two rooks in the same column one gets $(1,j)$ and the other $(n,j)$, so images are distinct.
    \item If one rook is assigned via the row-extreme case and the other via the column-extreme case, their images belong to disjoint sets: row-extreme images lie on the left or right border with row $i\in\{2,\dots,n-1\}$ (so column $1$ or $n$), while column-extreme images lie on the top or bottom border with column $j\in\{2,\dots,n-1\}$ (so row $1$ or $n$). The only possible overlap would be at the four corners, but none of these images is a corner because $i\notin\{1,n\}$ in the first case and $j\notin\{1,n\}$ in the second. Hence the images are distinct.
\end{itemize}
Thus $f$ is injective, and each $f(r)$ is indeed an empty border cell. Therefore $|X|\le |B\setminus S|$. $\blacksquare$

\textbf{Upper bound.} We have
\[
\begin{aligned} |S| &= |Y| + |X| = (|B| - |B\setminus S|) + |X| \\ &\le |B| = 4n-4. \end{aligned}
\]
Hence any admissible placement contains at most $4n-4$ rooks.

\textbf{Construction achieving the bound.} Place rooks on every border cell. This uses exactly $4n-4$ rooks. We verify that no rook attacks four others.
\begin{itemize}[itemsep=0em,leftmargin=1.2em,topsep=0.2em]
    \item A rook on the top row (row $1$): its row is fully occupied, so it has both a left and a right neighbour, giving $2$ attacks from the row. If it is not a corner (i.e., column $j$ with $2\le j\le n-1$), its column contains rooks only at rows $1$ and $n$ (since interior rows have no rook in that column), so it has exactly one neighbour in the column, the rook at $(n,j)$, giving $1$ attack. Total $3$. If it is a corner, say $(1,1)$, it is leftmost in its row (only right neighbour) and topmost in its column (only down neighbour), giving $2$ attacks.
    \item A rook on the bottom row is symmetric.
    \item A rook on the left border (column $1$) with $2\le i\le n-1$: its row $i$ contains exactly two rooks, at columns $1$ and $n$ (since the row is not top or bottom). Thus it is leftmost, so it has one neighbour in the row, the rook at $(i,n)$, giving $1$ attack. Its column $1$ is fully occupied, so it has both an up and a down neighbour, giving $2$ attacks. Total $3$.
    \item A rook on the right border is symmetric.
\end{itemize}
Thus every rook attacks at most three others, satisfying the condition. Therefore the maximum number of rooks for $n=2025$ is $4\cdot2025-4 = 8096$.
\[\boxed{8096}\]

\exswitch

\[
\textbf{Claim: the maximum number of rooks is }8096.
\]
Let $m$ be the number of rooks in a valid arrangement. Also let $r$ be the number of rows containing at least one rook and $c$ the number of columns containing at least one rook. For each rook $X$, define $h(X)$ to be the number of rooks attacked by $X$ in its row and $v(X)$ the number attacked in its column.

Since a rook can attack at most one rook to the left and at most one rook to the right, we have $h(X)\in\{0,1,2\}$. Similarly $v(X)\in\{0,1,2\}$. The condition in the problem says that no rook attacks $4$ others, so
\[
h(X)+v(X)\le 3 \quad\text{for every rook }X.
\]

\textbf{Step 1: Sum of horizontal and vertical attacks.} Consider one occupied row containing $k$ rooks. If we list those $k$ rooks from left to right, then the only attacking pairs in that row are the $k-1$ consecutive pairs. Each such pair contributes $1$ to the horizontal attack count of each of its two rooks, so the total horizontal contribution of this row is $2(k-1)$. Summing over all occupied rows,
\[
\sum_X h(X)=2(m-r).
\]
Exactly the same argument for columns gives
\[
\sum_X v(X)=2(m-c).
\]

\textbf{Step 2: Define a ``defect''.} For each rook $X$, define $\delta(X)=3-h(X)-v(X)$. Because $h(X)+v(X)\le 3$, we have $\delta(X)\ge 0$. Now sum over all rooks:
\[
\begin{aligned}
\sum_X \delta(X) &= \sum_X \bigl(3-h(X)-v(X)\bigr) \\
&= 3m-\sum_X h(X)-\sum_X v(X).
\end{aligned}
\]
Substituting the formulas from Step 1,
\[
\begin{aligned} \sum_X \delta(X) &= 3m-2(m-r)-2(m-c) \\ &= 2r+2c-m. \end{aligned}
\]
So
\[
\sum_X \delta(X)=2r+2c-m. \tag{1}
\]

\textbf{Step 3: The top and bottom occupied rows force at least $4$ total defect.} If all rooks lie in a single row, then $m\le 2025<8096$, so such an arrangement cannot beat $8096$. Thus, for the upper bound, we may assume there are at least two occupied rows.

Let $T$ be the top occupied row and $B$ the bottom occupied row. These are distinct. Take a rook $X$ in the top occupied row $T$. There is no rook above it, so in its column it can attack at most one rook. Hence $v(X)\le 1$, and therefore
\[
\delta(X)=3-h(X)-v(X)\ge 2-h(X).
\]
Summing over all rooks $X$ in row $T$,
\[
\sum_{X\in T}\delta(X)\ge \sum_{X\in T}(2-h(X)).
\]
Suppose row $T$ contains $k$ rooks. As above, the total horizontal attack count in that row is $2(k-1)$. Hence
\[
\sum_{X\in T}(2-h(X)) =2k-2(k-1)=2,
\]
so $\sum_{X\in T}\delta(X)\ge 2$. By exactly the same argument, the bottom occupied row $B$ also contributes at least $2$:
\[
\sum_{X\in B}\delta(X)\ge 2.
\]
Since $T$ and $B$ are different rows, these contributions come from different rooks, so altogether $\sum_X \delta(X)\ge 4$. Using (1),
\[
2r+2c-m\ge 4,
\]
hence $m\le 2r+2c-4$. Finally, since $r\le 2025$ and $c\le 2025$,
\[
m\le 2\cdot 2025+2\cdot 2025-4=8096.
\]
So no valid arrangement can have more than $8096$ rooks.

\textbf{Step 4: Construction with $8096$ rooks.} Place rooks in every boundary cell of the $2025\times 2025$ grid. The number of boundary cells is
\[
4\cdot 2025-4=8096.
\]
Now check the condition:
\begin{itemize}[itemsep=0em,leftmargin=1.2em,topsep=0.2em]
    \item A corner rook attacks exactly $2$ rooks: one along its row and one along its column.
    \item A non-corner rook on the top or bottom edge attacks $3$ rooks: the nearest rook to its left, the nearest rook to its right, and the rook on the opposite edge in its column.
    \item A non-corner rook on the left or right edge also attacks $3$ rooks: the nearest rook above, the nearest rook below, and the rook on the opposite edge in its row.
\end{itemize}
Thus no rook attacks $4$ others. So $8096$ rooks are achievable. Therefore the maximum possible number is
\[\boxed{8096}\]
\end{excompare}

\begin{exverdict}{Annotator verdicts --- \texttt{shortlist\_2025\_6} (\glmfive{} vs.\ \gptfivefour{})}
\centering
\begin{tabular}{lcc}
    \toprule
    \textbf{Which solution is \dots} & \textbf{Annotator 1} & \textbf{Annotator 2} \\
    \midrule
    more conceptually complex      & Solution 1 & Solution 2 \\
    \bottomrule
\end{tabular}
\end{exverdict}

The annotators split on cognitive simplicity. Solution 1 works through a more formal apparatus, notably in the statement and proof of its injectivity lemma and in the explicit definitions of border and interior cells. A reader who finds set-theoretic bookkeeping demanding may call it the more complex one. Solution 2 instead compresses the same content into a global counting identity, which is shorter on the page but requires the reader to hold an abstract quantity in mind and to see why the extremal rows must contribute to it. Thus, which solution is harder depends on the reader's background.

\subsubsection{Computational Ease: Sparse Notation Against Simple Steps} \label{app:example-computational}

This example split the annotators on computational ease, with one solution writing less but demanding more verification per line, and the other writing more but keeping each individual step trivial.

\begin{exproblem}{Problem \texttt{imo-bench\_combinatorics-011}}
A soccer player named Ronaldo stands on a point on a circle with circumference $1$. Given an infinite sequence of positive real numbers $c_1, c_2, c_3, \dots$, Ronaldo successively runs distances $c_1, c_2, c_3, \dots$ around the circle, each time choosing to run either clockwise or counterclockwise. Determine the largest constant $C > 0$ with the following property: for every sequence of positive real numbers $c_1, c_2, c_3, \dots$ with $c_i < C$ for all $i$, Ronaldo can (after studying the sequence) ensure that there is some point on the circle that he will never visit or run across.
\end{exproblem}

\begin{excompare}{gemini}{Solution 1 --- \geminiflash{}}{oai}{Solution 2 --- \gptfivefour{}}
To determine the largest constant $C > 0$ such that for every sequence of positive real numbers $c_1, c_2, c_3, \dots$ with $c_i < C$, Ronaldo can choose directions $s_i \in \{-1, 1\}$ such that he never visits or runs across some point $p$ on the circle, we analyze the conditions under which he can avoid covering the entire circle.

Let the circle be represented by the interval $[0, 1)$ with $0 \equiv 1$. Ronaldo starts at $x_0 = 0$. In each step $i$, he moves a distance $c_i$ to $x_i = (x_{i-1} + s_i c_i) \bmod 1$, covering the arc $L_i$ of length $c_i$ between $x_{i-1}$ and $x_i$.

\textbf{1.} Suppose $C > 1/2$. We show that Ronaldo can be forced to cover the circle. Let $c_1 = c_2 = 1/2 + \epsilon$ for some $0 < \epsilon < C - 1/2$.
\begin{itemize}[itemsep=0em,leftmargin=1.2em,topsep=0.2em]
    \item In step 1, Ronaldo moves distance $c_1$, covering the arc $L_1$ of length $1/2 + \epsilon$. Assume without loss of generality that he moves clockwise, so $x_1 = 1/2 + \epsilon$ and $L_1 = [0, 1/2 + \epsilon]$.
    \item In step 2, Ronaldo moves distance $c_2 = 1/2 + \epsilon$. If he moves clockwise ($s_2 = 1$), the arc $L_2$ is $[1/2 + \epsilon, 1 + 2\epsilon]$. In $\mathbb{R}/\mathbb{Z}$, this covers $[1/2 + \epsilon, 1] \cup [0, 2\epsilon]$. Combined with $L_1$, the covered set is $[0, 1/2 + \epsilon] \cup [1/2 + \epsilon, 1] = [0, 1]$, the entire circle.
    \item If he moves counter-clockwise ($s_2 = -1$), the arc $L_2$ is the set of points obtained by starting at $1/2 + \epsilon$ and moving $1/2 + \epsilon$ backwards. The complement of $L_1$ is the arc $(1/2 + \epsilon, 1)$ of length $1/2 - \epsilon$. Since $c_2 = 1/2 + \epsilon > 1/2 - \epsilon$, the arc $L_2$ covers the entire complement of $L_1$.
\end{itemize}
In either case, Ronaldo covers the entire circle after two steps. Thus, $C$ cannot be greater than $1/2$.

\textbf{2.} Suppose $c_i < 1/2$ for all $i$. We show that Ronaldo can always avoid some point. He can use the greedy strategy: choose $s_i = -\operatorname{sgn}(x_{i-1})$, where $x_{i-1}$ is represented in the range $(-1/2, 1/2]$.
\begin{itemize}[itemsep=0em,leftmargin=1.2em,topsep=0.2em]
    \item Starting with $x_0 = 0$, we have $x_1 = s_1 c_1 \in (-1/2, 1/2)$ since $c_1 < 1/2$.
    \item For each step $n$, if $x_{n-1} \in [0, 1/2)$, then $s_n = -1$ gives $x_n = x_{n-1} - c_n$. Since $0 < c_n < 1/2$, $x_n \in (-1/2, 1/2)$.
    \item Similarly, if $x_{n-1} \in (-1/2, 0)$, then $s_n = 1$ gives $x_n = x_{n-1} + c_n \in (-1/2, 1/2)$.
    \item This strategy ensures that $x_n \in (-1/2, 1/2)$ for all $n$. Each arc $L_n$ has length $c_n < 1/2$ and connects $x_{n-1}$ and $x_n$, both of which are in the open interval $(-1/2, 1/2)$. Because $c_n$ is the shorter arc distance (since $c_n < 1/2$), the arc $L_n$ is contained entirely within the interval $(-1/2, 1/2)$.
    \item Consequently, the union of all arcs $\bigcup_{i=1}^\infty L_i$ is contained in $(-1/2, 1/2)$, meaning the point $1/2$ is never visited or run across.
\end{itemize}
Thus, the largest such constant $C$ is $1/2$.
\[\boxed{1/2}\]

\exswitch

We claim that the largest possible constant is $\frac12$.

First, we use a convenient reformulation. If there is a point $P$ on the circle that Ronaldo never visits and never runs across, then we may \textbf{cut} the circle at $P$. What remains is an open interval of length $1$, which we identify with $(0,1)$. Ronaldo's starting point becomes some $x_0\in(0,1)$, and after each move his new position is
\[
x_n=x_{n-1}\pm c_n,
\]
because in the cut interval, ``clockwise'' and ``counterclockwise'' simply mean moving right or left by distance $c_n$. Also, every segment he runs is contained in $(0,1)$, since touching an endpoint would mean touching $P$.

Conversely, if we can choose signs so that a walk with steps $c_1,c_2,\dots$ stays entirely inside $(0,1)$, then after gluing the endpoints back together we get a strategy on the circle that avoids the cut point $P$. So the problem is equivalent to this: find the largest $C$ such that for every sequence $0<c_n<C$, one can choose left/right moves so that the whole walk stays inside an open interval of length $1$.

\textbf{1. Lower bound: $C=\frac12$ works.} Assume $0<c_n<\frac12$ for all $n$. Choose $P$ to be the point opposite Ronaldo's starting point on the circle. After cutting at $P$, the start is at the midpoint of the interval, so $x_0=\frac12$. Now use the following rule:
\begin{itemize}[itemsep=0em,leftmargin=1.2em,topsep=0.2em]
    \item if $x_{n-1}<\frac12$, move to the right, so $x_n=x_{n-1}+c_n$;
    \item if $x_{n-1}>\frac12$, move to the left, so $x_n=x_{n-1}-c_n$;
    \item if $x_{n-1}=\frac12$, either choice is fine.
\end{itemize}
We check that this always keeps $x_n$ inside $(0,1)$.

\emph{Case 1: $x_{n-1}\le \frac12$.} Then we move right, $x_n=x_{n-1}+c_n$. Since $x_{n-1}>0$ and $c_n>0$, we have $x_n>0$. Also,
\[
x_n=x_{n-1}+c_n<\tfrac12+\tfrac12=1.
\]
So $x_n\in(0,1)$.

\emph{Case 2: $x_{n-1}\ge \frac12$.} Then we move left, $x_n=x_{n-1}-c_n$. Because $x_{n-1}\ge \frac12$ and $c_n<\frac12$,
\[
x_n=x_{n-1}-c_n>\tfrac12-\tfrac12=0.
\]
Also $x_n<1$ because $x_{n-1}<1$. Thus again $x_n\in(0,1)$.

So by induction, every position $x_n$ stays in $(0,1)$. Hence the cut point $P$ is never visited and never crossed. Therefore every sequence with all terms $<\frac12$ is avoidable, so $C\ge \frac12$.

\textbf{2. Upper bound: no $C>\frac12$ can work.} Now suppose $C>\frac12$. We will construct a sequence with all terms $<C$ for which avoiding a point is impossible. Choose numbers $a,b$ such that
\[
\tfrac12<b<a<\min(C,1),
\]
and consider the infinite sequence $c_1=a$, $c_2=b$, $c_3=a$, $c_4=b,\dots$

Assume for contradiction that Ronaldo can avoid some point $P$. Cut the circle at $P$. Then his positions satisfy
\[
x_n=x_{n-1}+\varepsilon_n c_n, \qquad \varepsilon_n\in\{+1,-1\},
\]
with all $x_n\in(0,1)$. We now show this is impossible.

\emph{Claim: consecutive signs must be opposite.} Suppose for some $n$ we had $\varepsilon_{n+1}=\varepsilon_n$. Then
\[
\begin{aligned}
x_{n+1}-x_{n-1} &=(x_n-x_{n-1})+(x_{n+1}-x_n) \\
&=\varepsilon_n c_n+\varepsilon_{n+1} c_{n+1}.
\end{aligned}
\]
Since the signs are equal, this becomes $x_{n+1}-x_{n-1}=\pm(c_n+c_{n+1})$. But $c_n,c_{n+1}$ are one $a$ and one $b$, so
\[
|x_{n+1}-x_{n-1}|=a+b.
\]
Because $a>\frac12$ and $b>\frac12$, we get $a+b>1$. On the other hand, both $x_{n-1}$ and $x_{n+1}$ lie in the interval $(0,1)$, so their distance must be strictly less than $1$. Contradiction. Therefore $\varepsilon_{n+1}=-\varepsilon_n$ for every $n$, so the signs must alternate forever.

\emph{What alternating signs imply.} There are only two possibilities.

\emph{Possibility 1: $+,-,+,-,\dots$} Then every pair of moves changes the position by $a-b>0$. Indeed,
\[
x_2=x_0+a-b,\qquad x_4=x_2+a-b=x_0+2(a-b),
\]
and in general $x_{2k}=x_0+k(a-b)$. Since $a-b>0$, choose $k$ so large that $k(a-b)>1-x_0$. Then $x_{2k}>1$, impossible because $x_{2k}\in(0,1)$.

\emph{Possibility 2: $-,+,-,+,\dots$} Now every pair of moves changes the position by $-(a-b)<0$, so similarly $x_{2k}=x_0-k(a-b)$. Choose $k$ so large that $k(a-b)>x_0$. Then $x_{2k}<0$, again impossible.

Both possibilities lead to contradictions. So for this sequence $a,b,a,b,\dots$, no point can be left untouched. Hence the required property fails for every $C>\frac12$, and therefore $C\le \frac12$.

Since we proved both $C\ge \frac12$ and $C\le \frac12$, the largest constant is
\[\boxed{\tfrac{1}{2}}\]
\end{excompare}

\begin{exverdict}{Annotator verdicts --- \texttt{imo-bench\_combinatorics-011}}
\centering
\begin{tabular}{lcc}
    \toprule
    \textbf{Which solution is \dots} & \textbf{Annotator 1} & \textbf{Annotator 2} \\
    \midrule
    more computationally demanding & Solution 1 & Solution 2 \\
    \bottomrule
\end{tabular}
\end{exverdict}

The annotators split on computational ease, and the disagreement turns on what ``time to follow line-by-line'' means when the two solutions distribute their difficulty differently. Solution 1 uses much sparser notation and reads more cleanly, but verifying each of its claims about the arcs and the inequalities can take the reader considerably longer, because a single sentence often compresses a case check that the reader must reconstruct. Solution 2 introduces more symbols and performs more algebraic steps in total, yet each individual step is simpler to parse, since the reduction to a walk on $(0,1)$ turns every subsequent claim into an elementary inequality.

\subsubsection{Judge Sensitivity on Near-Identical Solutions} \label{app:example-judge}

We also study disagreement between annotators and the LLM judge. When two solutions are nearly identical, humans record a tie, whereas the judge may pick up on a small surface difference.

\begin{exproblem}{Problem \texttt{aime\_2025\_27}}
Let $A_1 A_2 A_3 \ldots A_{11}$ be an $11$-sided non-convex simple polygon with the following properties:
\begin{itemize}[itemsep=0em,leftmargin=2em,topsep=0.2em]
    \item For every integer $2 \le i \le 10$, the area of $\triangle A_i A_{1} A_{i+1}$ is equal to $1$.
    \item For every integer $2 \le i \le 10$, $\cos(\angle A_i A_{1} A_{i+1}) = \frac{12}{13}$.
    \item The perimeter of the $11$-gon $A_1 A_2 A_3 \dots A_{11}$ is equal to $20$.
\end{itemize}
Then $A_1 A_2 + A_1 A_{11} = \frac{m \sqrt{n} - p}{q}$ where $m, n, p$, and $q$ are positive integers, $n$ is not divisible by the square of any prime, and no prime divides all of $m, p$, and $q$. Find $m + n + p + q$.
\end{exproblem}

\begin{excompare}{qwen}{Solution 1 --- \qwenthreeshort{}}{oai}{Solution 2 --- \gptfivefour{}}
Let $x_i$ denote the length of the segment $A_1 A_i$ for $i = 1, 2, \dots, 11$. Since $A_1$ is a vertex, $x_1 = 0$. We are interested in finding the sum $x_2 + x_{11}$.

\textbf{Step 1: Analyze the properties of the triangles.}
For every integer $2 \le i \le 10$, consider the triangle $\triangle A_i A_1 A_{i+1}$. Let $\alpha_i = \angle A_i A_1 A_{i+1}$. We are given that $\cos(\alpha_i) = \frac{12}{13}$. Since $\alpha_i$ is an angle in a triangle, $\sin(\alpha_i) > 0$, so:
\[ \sin(\alpha_i) = \sqrt{1 - \left(\frac{12}{13}\right)^2} = \sqrt{\frac{25}{169}} = \frac{5}{13} \]
The area of $\triangle A_i A_1 A_{i+1}$ is given as $1$. Using the sine area formula:
\[ \text{Area} = \frac{1}{2} x_i x_{i+1} \sin(\alpha_i) = 1 \]
Substituting $\sin(\alpha_i) = \frac{5}{13}$:
\[ \frac{1}{2} x_i x_{i+1} \left(\frac{5}{13}\right) = 1 \implies x_i x_{i+1} = \frac{26}{5} \]
This relation holds for $i = 2, 3, \dots, 10$.

\textbf{Step 2: Determine the sequence of lengths $x_i$.}
From the relation $x_i x_{i+1} = \frac{26}{5}$, we can see that $x_{i+2} = \frac{26/5}{x_{i+1}} = x_i$. Thus, the sequence of lengths $x_2, x_3, \dots, x_{11}$ is periodic with period $2$. Let $x_2 = a$ and $x_3 = b$. Then $x_4 = a, x_5 = b$, and so on. The sequence is $a, b, a, b, a, b, a, b, a, b$. Specifically, $x_2 = a$ and $x_{11} = b$. We have the product $ab = \frac{26}{5}$. We need to find the sum $S = a + b$.

\textbf{Step 3: Express the perimeter in terms of $a$ and $b$.}
The perimeter of the polygon is the sum of the lengths of its edges:
\[ \begin{aligned} P &= A_1 A_2 + \sum_{i=2}^{10} A_i A_{i+1} + A_{11} A_1 \\ &= a + \sum_{i=2}^{10} A_i A_{i+1} + b \end{aligned} \]
There are $9$ terms in the summation. Let $L_i = A_i A_{i+1}$. By the Law of Cosines on $\triangle A_i A_1 A_{i+1}$:
\[ L_i^2 = x_i^2 + x_{i+1}^2 - 2 x_i x_{i+1} \cos(\alpha_i) \]
Since $x_i x_{i+1} = \frac{26}{5}$ and $\cos(\alpha_i) = \frac{12}{13}$, the term $2 x_i x_{i+1} \cos(\alpha_i)$ is constant:
\[ 2 \left(\frac{26}{5}\right) \left(\frac{12}{13}\right) = \frac{48}{5} \]
Thus, $L_i^2 = x_i^2 + x_{i+1}^2 - \frac{48}{5}$. For any $i$, the pair $(x_i, x_{i+1})$ is either $(a, b)$ or $(b, a)$. In both cases, $x_i^2 + x_{i+1}^2 = a^2 + b^2$. Therefore, all edge lengths $A_i A_{i+1}$ are equal. Let $L = \sqrt{a^2 + b^2 - \frac{48}{5}}$. The perimeter equation becomes:
\[ 20 = a + b + 9L = S + 9\sqrt{a^2 + b^2 - \frac{48}{5}} \]
Substituting $a^2 + b^2 = (a+b)^2 - 2ab = S^2 - \frac{52}{5}$:
\[ \begin{aligned} 20 &= S + 9\sqrt{S^2 - \frac{52}{5} - \frac{48}{5}} \\ &= S + 9\sqrt{S^2 - \frac{100}{5}} = S + 9\sqrt{S^2 - 20} \end{aligned} \]

\textbf{Step 4: Solve for $S$.}
Rearranging the equation:
\[ 9\sqrt{S^2 - 20} = 20 - S \]
Squaring both sides (noting that $20-S \ge 0$):
\[ 81(S^2 - 20) = (20 - S)^2 \]
\[ 81S^2 - 1620 = 400 - 40S + S^2 \]
\[ 80S^2 + 40S - 2020 = 0 \]
Dividing by $20$:
\[ 4S^2 + 2S - 101 = 0 \]
Using the quadratic formula for the positive root (since $S > 0$):
\[ \begin{aligned} S &= \frac{-2 + \sqrt{2^2 - 4(4)(-101)}}{2(4)} \\ &= \frac{-2 + \sqrt{4 + 1616}}{8} = \frac{-2 + \sqrt{1620}}{8} \end{aligned} \]
Simplifying $\sqrt{1620} = \sqrt{324 \times 5} = 18\sqrt{5}$:
\[ S = \frac{-2 + 18\sqrt{5}}{8} = \frac{9\sqrt{5} - 1}{4} \]

\textbf{Step 5: Final calculation.}
We have $A_1 A_2 + A_1 A_{11} = S = \frac{9\sqrt{5} - 1}{4}$. Comparing this to the form $\frac{m \sqrt{n} - p}{q}$, we identify $m = 9$, $n = 5$, $p = 1$, $q = 4$. These satisfy the conditions: $n=5$ is square-free, and $\gcd(9, 1, 4) = 1$. The required sum is:
\[ m + n + p + q = 9 + 5 + 1 + 4 = 19 \]
\[\boxed{19}\]

\exswitch

Let $x_i=A_1A_i$ for $2\le i\le 11$.

For each $i=2,3,\dots,10$, consider $\triangle A_iA_1A_{i+1}$. Since
\[ \cos\angle A_iA_1A_{i+1}=\frac{12}{13}, \]
and this is an angle of a triangle, its sine is positive, so
\[ \sin\angle A_iA_1A_{i+1} =\sqrt{1-\left(\frac{12}{13}\right)^2} =\frac{5}{13}. \]
Now use the area formula:
\[ \begin{aligned} 1 &= \text{Area}(\triangle A_iA_1A_{i+1}) \\ &= \tfrac12\,x_i x_{i+1}\sin\angle A_iA_1A_{i+1} \\ &= \tfrac12\,x_i x_{i+1}\cdot \frac{5}{13}. \end{aligned} \]
Hence
\[ x_i x_{i+1}=\frac{26}{5}\qquad (2\le i\le 10). \]
So all consecutive products are equal:
\[ x_2x_3=x_3x_4=\cdots=x_{10}x_{11}=\frac{26}{5}. \]
Because every $x_i>0$, from $x_i x_{i+1}=x_{i+1}x_{i+2}$ we may divide by $x_{i+1}$ to get $x_i=x_{i+2}$. Therefore the distances from $A_1$ alternate between two values. Set
\[ \begin{gathered} u=x_2=x_4=x_6=x_8=x_{10}, \\ v=x_3=x_5=x_7=x_9=x_{11}. \end{gathered} \]
Then $uv=\frac{26}{5}$.

Now compute the side lengths $A_iA_{i+1}$ for $i=2,\dots,10$. By the Law of Cosines in $\triangle A_iA_1A_{i+1}$,
\[ A_iA_{i+1}^2=x_i^2+x_{i+1}^2-2x_ix_{i+1}\cdot \frac{12}{13}. \]
Since $(x_i,x_{i+1})$ is always $(u,v)$ or $(v,u)$, all these nine side lengths are equal. Call their common length $s$. Then
\[ s^2=u^2+v^2-2uv\cdot \frac{12}{13}. \]
Let $t=u+v=A_1A_2+A_1A_{11}$. Because $u^2+v^2=t^2-2uv$, we get
\[ \begin{aligned} s^2 &= t^2-2uv-\frac{24}{13}uv \\ &= t^2-\left(2+\frac{24}{13}\right)uv \\ &= t^2-\frac{50}{13}\cdot \frac{26}{5}. \end{aligned} \]
Now $\frac{50}{13}\cdot \frac{26}{5}=20$, so $s^2=t^2-20$.

Next use the perimeter condition. The perimeter is
\[ \begin{gathered} A_1A_2+(A_2A_3+\cdots +A_{10}A_{11})+A_{11}A_1 \\ = u+9s+v=t+9s. \end{gathered} \]
Since the perimeter is $20$, we get $t+9s=20$, so $s=\frac{20-t}{9}$. Substitute this into $s^2=t^2-20$:
\[ \left(\frac{20-t}{9}\right)^2=t^2-20. \]
Multiply both sides by $81$:
\[ (20-t)^2=81(t^2-20). \]
Expand:
\[ 400-40t+t^2=81t^2-1620. \]
So $80t^2+40t-2020=0$. Divide by $20$:
\[ 4t^2+2t-101=0. \]
Now solve:
\[ \begin{aligned} t &= \frac{-2\pm\sqrt{2^2-4\cdot 4\cdot (-101)}}{8} \\ &= \frac{-2\pm\sqrt{4+1616}}{8} = \frac{-2\pm\sqrt{1620}}{8}. \end{aligned} \]
Since $\sqrt{1620}=\sqrt{324\cdot 5}=18\sqrt5$, we get
\[ t=\frac{-2\pm 18\sqrt5}{8} =\frac{-1\pm 9\sqrt5}{4}. \]
Because $t=u+v>0$, we must take $t=\frac{9\sqrt5-1}{4}$. Thus
\[ A_1A_2+A_1A_{11}=\frac{9\sqrt5-1}{4}. \]
So $m=9$, $n=5$, $p=1$, $q=4$, and $m+n+p+q=9+5+1+4=19$.
\[\boxed{19}\]
\end{excompare}

\begin{exverdict}{Verdicts --- \texttt{aime\_2025\_27}}
\centering
\footnotesize
\begin{tabular}{lccc}
    \toprule
    \textbf{Which solution is \dots} & \textbf{Annotator 1} & \textbf{Annotator 2} & \textbf{\gptoss{} judge} \\
    \midrule
    more conceptually complex      & tie & tie & --- \\
    more computationally demanding & tie & tie & Solution 1 \\
    \bottomrule
\end{tabular}
\end{exverdict}

Both annotators graded the comparison a tie, since the two solutions use essentially the same method and their computations barely differ. Namely, both derive $x_i x_{i+1} = 26/5$ from the area condition, both deduce the alternation of the distances from $A_1$, both apply the Law of Cosines to get $s^2 = S^2 - 20$, and both finish with the same quadratic. The LLM judge, by contrast, rated Solution 1 as the more computationally demanding one, most plausibly because it lays out its computations at greater length and labels them as numbered steps. However, because each model rating in \bench{} aggregates hundreds of such judgments, such discrepancies that are too small to decide any individual comparison average out across problems, while a genuine difference in quality recurs and becomes measurable. This is also why we report ratings across models rather than interpreting single pairwise verdicts.

\subsection{Importance of Nuance} \label{app:examples-nuance}

The examples we showcased above illustrate that even when two solutions are both correct, they can differ substantially in style and quality, and in varying degrees of nuance. Quantifying differences of varying granularity in a consistent and unbiased manner is important for a benchmark, as otherwise a qualitative inspection would suffice and a metric confirming the obvious would not be beneficial. We can quantify the balance between the two regimes directly for conciseness, where the pairwise decision is driven by the relative compressibility gap:
\begin{equation}
    \Delta(p_a, p_b) = \frac{|C(p_a) - C(p_b)|}{\max(C(p_a), C(p_b))},
\end{equation}
with $C(\cdot)$ denoting the compressibility of a solution. \Cref{fig:compressibility_gap} measures the distribution of $\Delta$ over all pairwise comparisons in \bench{}, showing that nuanced and clearly separated pairs are both well-represented in our benchmark. This feature is what allows our metrics to separate solutions and models that a reader might find difficult to distinguish pair-by-pair otherwise.

\begin{figure}[h]
\centering
% Distribution of the relative compressibility gap over all pairwise
% conciseness comparisons in ProofRank. Bins have width 0.05; the rightmost
% bar aggregates every pair with a gap above 0.75. Dashed rule: median gap.
\definecolor{gapbar}{HTML}{4C72B0}

\begin{tikzpicture}[font=\scriptsize]
\begin{axis}[
  width=10.5cm, height=4.6cm, scale only axis,
  xmin=0, xmax=0.8,
  xtick={0,0.1,0.2,0.3,0.4,0.5,0.6,0.7,0.8},
  xticklabels={$0$,$0.1$,$0.2$,$0.3$,$0.4$,$0.5$,$0.6$,$0.7$,${>}0.75$},
  ymin=0, ymax=10.4, ytick={0,2,4,6,8,10},
  yticklabel={\pgfmathprintnumber{\tick}\%},
  y tick label style={font=\scriptsize}, x tick label style={font=\scriptsize},
  tick align=outside, tick pos=left,
  ymajorgrids=true, grid style={black!12, line width=0.3pt},
  axis line style={black!55, line width=0.4pt},
  xlabel={Relative compressibility gap $\Delta$},
  ylabel={Fraction of pairs},
  xlabel style={font=\scriptsize}, ylabel style={font=\scriptsize},
]
  % histogram: 16 bins of width 0.05, given as bin edges (last y is ignored)
  \addplot[ybar interval=0.86, fill=gapbar!75, draw=gapbar!90!black,
           line width=0.4pt] coordinates {
    (0.00,8.2) (0.05,8.6) (0.10,8.8) (0.15,9.5)
    (0.20,8.8) (0.25,8.8) (0.30,7.8) (0.35,7.9)
    (0.40,7.0) (0.45,6.5) (0.50,5.9) (0.55,4.5)
    (0.60,3.2) (0.65,2.2) (0.70,1.3) (0.75,1.0)
    (0.80,0)};
  % median gap
  \draw[black!70, dashed, line width=0.7pt]
    (axis cs:0.285,0) -- (axis cs:0.285,9.25);
  \node[anchor=south, font=\scriptsize, text=black!70]
    at (axis cs:0.285,9.25) {median $0.28$};
\end{axis}
\end{tikzpicture}
\caption{Distribution of the relative compressibility gap $\Delta$ over all pairwise conciseness comparisons.}
\label{fig:compressibility_gap}
\end{figure}

\section{Prompts and Instructions} \label{app:prompts}

In this section, we provide the specific prompts used throughout our experiments for reproducibility.

\subsection{Human Validation Instructions} \label{app:human-validation-instructions}

Below we include the instructions provided to the human annotators for pairwise comparisons of the solutions on the different metrics. For every dimension, annotators were instructed that they could also answer 0.5 to indicate a tie or ``no preference''.

\begin{prompt}{Verbosity}
Decide which solution is more verbose: 0 = left, 1 = right. Judge verbosity by how much text is used to express the same content (extra explanations, repeated statements, long-winded phrasing, unnecessary intermediate steps), not by mathematical difficulty. If they feel equally verbose, pick the one that spends more words on redundant detail.
\end{prompt}

\begin{prompt}{Elegance}
Decide which solution is more elegant: 0 = left, 1 = right. High elegance feels clean and unified: it uses a clever trick, symmetry, an invariant, or a principled shortcut that avoids heavy calculation and produces an "aha" moment. Standard elegance uses the expected method efficiently but without surprise. Low elegance is brute force: long algebra/coordinate bashing when a cleaner approach exists, large/manual casework, or repetitive computations that obscure the main idea. If both are similarly elegant, choose the one with fewer moving parts and a clearer unifying idea.
\end{prompt}

\begin{prompt}{Insightfulness}
Decide which solution is more insightful: 0 = left, 1 = right. An insightful solution exposes a deeper structure, gives a clear "why this works" viewpoint, or makes a meaningful connection between ideas/areas (e.g., reframing number theory via combinatorics/invariants/geometry), and leaves the reader with a reusable lesson. A less insightful solution is mainly routine manipulation or casework, follows a standard template, or applies a "black-box" theorem/formula with little explanation of why it is the right fit. If both feel equally insightful or there isn't enough information to tell, pick the one that better explains the core idea rather than the one with more steps.
\end{prompt}

\begin{prompt}{Cognitive Simplicity}
Decide which solution is more conceptually complex (more ingenuity/non-obvious core idea, more specialized/advanced tools, or harder integration of multiple ideas). Ignore notation density, computation/bookkeeping time, and step count unless it reflects genuinely different conceptual ingredients. Output exactly one number and nothing else: 0 = Solution 1 is more complex, 0.5 = Tie/unclear from the text, 1 = Solution 2 is more complex.
\end{prompt}

\begin{prompt}{Computational Ease}
Decide which one takes more time to follow line-by-line due to:
 - heavier notation/symbol tracking (many variables/indices, dense formulas, switching conventions, overloaded symbols),
 - unclear or inconsistent definitions/references (symbols used before defined, ambiguous bounds/domains/constraints),
 - more explicit computation/bookkeeping (long chains of routine steps, expansions, substitutions, sign/exponent/index tracking).

Ignore conceptual cleverness, focus only on time-to-parse/track what is written.
 - 0 = Solution 1 imposes more time/load
 - 0.5 = Tie / cannot confidently rank from the text
 - 1 = Solution 2 imposes more time/load
\end{prompt}

\subsection{Solution Generation Prompts}\label{app:solver_prompts}
We use the following prompts to generate solutions for our evaluation. This includes our standard prompt that follows \citep{opc}, as well as variations for specific experiments: technique-constrained generation for adaptivity and two prompts emphasizing conciseness. The prompts used in the steerability analysis of \cref{sec:steerability} are given in \cref{app:steering_prompts}.

\begin{prompt}{Standard Generation Prompt}
Your task is to write a proof solution to the following problem, focusing on accuracy, thoroughness, and clarity. When you write your proof, follow these guidelines:

- You are creating a proof, not a proof outline. Each step should be carefully explained and documented. If not properly explained, the judge will assume that you cannot explain it, and therefore decrease your grade.
- You can use general theorems and lemmas, but only if they are well-known. As a rule of thumb: if the result has a name and is famous enough to have a Wikipedia page or something similar to describe it, it is allowed. Any result from papers that would not be taught in high-school or low-level bachelor courses in mathematics should not be used.
- Do not skip computation steps in your proof. Clearly explain what transformations were done and why they are allowed in each step of a calculation.
- Your proof should be self-contained.
- If you are not sure about a specific step, or do not know how to prove an intermediate result, clearly state this. It is much preferable to indicate your uncertainty rather than making incorrect statements or claims.
- Put your final answer within \\boxed{{}}.

{problem}

\end{prompt}

\begin{prompt}{Technique-constrained Generation Prompt}

Your task is to write a proof solution to the following problem. You will be required to solve the problem using a specific solving technique. When you write your proof, follow these guidelines:

- You are creating a proof, not a proof outline. Each step should be carefully explained and documented. If not properly explained, the judge will assume that you cannot explain it, and therefore decrease your grade.
- You can use general theorems and lemmas, but only if they are well-known. As a rule of thumb: if the result has a name and is famous enough to have a Wikipedia page or something similar to describe it, it is allowed. Any result from papers that would not be taught in high-school or low-level bachelor courses in mathematics should not be used.
- Do not skip computation steps in your proof. Clearly explain what transformations were done and why they are allowed in each step of a calculation.
- Your proof should be self-contained.
- Your proof should make use of the specified technique. You will not receive any credit if you fail to use the specified technique.
- If you are not sure about a specific step, or do not know how to prove an intermediate result, clearly state this. It is much preferable to indicate your uncertainty rather than making incorrect statements or claims.
- Put your final answer within \\boxed{{}}.

{problem}

Solve the problem using the following method: {technique}
\end{prompt}

\subsection{Directed Prompts} \label{app:steering_prompts}

Each instruction below is appended to the guideline list of the standard generation prompt of \cref{app:solver_prompts}, directly after its final guideline, and the rest of that prompt is left unchanged, with the exception of the conciseness prompts, and the single-prompt diversity, which are provided in full.

\begin{prompt}{Added Instruction: Cognitive Simplicity (Easier)}
- Prefer the most standard, well-known method for this type of problem. Avoid clever tricks, specialized theorems, and non-obvious reformulations whenever a more elementary argument exists, so long as the proof remains correct and complete.
\end{prompt}

\begin{prompt}{Added Instruction: Cognitive Simplicity (Harder)}
- Prefer a conceptually sophisticated approach: a clever key idea, a specialized theorem, or a non-obvious reformulation. Avoid the most standard textbook method whenever a more ingenious correct argument exists, so long as the proof remains correct and complete.
\end{prompt}

\begin{prompt}{Added Instruction: Computational Ease (Lighter)}
- Minimize explicit computation and notational bookkeeping: avoid long algebraic manipulations, heavy case analysis, dense notation, and long chains of routine steps whenever an argument with less bookkeeping exists, so long as the proof remains correct and complete. Do not silently skip computations that your argument requires; instead, prefer approaches that do not need them.
\end{prompt}

\begin{prompt}{Added Instruction: Computational Ease (Heavier)}
- Prefer a direct, computation-driven approach (e.g., coordinates, algebraic expansion, explicit case analysis) with every computation written out step by step, even when a slicker conceptual argument exists, so long as the proof remains correct and complete.
\end{prompt}

\begin{prompt}{Generation Prompt for Conciseness}
Your task is to write a proof solution to the following problem, focusing on accuracy, thoroughness, and clarity. You should make sure that your proof contains as little extra comments as possible, and is as concise as possible so long as the proof is correct. When you write your proof, follow these guidelines:

- You are creating a proof, not a proof outline. Each step should be carefully explained and documented. If not properly explained, the judge will assume that you cannot explain it, and therefore decrease your grade.
- You can use general theorems and lemmas, but only if they are well-known. As a rule of thumb: if the result has a name and is famous enough to have a Wikipedia page or something similar to describe it, it is allowed. Any result from papers that would not be taught in high-school or low-level bachelor courses in mathematics should not be used.
- Do not skip computation steps in your proof. Clearly explain what transformations were done and why they are allowed in each step of a calculation.
- Your proof should be self-contained.
- Make sure the proof is as concise as possible while still being correct. Avoid additional explanations or comments that are not strictly necessary for the correctness of the proof.
- Put your final answer within \\boxed{{}}.

{problem}
\end{prompt}

\begin{prompt}{Generation Prompt for Stronger Conciseness}
Your task is to write a proof solution to the following problem, focusing on conciseness. You should make sure that your proof contains as little extra comments as possible, and is as concise as possible so long as the proof is correct. When you write your proof, follow these guidelines:

  - You are creating a proof, not a proof outline. Each step should be carefully explained and documented.
  - Make sure the proof is as concise as possible while still being correct. Avoid additional explanations or comments that are not strictly necessary for the correctness of the proof.

{problem}
\end{prompt}

\begin{prompt}{Generation Prompt for Diversity}
Your task is to write proof solutions to the following problem. You must provide EXACTLY 4 substantially different solutions, using genuinely different core methods or key ideas (not minor variations of the same argument). When you write your proofs, follow these guidelines:

- You are creating a proof, not a proof outline. Each step should be carefully explained and documented. If not properly explained, the judge will assume that you cannot explain it, and therefore decrease your grade.
- You can use general theorems and lemmas, but only if they are well-known. As a rule of thumb: if the result has a name and is famous enough to have a Wikipedia page or something similar to describe it, it is allowed. Any result from papers that would not be taught in high-school or low-level bachelor courses in mathematics should not be used.
- Do not skip computation steps in your proof. Clearly explain what transformations were done and why they are allowed in each step of a calculation.
- Your proof should be self-contained.
- If you are not sure about a specific step, or do not know how to prove an intermediate result, clearly state this. It is much preferable to indicate your uncertainty rather than making incorrect statements or claims.
- Put your final answer within \\boxed{{}}.
- Each of the 4 solutions must be fully self-contained: do not refer to the other solutions, and restate any setup each solution needs.
- Each solution must end with the final answer within \\boxed{{}}.
- Wrap each solution in numbered tags, exactly as follows:

<solution_1>
(first complete solution)
</solution_1>

<solution_2>
(second complete solution)
</solution_2>

<solution_3>
(third complete solution)
</solution_3>

<solution_4>
(fourth complete solution)
</solution_4>

Do not put any text after the closing tag of the fourth solution.

{problem}
\end{prompt}

\subsection{Correctness Verification Prompts}
\subsubsection{Correctness Verification Prompt}\label{app:solution-verification-prompt}
We use the following prompt for running \gptfivefour to determine the overall correctness of a mathematical solution.

\begin{prompt}{Solution Verification Prompt}
You are judging the correctness of an LLM-generated proof for a math problem.

### Input:

Your input will consist of the following components:
- **Problem Statement**: A mathematical problem that the proof is attempting to solve.
- **Proof Solution**: The proof that you need to evaluate. This proof may contain errors, omissions, or unclear steps. The proof was generated by another language model, which was given the following instructions:
<model_prompt>
- You are creating a proof, not a proof outline. Each step should be carefully explained and documented. If not properly explained, the judge will assume that you cannot explain it, and therefore decrease your grade.
- You can use general theorems and lemmas, but only if they are well-known. As a rule of thumb: if the result has a name and is famous enough to have a Wikipedia page or something similar to describe it, it is allowed. Any result from papers that would not be taught in high-school or low-level bachelor courses in mathematics should not be used. Any use of such results will immediately give you a zero grade.
- Do not skip computation steps in your proof. Clearly explain what transformations were done and why they are allowed in each step of a calculation.
- You should use correct LaTeX notation to write equations and mathematical symbols. You should encompass these equations in appropriate symbols ("\\(" and "\\)" for inline math, "\\[" and "\\]" for block math) to enhance the clarity of your proof. Do not use any unicode characters.
- Your proof should be self-contained.
- If you are not sure about a specific step, or do not know how to prove an intermediate result, clearly state this. It is much preferable to indicate your uncertainty rather than making incorrect statements or claims.
</model_prompt>

### How the solution should be graded:
A solution should be considered correct even if it would earn 5+/7 points in a standard grading format. Examples of small penalties worth 1 point are if the solution:
- Makes a small computational mistake that can be easily fixed
- Misses an edge case which can be easily proven/disproven
- Skips over a step that follows without much reasoning or manual work
Depending on the severity and the context, you may also not penalise a given error. On the other hand, a solution should be marked as incorrect if:
- It marks a step as trivial, if it is not immediately obvious with little reasoning why this would be the case.
- It omits algebra-heavy computational steps, regardless of whether or not it has outlined the methodology. Skipping shorter computations should be permitted.
- Generalizes over a pattern without rigorously describing the pattern, or without proving any relevant properties.
- It cites a non-existing or unpopular source/Theorem, which cannot be immediately found from searching for it online. Thus, any theorems that can be immediately found and have a Wikipedia article are allowed.

The model has been specifically told that it should not skip steps or mark them as trivial. Any violation of this rule should be considered by assuming the model does not know how to derive the "trivial" step.

### Scoring instructions

If you believe the proof is correct, end your analysis with \\boxed{{correct}}. If you believe the proof is incorrect, end your analysis with \\boxed{{incorrect}}.

### Problem Statement:
{problem}

### Model Solution:
{solution}
\end{prompt}

\subsubsection{Answer Verification Prompt}\label{app:answer-verification-prompt}
Following \citep{imobench}, we use the following prompt to verify the correctness of the final answer provided in a model solution.
\begin{prompt}{Answer Verification Prompt}
# System Role: Deterministic Mathematical Autograder
You are a precise, automated grading system. Your sole function is to determine if the final answer provided in the Model Solution is mathematically equivalent to the Golden Answer. You must NOT grade the reasoning or steps, only the final result.

# 1. Grading Guidelines (Equivalence Rules)

Equivalence is mandatory for a correct grade. You must rigorously verify if the answers represent the exact same mathematical value or expression, even if the format differs.

* **Algebraic Equivalence:** e.g., 'n(n+1)/2' is equivalent to 'n^2/2 + n/2'. You must verify the algebra.

* **Numerical Equivalence:** e.g., '1/2' is equivalent to '0.5'; 'sqrt(2)/2' is equivalent to '1/sqrt(2)'.

* **Set/List Equivalence:** Unless specified as an ordered tuple/vector, the order of elements does not matter (e.g., {{1, 2}} is equivalent to {{2, 1}}).

* **Partial Credit:** No partial credit is allowed. If the answer is incomplete or partially incorrect, it is incorrect.

* **No Answers:** If no clear, unambiguous final answer can be extracted, the solution must be graded as incorrect.

# 3. Output Protocol (Strict Compliance Required)

You must execute the task using a two-part structure. Failure to follow this structure will result in task failure.

**Part 1: Analysis (Chain-of-Thought)**

You MUST perform your analysis within <thinking></thinking> tags. Make your thinking concise. This section details your reasoning process and must follow these steps sequentially:

1. **Golden Answer:** State the Golden Answer.

2. **Extracted Model Answer:** State the extracted answer based on the Extraction Protocol. If none found, state "No clear final answer found."

3. **Equivalence Analysis:** Compare the two answers using the Grading Guidelines. Detail the steps taken to verify mathematical equivalence (e.g., simplification, algebraic manipulation). You must actively try to prove they are the same before concluding they are different.

4. **Conclusion:** State the final determination ("Correct" or "Incorrect").

**Part 2: Final Grade**

Immediately following the closing </thinking> tag, output **ONLY** the final grade.

* If Correct: \\boxed{{Correct}}
* If Incorrect: \\boxed{{Incorrect}}

**CRITICAL CONSTRAINT: Do not add any text, explanations, or formatting outside the <thinking> tags or the final \\boxed{{}} output.**

-------
Output example:
<thinking>
1. **Golden Answer:** (-\infty, -4) \cup (4, \infty)
2. **Extracted Model Answer:** ∅ (the empty set)
3. **Equivalence Analysis:**
The Golden Answer is a non-empty set of real numbers. The Model Answer is the empty set. These two sets are not equivalent. The empty set contains no elements, while the Golden Answer contains an infinite number of elements.

4. **Conclusion:** Incorrect
</thinking>
\\boxed{{Incorrect}}

-------
# 4. Input Data
Here is the problem, model solution, and golden answer to grade:

### Problem Statement:
{problem}

### Model Solution:
{solution}

### Golden Answer:
{gold_answer}
\end{prompt}

\subsubsection{Completeness Verification Prompt}\label{app:completeness-verification-prompt}
We use the following prompt for running \gptfivefour to determine whether the reasoning presented in a mathematical solution is complete.

\begin{prompt}{Solution Verification Prompt}
You are judging the completeness of an LLM-generated proof for a math problem.

### Input:

Your input will consist of the following components:
- **Problem Statement**: A mathematical problem that the proof is attempting to solve.
- **Proof Solution**: The proof that you need to evaluate. This proof may contain errors, omissions, or unclear steps. The proof was generated by another language model, which was given the following instructions:
<model_prompt>
- You are creating a proof, not a proof outline. Each step should be carefully explained and documented. If not properly explained, the judge will assume that you cannot explain it, and therefore decrease your grade.
- You can use general theorems and lemmas, but only if they are well-known. As a rule of thumb: if the result has a name and is famous enough to have a Wikipedia page or something similar to describe it, it is allowed. Any result from papers that would not be taught in high-school or low-level bachelor courses in mathematics should not be used. Any use of such results will immediately give you a zero grade.
- Do not skip computation steps in your proof. Clearly explain what transformations were done and why they are allowed in each step of a calculation.
- You should use correct LaTeX notation to write equations and mathematical symbols. You should encompass these equations in appropriate symbols ("\\(" and "\\)" for inline math, "\\[" and "\\]" for block math) to enhance the clarity of your proof. Do not use any unicode characters.
- Your proof should be self-contained.
- If you are not sure about a specific step, or do not know how to prove an intermediate result, clearly state this. It is much preferable to indicate your uncertainty rather than making incorrect statements or claims.
</model_prompt>

### How the solution should be graded:
A solution should be considered complete if it has described every step of the proof with clear logical reasoning. Do NOT verify the overall correctness of the proof, but only check if sufficient details are presented to make the proof complete.

#### Small mistakes that do not affect the overall completeness of the proof:
- Makes a small computational mistake that can be easily fixed
- Misses an edge case which can be easily proven/disproven
- Skips over a step that follows without much reasoning or manual work

#### The following mistakes should be considered as making the proof incomplete:
- It marks a step as trivial, if it is not immediately obvious with little reasoning why this would be the case.
- It omits algebra-heavy computational steps, regardless of whether or not it has outlined the methodology. Skipping shorter computations should be permitted.
- Generalizes over a pattern without rigorously describing the pattern, or without proving any relevant properties.
- It cites a non-existing or unpopular source/Theorem, which cannot be immediately found from searching for it online. Thus, any theorems that can be immediately found and have a Wikipedia article are allowed.

The model has been specifically told that it should not skip steps or mark them as trivial. Any violation of this rule should be considered by assuming the model does not know how to derive the "trivial" step.

### Scoring instructions

If you believe the proof is complete, end your analysis with \\boxed{{complete}}. If you believe the proof is incomplete, end your analysis with \\boxed{{incomplete}}.

### Problem Statement:
{problem}

### Model Solution:
{solution}
\end{prompt}

\subsection{Clustering Prompts} \label{app:clustering_prompts}
\paragraph{Solution Summary Generation} We use the following prompt to cluster different solutions for easier parsing by the diversity clustering LLM. We find this significantly improves the consistency of the clustering, as it provides a concise, structured overview of the solutions, allowing the model to better identify the core techniques of each solution, without getting lost in the details of the full text.

\begin{prompt}{Solution Summary Generation Prompt}
You are an expert mathematical annotator tasked with extracting the core method fingerprint of a provided solution.

# Main task

Identify the central mathematical idea that makes this particular solution work.

Focus on the method actually present in the provided solution text.  
Do not solve the problem yourself.  
Do not repair missing steps.  
Do not add steps that are not explicitly supported by the solution.

# Analysis

Analyze the solution at three levels:

1. **Family / viewpoint**  
The broad style, such as algebraic, geometric, combinatorial, constructive, extremal, invariant-based.

2. **Backbone**  
The main structural move that sets up the solution, such as a reflection construction, polynomial-root encoding, greedy maintenance strategy, block decomposition, reduction to a theorem, or symmetry-based transformation.

3. **Closing engine**  
The theorem, lemma, or decisive move that turns the backbone into the conclusion.

The **backbone** is the most important field.  
The **closing engine** should be recorded when it is a real mathematical ingredient, not just routine simplification.

# Requirements

1. Identify what makes the solution work conceptually, not how to carry out the computations.
2. Do not solve the problem on your own.
3. Do not include chain-of-thought, hidden reasoning, or speculative reconstruction.
4. Do not fill in missing steps.
5. Work only with what is explicitly present or very strongly signaled in the solution text.
6. If the solution is incomplete but its intended method is still recognizable, describe that intended method.
7. If the text is too vague to identify a genuine nontrivial method, set `noCoreIdea` to `true` and use `null` where appropriate.
8. Avoid generating equations or calculations yourself. Short verbatim evidence quotes may contain formulas if they appear in the provided solution.
9. Write `coreIdea` and `supportingIdeas` in an imperative, hint-like style.
10. Keep labels method-specific. Avoid broad labels such as "use algebra," "use symmetry," or "use geometry" unless no more specific description is supported.

# Output format

Your response must include:

- **noCoreIdea**: whether the provided solution lacks an identifiable nontrivial core idea
- **solutionSummary**: a concise summary of the solution's method and structure, written descriptively
- **coreIdea**: one short imperative sentence giving the main hint
- **supportingIdeas**: zero to three short imperative phrases for secondary ideas
- **family**: the broad mathematical viewpoint
- **backbone**: the main structural move that defines the method
- **closingEngine**: the decisive theorem, lemma, or final mechanism, or `null` if none is identifiable
- **evidenceQuotes**: one to three short quotes or pinpointed references from the solution supporting your identification
- **confidence**: a number between $0$ and $1$

# Style guidelines

- `solutionSummary` should be descriptive, compact, and reconstruction-friendly.
- `coreIdea` should be short, concrete, and method-level.
- `supportingIdeas` should name supporting moves, not computations.
- Do not mention steps that are absent from the solution.
- Do not overstate confidence when the proof is sketchy or incomplete.

# Schema

```json
{{
"noCoreIdea": false,
"solutionSummary": "",
"coreIdea": "",
"supportingIdeas": [],
"family": "",
"backbone": "",
"closingEngine": null,
"evidenceQuotes": [],
"confidence": 0.0
}}
```
# Notes

- Use `noCoreIdea = true` only when the solution text does not support a recognizable central method.
- If the proof is short but clearly method-driven, do **not** mark it as lacking a core idea.
- If the proof names a theorem but does not actually use it, do not treat that theorem as the backbone or closing engine.
- If two descriptions would differ only in wording, prefer the more concrete mathematical description.

# Input

## Problem
{problem}

## Solution
{solution}
\end{prompt}
\paragraph{Solution Summary Clustering} We use the following prompt to cluster solutions by technique.

\begin{prompt}{Solution Clustering Prompt}
Your task is to analyze multiple candidate proofs of the same problem and cluster them by their core mathematical method.

# Main Goal

Given one problem and $N$ proof summaries, determine which proofs are essentially the same method and which are meaningfully different methods.

Your job is not to grade correctness or completeness, but to detect diversity of methods in a way that is stable and consistent.

# Pairwise consistency rule

For any pair of proofs, the same-cluster / different-cluster decision should be the same whether those two proofs are shown alone or inside a larger batch, regardless of the presence of other proofs.

# Method analysis framework

Analyze each prof at three levels:

1. **Family / viewpoint**  
The broad style, such as algebraic, geometric, combinatorial, constructive, extremal, invariant-based.

2. **Backbone**  
The main structural move that sets up the proof,  i.e. a geometric construction, polynomial-root encoding, decomposition, extremal choice, or reduction to a known theorem.

3. **Closing engine**  
The theorem, lemma, or decisive move that links the backbone and the conclusion.

## Clustering rule
Cluster primarily by the **backbone**.

Use the **closing engine** to split proofs only when it is an essential, non-routine reason the proof works.

Do **not** cluster by broad family alone.

# Pairwise clustering criteria

## Put two proofs in the same cluster if:
- they share the same backbone, and
- they reach the result for the same essential mathematical reason,
even if they differ in wording, notation, order, level of detail, or minor helper tools.

## Put two proofs in different clusters only if you can clearly state:
- "One proof hinges on X, while the other hinges on Y,"
or
- "One proof translates the problem into X, while the other instead uses Y,"

If you cannot articulate a clear method-level difference, prefer merging.

# Differentiating the proof scaffold vs engine

If two proofs share the same setup or initial construction, do not automatically merge them. Instead, check whether they differ in the following manner:
- Same scaffold + same essential reason for finishing => same cluster
- Same scaffold + different essential non-routine finishing engine => different clusters
- Same scaffold + different routine algebra => same cluster

# Determining equivalent methods

Treat two proofs as the same method when they use mathematically equivalent realizations of the same backbone, even if they:
- introduce different auxiliary objects
- phrase the same construction differently
- derive the same key relation through different local manipulations
- present the same symmetry or transform through different language

If the mathematical role is the same, prefer merging.

# What does not make a new cluster

Do not split proofs just because of:
- different notation or variable names
- different exposition style
- different order of steps
- one proof being more polished or detailed
- naming a theorem versus using it implicitly
- minor local casework inside the same global plan
- different algebraic packaging of the same derived relation
- a cosmetic lemma that merely restates the key step
- a different auxilary object playing the same structural role

# What makes a new cluster

Split proofs when they have a genuinely different:
- strategic paradigm
- representation of the problem
- backbone construction or reduction
- invariant, extremal quantity, or inductive mechanism
- decisive theorem or lemma, if that theorem is the real engine
- constructive mechanism
- reason the proof closes

# Cluster naming rule

Labels such as "algebraic," "geometric," "reflection," "greedy," "Vieta," "casework," or "symmetry" are too broad by themselves.

A cluster name should identify the actual backbone and, when essential, the closing engine.

# Incorrect or incomplete proofs

- If a proof is incorrect but clearly intends a recognizable method, cluster it by intended method.
- If it is too vague to identify a real method, place it in an `Unclear / Non-proof` cluster.
- Do not invent missing key steps.
- Do not assume a standard argument unless the text supports it.

# Pipeline

## Step 1: Build a method fingerprint for each proof
For each `Proof i`, identify:
- primary approach
- secondary techniques
- key pivot step
- evidence quotes

The fingerprint must be based only on what appears in the proof.

## Step 2: Make pairwise equivalence judgments
For each pair, ask:
"Would a knowledgeable mathematician describe these as the same proof idea?"

This judgment must depend only on the two proofs, not on the rest of the batch.

## Step 3: Form clusters
Create clusters from those pairwise judgments.

Each cluster must include:
- a concise cluster name
- the defining approach
- member proofs
- a short list of distinct features

## Step 4: Consistency check
Before finalizing:
- merge clusters that differ only in phrasing
- split clusters that hide clearly different essential engines
- check whether any near-duplicate proofs were accidentally separated
- check whether any proofs were merged only because they share a broad family label

## Output format

Return exactly the following JSON shape, with no extra fields and no commentary outside the JSON.

```json
{{
"N": 0,
"K": 0,
"diversity_score_D": 0.0,
"clusters": [
    {{
    "cluster_id": "C1",
    "cluster_name": "",
    "defining_approach": "",
    "defining_features": ["", ""],
    "members": [1]
    }}
],
"proof_fingerprints": [
    {{
    "proof_id": 1,
    "primary_approach": "",
    "secondary_techniques": ["", ""],
    "key_pivot_step": "",
    "evidence_quotes": ["", ""]
    }}
],
"warnings": ["", ""]
}}
```

## Input

### Problem

{problem}

### Solutions

{solution_summary}
\end{prompt}

\subsection{Topic Classification Prompt}\label{app:topic_classification}
We use the following prompt to classify each problem into one of four mathematical topics: Algebra, Geometry, Number Theory, and Combinatorics.
\begin{prompt}{Topic Classification Prompt}
Your task is to classify the following math problem into one of several topics based on its content and the mathematical concepts it involves. The topics to choose from are:
- Algebra
- Geometry
- Number Theory
- Combinatorics
- Calculus

Output the topic that best fits the problem in \\boxed{{}}, i.e. \\boxed{{Algebra}}, at the end of your response.

Here are the problem statement and solution:

Problem:

{problem}

Solution:

{solution}
\end{prompt}
\subsection{Solution Shortening Prompt}\label{app:conciseness_rephraser}
We use the following prompt to shorten LLM-generated proofs to estimate their textual conciseness.
\begin{prompt}{Solution Shortening Prompt}
Your task is to rephrase the following solution to a mathematical problem to make it less verbose without omitting any important details. Ensure that the rephrased solution remains clear and accurate while being more concise. Follow these guidelines:
- Retain all critical steps and explanations necessary for understanding the solution.
- Do not introduce any new information or change the meaning of the original solution.
- The new solution must be equivalent to the original in terms of correctness and completeness.
- Do not omit any computations that were critical to the solution.


Original Problem:
{problem}

Original Solution:
{solution}
\end{prompt}
\subsection{Shortening Verification Prompt}\label{app:verbosity_verifier}
We use the following prompt to verify whether the shortened proofs are equivalent to the original.
\begin{prompt}{Shortening Verification Prompt}
Your task is to verify whether the following solution to a mathematical problem has been successfully rephrased to be less verbose while retaining all important details. Follow these guidelines:
- Check that all critical steps and explanations necessary for understanding the solution are still present.
- Ensure that no new information has been introduced and that the meaning of the original solution has not been changed.
- Confirm that the rephrased solution is equivalent to the original in terms of correctness and completeness.
- Verify that no computations critical to the solution have been omitted.
- If the rephrased solution meets all these criteria, conclude with "\\boxed{{Yes}}". If it fails any of these criteria, conclude with "\\boxed{{No}}".


Original Problem:
{problem}

Original Solution:
{solution}

Shortened Solution:
{outputs}
\end{prompt}

\subsection{Technique Verification Prompt}\label{app:technique_verification}
In the following, we provide the prompt used to verify whether a given solution correctly uses a specified technique to solve a mathematical problem. Correctness is established separately by the filter of \cref{sec:methodology-dataset}, so this judge only assesses technique usage.

\begin{prompt}{Technique Verification Prompt}
Your task is to verify whether the following solution to a mathematical problem has correctly used a pre-specified technique to solve the problem.

### Input:

Your input will consist of the following components:
- **Problem Statement**: A mathematical problem that the proof is attempting to solve.
- **Required Technique**: A specific mathematical technique that the proof is required to use in order to solve the problem.
- **Proof Solution**: The proof that you need to evaluate. This proof may contain errors, omissions, or unclear steps. The proof was generated by another language model, which was given the following instructions:
<model_prompt>
- You are creating a proof, not a proof outline. Each step should be carefully explained and documented. If not properly explained, the judge will assume that you cannot explain it, and therefore decrease your grade.
- You can use general theorems and lemmas, but only if they are well-known. As a rule of thumb: if the result has a name and is famous enough to have a Wikipedia page or something similar to describe it, it is allowed. Any result from papers that would not be taught in high-school or low-level bachelor courses in mathematics should not be used. Any use of such results will immediately give you a zero grade.
- Do not skip computation steps in your proof. Clearly explain what transformations were done and why they are allowed in each step of a calculation.
- You should use correct LaTeX notation to write equations and mathematical symbols. You should encompass these equations in appropriate symbols ("\\(" and "\\)" for inline math, "\\[" and "\\]" for block math) to enhance the clarity of your proof. Do not use any unicode characters.
- Your proof should be self-contained.
- Your proof should make use of the specified technique. You will not receive any credit if you fail to use the specified technique.
- If you are not sure about a specific step, or do not know how to prove an intermediate result, clearly state this. It is much preferable to indicate your uncertainty rather than making incorrect statements or claims.
</model_prompt>

Assume that the solution as a whole is correct, unless you easily find a clear error in the solution. You should not assume that the solution is incorrect just because it is not clear to you, or because it is not well explained.

The model has been specifically told that it should not skip steps or mark them as trivial. Any violation of this rule should be considered by assuming the model does not know how to derive the "trivial" step.

### How the solution should be graded with respect to technique usage:

The solution should be considered to have correctly used the specified technique if:
- You can identify that the solution has explicitly stated the use of the technique, or an equivalent method.
- The solution has correctly applied the technique, and has not simply wrapped a different method using the name of the specified technique.
- The solution has made significant progress towards solving the problem using the specified technique. This means that the technique should be central to the solution, and not just mentioned in passing or used for minor steps.

The solution should be considered to have incorrectly used the specified technique if:
- The solution does not mention the technique at all, or only mentions it in passing without using it in any significant way.
- The solution uses a different technique that is not equivalent to the specified technique.
- The solution only uses the specified technique for minor steps, while the majority of the solution relies on other techniques.

### Scoring instructions

If you believe the proof is correct and has used the specified technique, end label the solution as correct. If you believe the proof is incorrect, or it has not followed the specified technique, end your analysis with incorrect.

### Invalid Technique Specification

Sometimes, the technique specified may not exist as a mathematical technique or theorem, but is instead a fact that appears often in mathematical competitions, which has been given a name. In cases where you are unaware of this technique, you should output "warning".

### Output format

At the end of your reasoning, output the following structure:

```json
{{
"verdict": "<one of: correct, incorrect, warning>",
"reason": "<one of: correctness, technique, both, unknown>",
"comments": "<your detailed reasoning here>"
}}
```

### Problem Statement:
{problem}

### Required Technique:
{technique}

### Model Solution:
{solution}
\end{prompt}

\subsection{Pairwise Comparison Prompts} \label{app:pairwise_prompts}

\paragraph{Pairwise Cognitive Simplicity Comparison Prompt}
We use the following prompt to compare the cognitive simplicity of different solutions based on their solving techniques.

\begin{prompt}{Pairwise Cognitive Simplicity Comparison Prompt}
You are an impartial LLM-as-a-judge. Your task is to compare Solution 1 vs Solution 2 and decide which one exhibits MORE **ideological complexity** to a technically competent reader, specifically due to:
- **Ingenuity / non-obviousness of the core idea** (key trick, surprising invariant, clever construction, unexpected transformation, "aha" step).
- **Niche-ness / sophistication of mathematical tools** (specialized lemmas, advanced theorems, subject-specific machinery; e.g., olympiad-specific techniques vs routine algebra; whether it relies on tools unlikely to be known outside contest/upper-undergrad/graduate contexts).
- **Difficulty of integrating multiple ideas** (linking several distinct concepts/lemmas; multi-layer strategy; reduction chains; combining geometry + algebra + number theory; nontrivial case architecture as a conceptual device).

Do NOT attempt to solve the problem and do NOT aim to determine which solution is correct.

Key definition (use consistently):
"Ideological complexity" means the degree of **conceptual ingenuity, specialized tool use, and difficulty of linking multiple ideas** in the presented solution. It is not about notation density, step count, or computation/time-to-follow.
Judge from: "How conceptually demanding is the *method/idea stack* here?" NOT: "How long does it take to parse the notation or execute computations?"

Complexity interpretation:
- Treat "overall complexity" as an estimate of **conceptual sophistication and methodological depth**, not time-to-follow.
- A short proof can be highly complex if it uses a deep theorem or a very non-obvious idea.
- A long proof can be low complexity if it is mostly routine expansions, bookkeeping, or standard steps.

Do NOT consider:
- **Computational/cognitive load as written**: long algebra, tedious arithmetic, dense notation, symbol tracking, or bookkeeping time.
- **Verbosity or step count** unless it reflects *genuinely* multiple conceptual ingredients (not just expanded algebra).
- Missing derivations as "hard work" unless the omission clearly hides a deep theorem/idea (again discounting routine algebraic manipulations).
- Correctness, rigor, or whether gaps can be filled. Do not fact-check theorem applicability.

What DOES count toward ideological complexity:
1) **Ingenuity of the key move**
- Non-standard substitution or viewpoint shift (e.g., turning a Diophantine problem into a geometric/graph/invariant argument).
- Introduction of an invariant/monovariant, extremal principle, or clever construction that is not routine.
- A reduction that is conceptually subtle (e.g., "encode as generating function," "apply probabilistic method," "use compactness," etc.).

2) **Tool sophistication / niche-ness**
- Use of advanced or specialized results (e.g., Jensen/Karamata/Muirhead in a nontrivial way; lifting exponent, LTE; Zsigmondy; projective geometry lemmas; complex numbers/barycentric coordinates; group actions; p-adics; generating functions; spectral methods; etc.).
- Reliance on domain-specific frameworks (e.g., functional equations classification tricks, invariant theory, combinatorial nullstellensatz, etc.).
- The extent to which the proof requires familiarity with "IMO-level" technique stacks versus broadly taught basics.

3) **Integration burden (conceptual linking)**
- Number of distinct ideas that must be coordinated (e.g., inequality + symmetry + convexity + tangent line method).
- Multi-stage architecture (reduction $\to$ lemma $\to$ transformation $\to$ final synthesis).
- Nontrivial case splits that represent fundamentally different conceptual regimes (not mere arithmetic branching).

Edge-case handling rules (apply as needed):
1) Extremely short vs very detailed:
- Do not treat brevity as simplicity; judge the *depth of ideas/tools* used.
2) Do not judge correctness or fill gaps; judge what the solution *claims to use*.
3) If both solutions are too high-level/vague to identify tools/ideas (e.g., "clearly follows" with no method indicated), output "0" (Tie/Indeterminate) due to insufficient evidence.
4) If one solution is computation-heavy but conceptually routine, it should rate LOW on ideological complexity even if it is hard to follow.
5) If a solution invokes a deep theorem without explanation, you MAY count that as high ideological complexity (tool sophistication), but do not penalize the other for not expanding computations.

You must perform a PAIRWISE COMPARISON ONLY:
- Output a single verdict: "1", "2", or "0" (Tie).
- Provide brief reasoning citing concrete features from the solutions (quote short snippets or refer to distinctive phrases like "apply $\dots$ lemma/theorem," "consider invariant," "use generating function," etc.), but do not expand into solving steps.


Do not attempt to complete/solve the problem:
- Do not compute final answers.
- Do not re-derive results to check correctness.
- Do not introduce new math/logic beyond describing the conceptual/tooling complexity characteristics of what is already written.
- Do not "repair" a solution, propose alternatives, or add missing steps.

Tie rules (must follow; explicit):
Return "Tie/Indeterminate" if and only if at least one of the following holds:
1) **Near-equal complexity:** Neither solution is clearly more conceptually sophisticated after comparing ingenuity, tool niche-ness, and integration.
2) **Near-identical methodology** Both solutions rely on the same core concepts, tools and strategy, with a similar method of execution.
3) **Orthogonal tradeoffs:** One uses a deep theorem but in a single-step way, while the other uses several moderately advanced ideas in an integrated way, and you cannot confidently rank overall ideological complexity without guessing hidden details.
4) **Insufficient evidence:** One or both solutions are too vague/underspecified to identify the conceptual toolkit or strategy.
5) **Both equally sophisticated:** Both rely on similarly niche tools and similarly non-obvious strategy.

Non-tie constraint:
Do not output Tie merely because both might be correct/incorrect/uncertain; correctness is irrelevant. Tie is only about inability to confidently rank ideological complexity from the written text alone.

Decision procedure (follow silently; do not output these steps):
- Identify the main conceptual moves and any named tools/results in Solution 1 and Solution 2.
- Compare (a) ingenuity, (b) niche-ness/sophistication of tools, (c) integration of multiple ideas.
- Choose 1/2/0 (Tie/Indeterminate) and justify with concise, text-anchored evidence.

Output format (must follow exactly; no extra sections, no bullets, no numbering):
Solution 1 Complexity: <one concise paragraph describing the ingenuity/tools/integration complexity of Solution 1, citing 1-3 concrete text anchors (short quotes or distinctive phrases).>
Solution 2 Complexity: <one concise paragraph describing the ingenuity/tools/integration complexity of Solution 2, citing 1-3 concrete text anchors (short quotes or distinctive phrases).>
Decision Reasoning: <one concise paragraph stating the verdict (1/2/0) and explaining why, directly contrasting the biggest ideological-complexity drivers; must reference at least one concrete feature from each solution.>
Confidence: High/Medium/Low.
Verdict: $$\boxed{{0 | 1 | 2}}$$

Now evaluate the following.

Problem:
{problem}

Solution 1:
{solution_1}

Solution 2:
{solution_2}
\end{prompt}
\paragraph{Pairwise Computational Ease Comparison Prompt}
We use the following prompt to compare the brute-force computational ease of different solutions based on their solving techniques.

\begin{prompt}{Pairwise Computational Ease Comparison Prompt}
You are an impartial LLM-as-a-judge. Your task is to compare Solution 1 vs Solution 2 and decide which one imposes more **notation/computation TIME load** on a careful reader, specifically due to:
- **Notation density / symbol-tracking burden** (dense symbolic expressions; many variables/indices; summations/products; matrices/tensors; deeply nested parentheses; frequent switching conventions; overloaded symbols).
- **Definition hygiene / referential clarity** (variables or functions used before definition; unclear domains/quantifiers; ambiguous constraints; inconsistent naming; redefining the same symbol for different objects; unclear dependence on parameters).
- **Computation burden as written (time-to-execute/track)** (many explicit steps even if each is "simple"; long simplifications; repeated expansions; sign/exponent/index bookkeeping; numeric work that is easy to slip on).
- **Mechanical manipulation burden** (multiple substitutions, changes of variables, rearrangements, coordinate transforms) **only insofar as** they increase bookkeeping, symbol tracking, or explicit computation/time load.

Do NOT attempt to solve the problem and do NOT aim to determine which solution is correct.

Key definition (use consistently):
"Cognitive load" here means the mental effort and **time** required for a technically competent reader to **parse and track the notation, definitions, and explicit computations** in the presented solution as written. It is not about correctness and not about the conceptual difficulty of the underlying idea.
Judge from "If I want to parse and follow what is written here line-by-line, how much time and mental effort would it take?" not "If I want to verify/certify every missing step, how hard would it be?"

Time Interpretation:
- Treat "overall load" as an estimate of **time-to-follow as written** for a careful, technically competent reader.
- **Many long but routine steps can be more time-consuming** than a small number of dense/clever steps.
- Count time cost coming from **step count + bookkeeping + symbol tracking**, not from conceptual depth.

Do NOT consider:
- Difficulty due to **conceptual/argument complexity** (e.g., using advanced theorems, clever insights, high-level strategy, non-obvious ideas), **unless** it directly increases notation/definition/computation burden on the page.
- "Leaps in logic" as a correctness/rigor issue. Treat them as relevant **only** when they make the text hard to *parse* (e.g., new symbols appear without explanation, variable roles change silently, indices/bounds are unclear).
- Whether the solution is "deep," "elegant," or "insightful."

Missing steps rule:
Do not guess the amount of work required to fill in omitted derivations unless the text itself explicitly introduces **notation/definition/computation opacity** (e.g., new unintroduced symbols appear, variable meanings shift, indices/bounds/domains are unstated).

You must perform a pairwise comparison:
- Output a single verdict: "1", "2", or "0" (Tie).
- Provide brief reasoning that cites concrete features from the solutions (quote short snippets or refer to distinctive phrases), but do not expand into solving steps.

STRICT NON-SOLVING RULES (must follow):
- Do not compute final answers.
- Do not re-derive results to check correctness.
- Do not introduce new math/logic beyond describing notation/definition/computation complexity characteristics of what is already written.
- Do not "repair" a solution, propose alternatives, or add missing steps.
- Do not reward/penalize based on verbosity alone **when it is purely redundant prose**; however, **do** count verbosity that materially increases **time-to-follow** because it adds many explicit computational/notation-tracking steps.

What to compare (qualitative rubric; no scores):
1) **Notation & symbol-tracking load (time to parse)**
- Many symbols/indices; nested expressions; heavy $\sum/\prod$ notation; matrices; function composition chains.
- Overloaded symbols; notation churn; switching conventions mid-stream.
- Symbols used before being defined; unclear bounds/indices/domains.

2) **Definition hygiene / clarity of references (time lost to ambiguity)** (weigh this factor less heavily than notational and algebraic load)
- Whether each variable/function/set is clearly introduced when first used.
- Whether constraints and quantifiers (for all/exists, domain restrictions) are explicit.
- Whether it's clear what is fixed vs varying, and what depends on what.

3) **Explicit computation & bookkeeping load (time to carry out as written)**
- Long algebraic manipulations, simplifications, expansions, casewise numeric work.
- Error-prone sign/exponent/index tracking; multiple intermediate quantities to remember.
- Multiple substitutions/rewrites that create bookkeeping overhead (regardless of conceptual motivation).
- **High step-count matters:** a long chain of straightforward arithmetic/algebra can dominate time even if each step is easy.

Edge-case handling rules (apply as needed):
1) Extremely short vs very detailed:
- A one-line claim can be low time/load to read even if conceptually deep; do not penalize conceptual depth.
- If both are too thin to compare on notation/definition/computation burden, output "0" (Tie/Indeterminate).
2) Do not judge correctness or theorem applicability; judge **notation/definition/computation time-to-follow**.
3) Do not fix apparent errors/contradictions.
4) Penalize frequent redefinition of variables, switching conventions, or overloading symbols because it increases tracking burden and time.
5) Many tiny trivial steps can be **more time-consuming overall** than a few dense steps; judge **overall time-to-follow**, considering both density and total step/bookkeeping count.

Tie rules

Do not be too strict in judging one way or another if there is not a clearly dominating solution in terms of computational load. You can call a Tie if the solutions are close enough that a reasonable reader might disagree on which is heavier, or if they have different types of load that are hard to compare. Use the following criteria to decide whether to return **"Tie/Indeterminate"** if you are uncertain which solution has more notation/computation load:
1) **Near-equal time/load:** After comparing notation/symbol-tracking, definition hygiene, and explicit computation/bookkeeping, neither solution will clearly dominate the other from a human perspective. For example, if the notation introduced is not heavy to follow, that aspect would not matter for your decision-making.
2) **Similar Solutions:** If the solution structures are for the most part similar, or the same, and the differences are mostly in surface-level details that do not materially affect the time-to-follow (e.g., one is slightly more verbose, while the other is more concise but uses slightly denser notation), you may judge them as a Tie.
3) **Orthogonal tradeoffs:** One solution is heavier mainly in **notation density**, while the other is heavier mainly in **long step-by-step computation**, and you cannot confidently rank overall time-to-follow without guessing reader preferences.
4) **Insufficient evidence:** One or both solutions are too short/underspecified to assess notation/definition/computation burden **as written** (e.g., both are mostly high-level claims with minimal notation and no explicit computation), so a comparison would require guessing omitted details.
5) **Both equally problematic:** Both exhibit similar levels of symbol overload, undefined variables, or dense/long manipulations such that a clear "more time/load" choice cannot be justified from the text alone.

Decision procedure (follow silently; do not output these steps):
- Identify the main notation/definition/computation time-load drivers in Solution 1 and in Solution 2.
- Compare which would demand more careful symbol tracking and computation/bookkeeping time to parse.
- Choose 1/2/0 (Tie/Indeterminate) and justify with concise, text-anchored evidence.

Output format (must follow exactly; no extra sections, no bullets, no numbering):
Solution 1 Load: <one concise paragraph describing the notation/definition/computation burden in Solution 1, citing 1-3 concrete text anchors (short quotes or distinctive phrases).>
Solution 2 Load: <one concise paragraph describing the notation/definition/computation burden in Solution 2, citing 1-3 concrete text anchors (short quotes or distinctive phrases).>
Decision Reasoning: <one concise paragraph stating the verdict (1/2/0) and explaining *why* it was chosen by directly contrasting the biggest time/load drivers; must reference at least one concrete feature from each solution.>
Confidence: High/Medium/Low.
Verdict: $$\boxed{{0 | 1 | 2}}$$

Now evaluate the following.

Problem:
{problem}

Solution 1:
{solution_1}

Solution 2:
{solution_2}
\end{prompt}

\clearpage
\section*{NeurIPS Paper Checklist}

\begin{enumerate}

\item {\bf Claims}
    \item[] Question: Do the main claims made in the abstract and introduction accurately reflect the paper's contributions and scope?
    \item[] Answer: \answerYes{} % Replace by \answerYes{}, \answerNo{}, or \answerNA{}.
    \item[] Justification: The abstract and introduction clearly state the main claims of the paper, including the introduction of \bench, the evaluation of multiple proof quality metrics, and the findings regarding differences in proof quality across models. The claims are supported by the experimental results presented in the paper.

\item {\bf Limitations}
    \item[] Question: Does the paper discuss the limitations of the work performed by the authors?
    \item[] Answer: \answerYes{} % Replace by \answerYes{}, \answerNo{}, or \answerNA{}.
    \item[] Justification: Yes, we specifically discuss limitations in \cref{sec:limitations}.

\item {\bf Theory assumptions and proofs}
    \item[] Question: For each theoretical result, does the paper provide the full set of assumptions and a complete (and correct) proof?
    \item[] Answer: \answerNA{} % Replace by \answerYes{}, \answerNo{}, or \answerNA{}.
    \item[] Justification: We do not include theoretical results in this paper, so this question is not applicable.

    \item {\bf Experimental result reproducibility}
    \item[] Question: Does the paper fully disclose all the information needed to reproduce the main experimental results of the paper to the extent that it affects the main claims and/or conclusions of the paper (regardless of whether the code and data are provided or not)?
    \item[] Answer: \answerYes{} % Replace by \answerYes{}, \answerNo{}, or \answerNA{}.
    \item[] Justification: Yes, we provide detailed information in our paper, and release both our code and data to ensure that our results are reproducible. We also provide detailed instructions for how to reproduce our results in the code.

\item {\bf Open access to data and code}
    \item[] Question: Does the paper provide open access to the data and code, with sufficient instructions to faithfully reproduce the main experimental results, as described in supplemental material?
    \item[] Answer: \answerYes{} % Replace by \answerYes{}, \answerNo{}, or \answerNA{}.
    \item[] Justification: Yes, as required by the evaluation and datasets track, we make our code and data publicly available, and provide detailed instructions for how to reproduce our results.

\item {\bf Experimental setting/details}
    \item[] Question: Does the paper specify all the training and test details (e.g., data splits, hyperparameters, how they were chosen, type of optimizer) necessary to understand the results?
    \item[] Answer: \answerYes{} % Replace by \answerYes{}, \answerNo{}, or \answerNA{}.
    \item[] Justification: Yes, we specifically provide all inference parameters in \cref{app:additional-experimental-details}, and provide instructions for how to reproduce our results in the code.

\item {\bf Experiment statistical significance}
    \item[] Question: Does the paper report error bars suitably and correctly defined or other appropriate information about the statistical significance of the experiments?
    \item[] Answer: \answerYes{} % Replace by \answerYes{}, \answerNo{}, or \answerNA{}.
    \item[] Justification: Yes, we include the $95\%$ confidence intervals for our main results in \cref{app:main-results-ci}.

\item {\bf Experiments compute resources}
    \item[] Question: For each experiment, does the paper provide sufficient information on the computer resources (type of compute workers, memory, time of execution) needed to reproduce the experiments?
    \item[] Answer: \answerYes{} % Replace by \answerYes{}, \answerNo{}, or \answerNA{}.
    \item[] Justification: Yes, we provide detailed information on the compute resources used for our experiments in \cref{app:additional-experimental-details}.
    
\item {\bf Code of ethics}
    \item[] Question: Does the research conducted in the paper conform, in every respect, with the NeurIPS Code of Ethics \url{https://neurips.cc/public/EthicsGuidelines}?
    \item[] Answer: \answerYes{} % Replace by \answerYes{}, \answerNo{}, or \answerNA{}.
    \item[] Justification: Yes, we abide by the NeurIPS Code of Ethics in all aspects of our research, including the design and execution of our experiments, the reporting of our results, and the way we handle data and code.

\item {\bf Broader impacts}
    \item[] Question: Does the paper discuss both potential positive societal impacts and negative societal impacts of the work performed?
    \item[] Answer: \answerYes{} % Replace by \answerYes{}, \answerNo{}, or \answerNA{}.
    \item[] Justification: Yes, we discuss broader impact in \cref{app:impact}. We do not see any negative societal impacts that form risks for misuse.
    
\item {\bf Safeguards}
    \item[] Question: Does the paper describe safeguards that have been put in place for responsible release of data or models that have a high risk for misuse (e.g., pre-trained language models, image generators, or scraped datasets)?
    \item[] Answer: \answerNA{} % Replace by \answerYes{}, \answerNo{}, or \answerNA{}.
    \item[] Justification: No, the paper does not pose such risks for misuse.

\item {\bf Licenses for existing assets}
    \item[] Question: Are the creators or original owners of assets (e.g., code, data, models), used in the paper, properly credited and are the license and terms of use explicitly mentioned and properly respected?
    \item[] Answer: \answerYes{} % Replace by \answerYes{}, \answerNo{}, or \answerNA{}.
    \item[] Justification: Yes, our only data sources are MathArena \citep{matharena} and IMOBench \citep{imobench}, and we follow the licenses and terms of use for both datasets. Although we use AoPS data for capturing diverse problems and techniques, we only release the technique list we consider, without any identifiable information, or details from the community-sourced solutions.

\item {\bf New assets}
    \item[] Question: Are new assets introduced in the paper well documented and is the documentation provided alongside the assets?
    \item[] Answer: \answerYes{} % Replace by \answerYes{}, \answerNo{}, or \answerNA{}.
    \item[] Justification: Yes, as required by the evaluation and datasets track, we make our code and data publicly available, and provide detailed documentation for how to use our code and data, including their license.

\item {\bf Crowdsourcing and research with human subjects}
    \item[] Question: For crowdsourcing experiments and research with human subjects, does the paper include the full text of instructions given to participants and screenshots, if applicable, as well as details about compensation (if any)?
    \item[] Answer: \answerYes{} % Replace by \answerYes{}, \answerNo{}, or \answerNA{}.
    \item[] Justification: Nearly all human judgments are internal annotations by paper authors for validation. A single person outside the author team contributed a small number of additional annotations, and is acknowledged in the paper. We do not use crowdsourcing. The full text of the instructions given to all annotators is provided in \cref{app:human-validation-instructions}.

\item {\bf Institutional review board (IRB) approvals or equivalent for research with human subjects}
    \item[] Question: Does the paper describe potential risks incurred by study participants, whether such risks were disclosed to the subjects, and whether Institutional Review Board (IRB) approvals (or an equivalent approval/review based on the requirements of your country or institution) were obtained?
    \item[] Answer: \answerNA{} % Replace by \answerYes{}, \answerNo{}, or \answerNA{}.
    \item[] Justification: Human judgments are internal annotations by paper authors, apart from a small number of additional annotations contributed by a single colleague outside the author team, as requested by reviewers. The task consists solely of comparing pairs of machine-generated mathematical proofs, involves no personal data and no foreseeable risk to participants, and therefore did not require IRB review at our institution.

\item {\bf Declaration of LLM usage}
    \item[] Question: Does the paper describe the usage of LLMs if it is an important, original, or non-standard component of the core methods in this research? Note that if the LLM is used only for writing, editing, or formatting purposes and does \emph{not} impact the core methodology, scientific rigor, or originality of the research, declaration is not required.
    %this research? 
    \item[] Answer: \answerYes{} % Replace by \answerYes{}, \answerNo{}, or \answerNA{}.
    \item[] Justification: We explain and document our usage of LLMs to grade our metrics extensively in \cref{sec:methodology}.

\end{enumerate}

%%%%%%%%%%%%%%%%%%%%%%%%%%%%%%%%%%%%%%%%%%%%%%%%%%%%%%%%%%%%%%%%%%%%%%%%%%%%%%%
%%%%%%%%%%%%%%%%%%%%%%%%%%%%%%%%%%%%%%%%%%%%%%%%%%%%%%%%%%%%%%%%%%%%%%%%%%%%%%%

\end{document}